# Foundation Models - A Panacea for Artificial Intelligence in Pathology?

Nita Mulliqi[1], Anders Blilie[2,3], Xiaoyi Ji[1], Kelvin Szolnoky[1], Henrik Olsson[1], Sol Erika Boman[1,4], Matteo Titus[1], Geraldine Martinez Gonzalez[1], Julia Anna Mielcarz[1], Masi Valkonen[5], Einar Gudlaugsson[2], Svein R. Kjosavik[6,7], José Asenjo[8], Marcello Gambacorta[9], Paolo Libretti[9], Marcin Braun[10], Radzislaw Kordek[10], Roman Łowicki[11], Kristina Hotakainen[12,13], Päivi Väre[14], Bodil Ginnerup Pedersen[15,16], Karina Dalsgaard Sørensen[16,17], Benedicte Parm Ulhøi[18], Pekka Ruusuvuori[5,19,20], Brett Delahunt[21,22], Hemamali Samaratunga[23], Toyonori Tsuzuki[24], Emilius A.M. Janssen[2,25,26], Lars Egevad[22], Martin Eklund[1], Kimmo Kartasalo[27]

1. Department of Medical Epidemiology and Biostatistics, Karolinska Institutet, Stockholm, Sweden
2. Department of Pathology, Stavanger University Hospital, Stavanger, Norway
3. Faculty of Health Sciences, University of Stavanger, Stavanger, Norway
4. Department of Molecular Medicine and Surgery, Karolinska Institutet, Stockholm, Sweden
5. Institute of Biomedicine, University of Turku, Turku, Finland
6. The General Practice and Care Coordination Research Group, Stavanger University Hospital, Norway
7. Department of Global Public Health and Primary Care, Faculty of Medicine, University of Bergen, Norway
8. Department of Pathology, Synlab, Madrid, Spain
9. Department of Pathology, Synlab, Brescia, Italy
10. Department of Pathology, Chair of Oncology, Medical University of Lodz, Lodz, Poland
11. 1st Department of Urology, Medical University of Lodz, Lodz, Poland
12. Department of Clinical Chemistry and Hematology, University of Helsinki, Helsinki, Finland
13. Laboratory Services, Mehiläinen Oy, Helsinki, Finland
14. Department of Pathology, Mehiläinen Länsi-Pohja Hospital, Kemi, Finland
15. Department of Radiology, Aarhus University Hospital, Aarhus, Denmark
16. Department of Clinical Medicine, Aarhus University, Aarhus, Denmark
17. Department of Molecular Medicine, Aarhus University Hospital, Aarhus, Denmark
18. Department of Pathology, Aarhus University Hospital, Aarhus, Denmark
19. InFLAMES Research Flagship, University of Turku, Turku, Finland
20. Faculty of Medicine and Health Technology, Tampere University, Tampere, Finland
21. Malaghan Institute of Medical Research, Wellington, New Zealand
22. Department of Oncology and Pathology, Karolinska Institutet, Stockholm, Sweden
23. Aquesta Uropathology and University of Queensland, QLD, Brisbane, Australia
24. Department of Surgical Pathology, School of Medicine, Aichi Medical University, Nagoya, Japan
25. Department of Chemistry, Bioscience and Environmental Engineering, University of Stavanger, Stavanger, Norway
26. Institute for Biomedicine and Glycomics, Griffith University, Queensland, Australia
27. Department of Medical Epidemiology and Biostatistics, SciLifeLab, Karolinska Institutet, Stockholm, Sweden

Corresponding author: Kimmo Kartasalo, kimmo.kartasalo@ki.se.

# Abstract

The role of artificial intelligence (AI) in pathology has evolved from aiding diagnostics to uncovering predictive morphological patterns in whole slide images (WSIs). Recently, foundation models (FMs) leveraging self-supervised pre-training have been widely advocated as a universal solution for diverse downstream tasks. However, open questions remain about their clinical applicability and generalization advantages over end-to-end learning using task-specific (TS) models.

Here, we focused on AI with clinical-grade performance for prostate cancer diagnosis and Gleason grading. We present the largest validation of AI for this task, using over 100,000 core needle biopsies from 7,342 patients across 15 sites in 11 countries. We compared two FMs with a fully end-to-end TS model in a multiple instance learning framework. Our findings challenge assumptions that FMs universally outperform TS models. While FMs demonstrated utility in data-scarce scenarios, their performance converged with—and was in some cases surpassed by—TS models when sufficient labeled training data were available. Notably, extensive task-specific training markedly reduced clinically significant misgrading, misdiagnosis of challenging morphologies, and variability across different WSI scanners. Additionally, FMs used up to 35 times more energy than the TS model, raising concerns about their sustainability.

Our results underscore that while FMs offer clear advantages for rapid prototyping and research, their role as a universal solution for clinically applicable medical AI remains uncertain. For high-stakes clinical applications, rigorous validation and consideration of task-specific training remain critically important. We advocate for integrating the strengths of FMs and end-to-end learning to achieve robust and resource-efficient AI pathology solutions fit for clinical use.

**Keywords:** Artificial Intelligence, Clinical Validation, Computational Pathology, End-to-End Learning, Foundation Models, Generalization, Gleason Grading, Model Efficiency, Prostate Cancer, Whole Slide Imaging

# Introduction

The pathological assessment of tissue specimens remains a cornerstone of diagnostic and therapeutic decision making. The ongoing digitization of pathology has been closely followed by the development of computational methods for analyzing whole slide image (WSI) data[1]. Initially, computational pathology mainly targeted improved diagnostic efficiency and accuracy by aiding pathologists with decision support and automation. Recently, the focus has increasingly shifted towards also discovering novel morphological patterns that are currently unknown to pathologists but are predictive of clinical outcomes, therapy response, or underlying molecular biomarkers. Besides the growing amount of digitized specimens, progress in computational pathology has been fueled by developments in artificial intelligence (AI).

The first breakthroughs in pathology image analysis were triggered by abandoning traditional image processing and feature engineering in favor of end-to-end learning[2,3]. In the former, researchers handcraft quantitative descriptors of tissue morphology to feed machine learning models, whereas the latter involves training deep neural networks to discover relevant patterns without human intervention. End-to-end learning generally produces AI models with superior performance but requires large amounts of labeled training data[4,5]. Collecting sufficient annotated data to cover all pathology tasks is infeasible, particularly for rare diseases or infrequently observed molecular biomarkers. Additionally, in contrast to the earlier analysis of micrographs or small image patches, applying end-to-end learning to the more clinically relevant high-resolution WSIs has been technically challenging and has forced developers to resort to multi-step approaches featuring a chain of separately trained AI models. Typically, WSIs are split into patches and the outputs of a patch-level model are aggregated by another AI model to obtain slide- or patient-level predictions[6,7].

A string of seminal papers on foundation models (FM)[8–14] has arguably marked the beginning of a second period of rapid breakthroughs in computational pathology. This new paradigm in medical AI[15] circumvents the collection of vast datasets for supervised training of 'narrow' task-specific (TS) models. Instead, 'broad' generalist FMs are pre-trained in an unsupervised manner using an unlabeled mixture of tissue and specimen types to try and capture a generally useful quantitative representation, or feature embedding, of the data. An off-the-shelf FM can then be used as a feature encoder with very little or no task-specific training to extract descriptors of patches or WSIs for downstream applications. In this sense, FMs represent a step back from end-to-end learning to a two-stage approach, where the data are first summarized using predefined features, which are then used for training relatively simple task-specific models.

The primary focus of most studies on FMs has been their applicability as a one-size-fits-all transfer learning platform to accelerate the development of AI models for a plethora of tasks, including those with very limited training data available. This viewpoint is supported by conclusive evidence from broad (but often superficial) evaluations across a range of tissue types and pathology tasks. However, the role envisioned for FMs in medical AI extends far beyond serving as an upgrade to transfer learning models pre-trained on non-medical data (e.g. ImageNet[16]). A more controversial hypothesis underlying the 'paradigm shift' expectations placed on FMs is that the unprecedented scale of these models and their unlabeled training data will give rise to generalist AI models, which exhibit unexpected skills, or 'emergent abilities', and outperform TS models even in tasks that the 'narrow' models specialize in[8,17,18]. For research and rapid prototyping, the prospect of satisfactory baseline performance in most tasks even without any task-specific training is highly appealing. However, from a clinical standpoint, the difference between an off-the-shelf prototype and a clinical-grade model with validated robust performance in a specific task can be a matter of life and death. From this perspective, the main promise of FMs lies in their hypothesized ability to 'understand' the data, which should translate to more consistent performance in real-world clinical settings where overfitting, batch effects and various biases are significant challenges for AI[19].

Currently, there is a knowledge gap regarding the expected universal generalization advantage of FMs over end-to-end trained TS models for WSI analysis, potentially due to the sparsity of large-scale labeled datasets and the computational challenges of analyzing full WSIs. In the absence of direct comparisons, the superiority of FMs' feature embeddings compared to those tailored for a given task via end-to-end learning remains a conjecture. Handling entire WSIs is computationally demanding, but evolving graphics processing units (GPU) and algorithmic innovations are overcoming this obstacle and making end-to-end learning a feasible option for modern computational pathology[20–22]. Due to the large number of model parameters that need to be optimized jointly, end-to-end training requires a large and diverse training cohort to avoid overfitting and capturing biases due to e.g. confounding between task labels and scanner-dependent variations in WSI appearance. Collecting such a dataset is understandably not feasible in the case of e.g. rare diseases, but vast amounts of WSIs are already generated from tissue types frequently analyzed in pathology laboratories, such as prostate specimens.

Prostate cancer is the most common cancer in men globally, and more than 2 million people undergo prostate biopsy annually in Europe and the US[23,24]. Pathological assessment for diagnosing and Gleason grading is crucial for the clinical management of the disease, but considerable inter-observer variation and the limited prognostic and treatment-predictive ability of the Gleason grading system complicate decision making[25]. Applying AI to harmonize grading of prostate biopsies has long been a widely studied topic in computational pathology[4,6,7,26–28], and recent studies have also aimed at improved prognostication of patient outcomes[29] or treatment response[30]. Some prostate AI models are also already commercially available and validated to varying degrees for clinical use[31–36]. Also FMs have been applied to prostate cancer detection[8,12] and Gleason grading[9], but no extensive validation covering different patient populations, laboratories, and scanners has been reported.

Here, we approach the problem of AI-based prostate cancer diagnosis and Gleason grading with a weakly supervised, task-specific model trained in true end-to-end fashion on entire WSIs using slide-level labels. The model was trained and retrospectively validated on cohorts that represent the most extensive external validation of AI for prostate cancer grading to date, including approximately 100,000 core needle biopsies from 7,342 patients, collected across 15 clinical sites and trials spanning 11 countries (**Figure 1**). We address the challenges associated with validating AI models—namely, the absence of standardized study designs, pre-registered protocols, and appropriate external cohort sampling[37,38]—through rigorously following a pre-specified study protocol, which also includes detailed descriptions of the patient cohorts included in this study (**Supplementary Appendix 1**)[39]. Using varying amounts of task-specific training data, we apply two recent FMs to this task and present the first direct comparison between foundation models and supervised WSI-level end-to-end learning in the context of a clinically relevant, large-scale retrospective validation of AI diagnostics.

# Results

## Foundation models' Gleason grading is improved with task-specific training

Previous studies have advocated using FMs as feature encoders in a few-shot learning setup without extensive training data for a specific downstream task[8,9]. To investigate whether FMs require large task-specific training datasets to reach clinically optimal performance in the diagnosis and Gleason grading of prostate cancer, we trained models with increasing amounts of WSIs of prostate core needle biopsies ranging between 1%-100% of the full training set (**Extended Data Figure 1E**) collected in the Swedish STHLM3 trial[40] and at Stavanger University Hospital, Norway and Capio S:t Göran Hospital, Sweden, and evaluated the models on 12 international cohorts (**Figure 1**). The slides were sampled randomly and added cumulatively. We trained models based on the UNI (UFM)[9] and Virchow2 (VFM)[8,41] foundation models as frozen feature encoders alongside an in-house developed, fully end-to-end TS model relying on a trainable EfficientNet-V2S encoder (initialized with ImageNet weights)[42]. All three models used identical pre-processing for extracting tissue patches from the WSIs, the well-established attention-based multiple instance learning (ABMIL) architecture[20] for aggregating patch-level feature embeddings into WSI-level feature vectors, and additional classification layers for predicting the primary and secondary Gleason patterns for a WSI (**Figure 2A**). While having been trained in weakly supervised fashion using only slide-level Gleason score (GS) labels, the model architecture still allows for efficiently capturing and visually presenting the spatial distribution of distinct Gleason patterns (**Figure 2B-2D**).

The models were evaluated on held-out validation datasets representing both internal cohorts, i.e. data originating from the same laboratory and/or WSI scanner as training data but independent patients (n=14,808 WSIs), and fully external cohorts, i.e. data from different laboratories, scanners, and patients than the training data (n=10,801 WSIs). We quantified the models' concordance with pathologists in terms of Cohen's quadratically weighted kappa (QWK) for the International Society of Urological Pathology (ISUP) grade, a five-level grouping of Gleason scores also known as the WHO grade or grade group[43]. All models benefited from increased amounts of data and reached their maximal performance in internal (**Figure 2E**) and external (**Figure 2F**) validation when using 100% of the training set (n=55,798 WSIs). Considering the expected ability of FMs to function well even with limited task-specific training, it is surprising that with 1% of the training set (n=524 WSIs), the UFM and TS models performed comparably. In contrast, VFM met the expectation of performing relatively well already with 1% of the data but still exhibited markedly improved performance with more training. From a clinical standpoint, the improvements in QWK observed for UFM (0.672 to 0.915 on internal data, 0.637 to 0.856 on external data) and for VFM (0.862 to 0.911 on internal data, 0.821 to 0.849 on external data) are considerable and suggest FMs may still need relatively large task-specific training datasets for optimal performance.

In view of the assumed superior generalization performance of FMs, their advantage appears to be limited to the 1%-15% training data regime, with the TS model slightly outperforming UFM and VFM in terms of overall QWK on the external cohorts when more than 20% of the training data (n=10,326 WSIs) became available (**Figure 2F**). The differences in performance between 1% (**Figure 2G**) and 100% (**Figure 2H**) training data were to a large degree consistent across different cohorts. Corresponding results in terms of GS are provided in **Extended Data Figure 1** and full per-cohort results are presented in **Extended Data Figure 2**. In the remainder of the paper, we focus on the three models trained on 100% of the task-specific training data and additionally evaluate the feasibility of the FM-based few-shot learning approach relying on 1% of the data.

## Gleason grading by AI models is comparable to pathologists

We further analyzed the maximum performance of the models (using 100% training data) on a cohort-by-cohort basis. All models exhibited high concordance relative to each cohort's original reference standard on all tuning and internal validation datasets, but more pronounced variation could be observed across the external validation cohorts with QWK varying between 0.48-0.90 for GS and 0.62-0.90 for ISUP grade (**Figure 3A**). The TS model exhibited a minor but relatively consistent advantage over the FMs on the external data, outperforming both of them on 5/7 and 6/10 cohorts in terms of QWK for GS and ISUP, respectively. The sensitivity and specificity for cancer detection were fairly consistent across all cohorts for all models, ranging between 87% and 100%, but UFM and VFM tend to have somewhat higher sensitivity than the TS model at the expense of lower specificity (**Figure 3B**).

Measuring AI generalization across patient populations, laboratories, and scanners is complicated by inter-observer variability between site-specific reference standards provided by different local pathologists and by variations in the granularity of the reporting (i.e. grading per slide, per a set of slides from the same anatomical region, or per patient). To directly measure the models' capabilities to generalize across cohorts without the confounding effect of varying reference standards, we established a uniform reference standard based on a per-slide re-assessment of a randomly selected set of slides (stratified by ISUP grade according to the original grading) from each cohort by the lead study pathologist (L.E.). Measured against this uniform reference, all models exhibited QWK > 0.75 in ISUP grading consistently across all external cohorts with the exception of the FMs on the SPROB20 cohort (ISUP QWK 0.580 and 0.680 for UFM and VFM, respectively) (**Figure 3C**).

On the cohorts where the original grading was conducted per slide, a direct comparison between the original and uniform reference standards was possible. The concordance between the models and the uniform reference standard (mean QWK across all cohorts for ISUP) was higher at 0.883 (TS), 0.850 (UFM), and 0.870 (VFM) than the concordance between the original grading by local pathologists and the uniform reference standard at 0.801. The lead study pathologist is an experienced uropathology specialist and has been shown to be highly concordant with other specialists in earlier studies[6,26,44], but relying on a single reader as the sole reference is still problematic in view of the subjective nature of Gleason grading. To place AI grading variability in the context of inter-observer variability between pathologists, we evaluated pairwise grading concordance on subsets of the STHLM3 internal validation set (ImageBase[45]) and the RUMC external validation set[26], as well as on the full UKK and WNS external validation sets[46], all graded independently by panels of pathologists and the AI models (**Figure 4**). The mean pairwise ISUP QWK between pathologists ranged from 0.67 to 0.93 in the four cohorts. The corresponding values achieved by the three models were comparable to the pathologists, with the TS model showing a minimal advantage over the FMs in all four cohorts.

We additionally analyzed model performance in the following subgroups: malignant samples only (**Extended Data Figure 3**), across patient age groups (**Extended Data Figure 4A**), and among patients treated for benign prostatic hyperplasia with 5-α-reductase inhibitors or alpha-blockers (which has been hypothesized to influence prostate morphology[47,48]) (**Extended Data Figure 4B**). Results in terms of Cohen's linearly weighted kappa (LWK) for GS and ISUP are shown in **Extended Data Table 1** (original reference standard) and **Extended Data Table 2** (uniform reference standard). Results in terms of Area Under the Receiver Operating Characteristic Curve (AUROC) are further shown in **Extended Data Figure 5**.

## Diagnosing difficult and rare cases requires task-specific training

After confirming that the AI models achieve overall pathologist-level performance across diverse patient cohorts, we focused on difficult and rare cases. The most severe consequences of misgrading are associated with high-grade cancers, leading to under- or overtreatment of patients, whereas errors between benign and low-grade cancers have less clinical significance due to the indolence of ISUP grade 1 prostate cancer[49]. We adopted the following pre-specified definition for clinically significant errors: a cancer of at least ISUP grade 2 predicted as benign or a benign sample predicted as a cancer of at least ISUP grade 2. Following this definition, we quantified significant errors committed by the models across all validation cohorts where the original reference standard was provided per slide (n=11,406 WSIs). When trained with 100% of training data, all models arrived at a similar rate of errors (110 WSIs, 0.96% for UFM and VFM, 111 WSIs, 0.97% for TS). Of the slides with errors, 49 were common to all models. Importantly, extensive task-specific training considerably decreased the number of clinically significant errors committed by the FMs trained with 1% vs. 100% of the data: from 139 to 110 WSIs (-20.9%) for UFM, and from 136 to 110 (-19.1%) for VFM. For the TS model, this result (from 366 to 111 WSIs, -69.7% decrease) was naturally expected.

To further evaluate the nature of the significant errors, the lead study pathologist re-graded all the slides (n=111) with significant errors committed by one of the models (TS), blinded to the original reference standard and the AI grading. Compared to this re-assessment, errors in 63/111 slides (57%) were resolved, meaning they could be attributed to database issues (e.g. mistyped information in the reference standard, mixed-up slide identifiers, and WSI scanning issues in cases where the original grading was done through microscope, etc.). Out of the 49 slides common to the three models, 42 fell into this category. Assuming the remaining 48 slides represent true AI errors, the TS model committed clinically significant errors in 0.42% of the validation slides (see **Extended Data Figure 6** for GS and ISUP QWK relative to the refined reference standard). These slides were characterized by the presence of minimal lesions, crush artifacts, overstained sections, out of focus regions, and unusual morphologies. Furthermore, many of these slides would require immunohistochemistry (IHC) in clinical routine to confirm primary prostatic adenocarcinoma.

We additionally evaluated the models on biopsies from three validation cohorts (AQ, KUH-2, STHLM3) representing challenging morphologies (n=304 WSIs), such as benign mimickers of prostate cancer and rare prostate cancer subtypes, which are difficult to diagnose but will inevitably be encountered by AI models in clinical use[50] (**Figure 5A**). Model performance was quantified in terms of sensitivity (rare malignant morphologies) and specificity (benign mimickers), as Gleason grading is not applicable to the majority of these morphologies. All models exhibited high sensitivity (92-100%) for detecting cancers with rare morphologies (**Figure 5B**) but struggled with a considerable number of false positives when assessing unusual benign tissue (**Figure 5C**). The resulting specificity varied dramatically depending on the model and the amount of task-specific training. The specificity of the TS model pooled across the cohorts was 0.690. For UFM and VFM, the pooled specificity improved from 0.261 to 0.584 and from 0.632 to 0.710, respectively, when increasing the fraction of training data used from 1% to 100%.

# Cross-scanner reproducibility of AI Gleason grading varies with task-specific training

Given that our tuning and validation cohorts include data collected with 14 different individual scanner instruments, representing 9 models from 5 vendors (**Figure 6A**), the overall results provide an indirect evaluation of AI generalization across scanners. However, differences in patient demographics and sample preparation contribute to the overall variation. To directly measure reproducibility across scanners, we quantified the models' cross-scanner concordance in terms of QWK for ISUP grading between subsets of slides re-scanned with multiple scanner models in the MUL (n=481 slides, 2 scanners) and STHLM3 (n=48 slides, 5 scanners) cohorts (**Figure 6B-C**).

In line with the other experiments, UNI trained with 1% of training data exhibited very low cross-scanner concordance but improved considerably with the full training dataset (mean pairwise QWK between scanners improved from 0.622 to 0.937 for STHLM3 and from 0.577 to 0.908 for MUL). In contrast, VFM exhibited high cross-scanner concordance overall but the effect of additional training data was cohort-dependent (mean pairwise QWK between scanners degraded from 0.952 to 0.904 for STHLM3 but improved from 0.870 to 0.912 for MUL). The TS model (trained fully with 100% task-specific data) exhibited mean cross-scanner pairwise QWK of 0.963 (STHLM3) and 0.918 (MUL) and outperformed both FMs on 9 of the 11 scanner pairs. Corresponding results in terms of cross-scanner QWK for GS are presented in **Extended Data Figure 7**.

# Foundation models consume up to 35x more energy than a task-specific model

Foundation models typically rely on more complex neural network architectures (e.g. Vision Transformers (ViT)[51]) than task-specific models: UFM (ViT-L/16) and VFM (ViT-H/14) have approximately 300 million and 632 million parameters, respectively, compared to the approximately 22 million parameters of TS (EfficientNet-V2S architecture)[52]. The size of a model influences its energy consumption and runtime, which has important implications for the financial and environmental sustainability of clinically deployed AI solutions.

To compare the relative efficiency of the three models in this study, we recorded the total amount of power consumed by the GPU to produce the Gleason grading predictions for our tuning set (n=801 slides). The models consumed 0.51 kWh (TS), 5.40 kWh (UFM) and 17.70 kWh (VFM), with runtimes of 2.50 GPUh, 15.25 GPUh, and 45.57 GPUh, respectively. The models spent 0.63 Wh (TS), 6.74 Wh (UFM), and 22.09 Wh (VFM) per prostate biopsy.

# Discussion

Previous studies on FMs have demonstrated their ability to operate effectively across a variety of tasks by relying on few-shot learning with limited training data. This is particularly advantageous in the medical domain, where labeled data are often scarce. At the same time, medical applications involving high-stakes decisions pose stringent requirements for AI accuracy[53]. Regulatory frameworks may need to be adapted in view of generalist AI models[54], but it will remain essential that all AI models be thoroughly evaluated with clinically relevant performance metrics before being deployed for a specific task. Here, we present the most comprehensive validation study to date on AI-based assessment of biopsies for diagnosis and Gleason grading of prostate cancer, relying on both FMs and state-of-the-art end-to-end learning. Our first key result is that to reach AI performance optimal for clinical deployment, even FMs require substantial amounts of task-specific training data. Moreover, with sufficient training data available, FMs do not possess any intrinsic generalization advantage over an end-to-end trained task-specific model. On the contrary, while all models exhibited nearly identical performance in internal validation, the overall performance of the FMs was slightly lower than the TS model on fully external validation cohorts. While the value proposition of FMs is clear for applications where data are scarce (e.g. rare diseases), these results suggest against assuming universal superiority of FMs over end-to-end models and advocate for caution with clinical deployment of FMs relying on limited task-specific training, which may risk committing clinically critical errors.

Sensitivity of AI models to variability introduced by different WSI scanners is a well-known problem in computational pathology, greatly hampering the clinical applicability of AI[55–59]. Despite hopes of FMs possessing inherent tolerance for scanner variation and solving the generalization problem once and for all, limited evidence suggests that the composition of FM pretraining data does influence downstream task performance on different tissue types[60] and that FMs are not immune to batch effects and biases present during the self-supervised pre-training[59,61–63]. Our second key result, handling the models' cross-scanner concordance, supports these earlier findings, demonstrating that UNI in particular suffered from poor diagnostic reproducibility when the same slides were imaged using different scanners. Interestingly, the cross-scanner concordance of the model was greatly improved with extensive task-specific training. This seems to suggest that the feature representation learned by the FM captures scanner-specific data characteristics, and a considerable number of task-specific examples are needed to adapt the auxiliary aggregation and classification layers to learn to correctly disregard these biases. Virchow2 exhibited more consistent performance even with limited task-specific training, but the highest cross-scanner concordance was still obtained by the end-to-end TS model.

Our third key result handles the energy consumption of AI in pathology. One thus far largely neglected consequence of increasing model complexity is the spiraling energy consumption. While the absolute power consumption per sample depends on various factors such as the type and efficiency of hardware (also including other components than the GPU) and the amount of tissue per slide, our experiments provide an estimate of the relative energy demands of analyzing prostate biopsies using FMs in comparison to a task-specific model (approx. 11x for UFM, 35x for VFM). This result is in line with the earlier reported directly proportional relationship between the number of model parameters and the resulting $CO_2$ emissions[64]. Increased demands for computing are also associated with growing use of freshwater resources for cooling data centers[65]. Amidst the global climate crisis, the increased environmental toll of FMs and other large models has to be weighed against the benefits they can provide in a given task compared to more streamlined, task-specific models that consume a fraction of the energy. Besides environmental sustainability, laboratories also need to consider the operational costs of AI computing from a financial perspective. In the case of AI-based prostate cancer diagnosis and grading, our results do not support the use of FMs in this task in place of simpler, task-specific models, provided that extensive training data are available and the aim is clinical-level diagnostic performance.

Our fourth key result is the confirmation of earlier findings[6,7,26,27] indicating that AI can diagnose and Gleason grade prostate cancer comparably to experienced pathologists, marking the most extensive validation of AI for this task to date. In this study, FM-based models did not demonstrate any clear advantages over the end-to-end trained TS model. However, when trained with extensive task-specific data, all the models in this study reached performance comparable to each other, to pathologists, and to earlier studies. It seems likely that performance in Gleason grading has reached a plateau that probably cannot be considerably improved on with further advances in AI. However, the picture may be markedly different in other tasks. For example, for direct prognostication of patient outcomes or treatment response from tissue morphology, novel (potentially subtle) patterns need to be discovered, which has led some authors to advocate for further research into end-to-end approaches[30]. Despite varying definitions for "end-to-end" training, current state-of-the-art models for prognosticating patient outcomes[66] or molecular biomarkers[67] from WSIs do not represent true end-to-end learning, where all parameters of the model are optimized jointly during training. Abandoning this key ingredient of the first wave of success in AI pathology precludes supervised data-driven discovery of patterns that are most informative for the task at hand.

Our study has some limitations. Firstly, while the validation cohorts were fully external to the evaluated models and captured a broad spectrum of clinical sites, laboratories and scanners, they represent populations with primarily Caucasian ancestry. Second, all scanners included in the cross-scanner analysis except for the Grundium device were present in the models' training data. Even more drastic cross-scanner variation in AI performance is thus likely to be seen if a similar evaluation is repeated on additional, unseen scanners. We plan to address these limitations in future validation studies representing more diversity in terms of patient ethnicity and scanner models.

Despite the concerns and shortcomings raised by us and others, the value of FMs is clear for accelerating research and development, and as a solution when labeled data are scarce. In addition, when moving from more fine-grained tasks (e.g. detection of nuclei) towards larger scales (e.g. prognostication of patient survival), the number of labeled data points (e.g. patients vs. pixels) shrinks, which complicates fully supervised end-to-end learning. Addressing complex multimodal applications may also be infeasible without FMs that are capable of integrating visual data with textual[10] or molecular information[68]. However, in its current form, the accelerating proliferation of FMs resembles the era of feature engineering, where machine learning practitioners solving a given task were faced with a semi-empirical *ad hoc* process of picking their favorites from a vast and expanding pool of various morphological descriptors. Our results clearly demonstrate that for a given task, not all FMs are created equal, and evidence is mounting on the existence of different biases encoded in their feature representations.

We believe future research will lead to optimized, data-driven ways of selecting, fusing, and pruning FMs[60] to dissect the relevant components of their embedded representations for a given task, while at the same time mitigating their growing environmental footprint. Another exciting opportunity in this direction are AI agents capable of autonomously selecting and using the best tools for each task[69]. For clinical applicability, stringent validation studies will still be needed to ascertain that biases and batch effects arising e.g. from the variability between WSI scanners are controlled for. Despite their unquestionable potential to revolutionize medical AI, without comprehensive studies addressing these points, FMs risk being prematurely positioned as a universal replacement for established approaches and discouraging the continued development of end-to-end learning for computational pathology.

# Ethical considerations

This study included data gathered in one or more collection rounds at participating international sites between 2012 and 2024. All datasets were de-identified at their respective sites and subsequently transferred to Karolinska Institutet in an anonymized format. This study complies with the Helsinki Declaration. The patient sample collection was approved by the Stockholm Regional Ethics Committee (permits 2012/572-31/1, 2012/438-31/3, and 2018/845-32), the Swedish Ethical Review Authority (permit 2019-05220), and the Regional Committee for Medical and Health Research Ethics in Western Norway (permits REC/Vest 80924, REK 2017/71). Informed consent was obtained from patients in the Swedish dataset and was waived for other data cohorts due to the use of de-identified prostate specimens in a retrospective setting. Patient involvement in this study was supported by the Swedish Prostate Cancer Society.

# Acknowledgments

A.B. received a grant from the Health Faculty at the University of Stavanger, Norway. B.G.P and K.D.S received funding from Innovation Fund Denmark and Nordforsk (Grant no. 8114-00014B) for the Danish branch of the NordCaP project. K.H. received funding from Business Finland (BF) for the Finnish branch of the NordCaP project. P.R. received funding from the Research Council of Finland (Grant no. 341967) and the Cancer Foundation Finland. M.E. received funding from the Swedish Research Council, Swedish Cancer Society, Swedish Prostate Cancer Society, Nordic Cancer Union, Karolinska Institutet, and Region Stockholm. K.K. received funding from the SciLifeLab & Wallenberg Data Driven Life Science Program (KAW 2024.0159), David and Astrid Hägelen Foundation, Instrumentarium Science Foundation, KAUTE Foundation, Karolinska Institute Research Foundation, Orion Research Foundation and Oskar Huttunen Foundation. We want to thank Carin Cavalli-Björkman, Astrid Björklund and Britt-Marie Hune for assistance with scanning and database support. We would also like to thank Simone Weiss for assistance with scanning in Aarhus, and Silja Kavlie Fykse and Desmond Mfua Abono for scanning in Stavanger. We would like to acknowledge the patients who participated in the STHLM3 diagnostic study and the OncoWatch and NordCaP projects and contributed the clinical information that made this study possible. High-performance computing was supported by the National Academic Infrastructure for Supercomputing in Sweden (NAISS) and the Swedish National Infrastructure for Computing (SNIC) at C3SE partially funded by the Swedish Research Council through grant agreement no. 2022-06725 and no. 2018-05973, by the supercomputing resource Berzelius provided by the National Supercomputer Centre at Linköping University and the Knut and Alice Wallenberg Foundation, and by CSC - IT Center for Science, Finland.

# Competing interests

N.M., L.E., K.K. and M.E. are shareholders of Clinsight AB.

# Figures and Tables

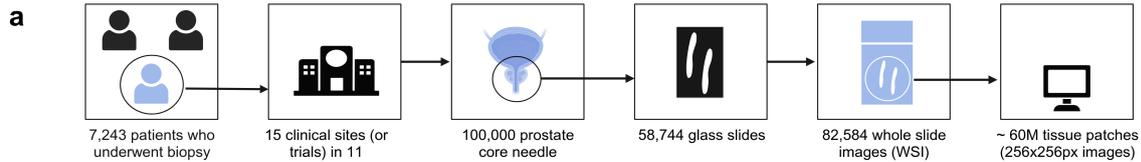

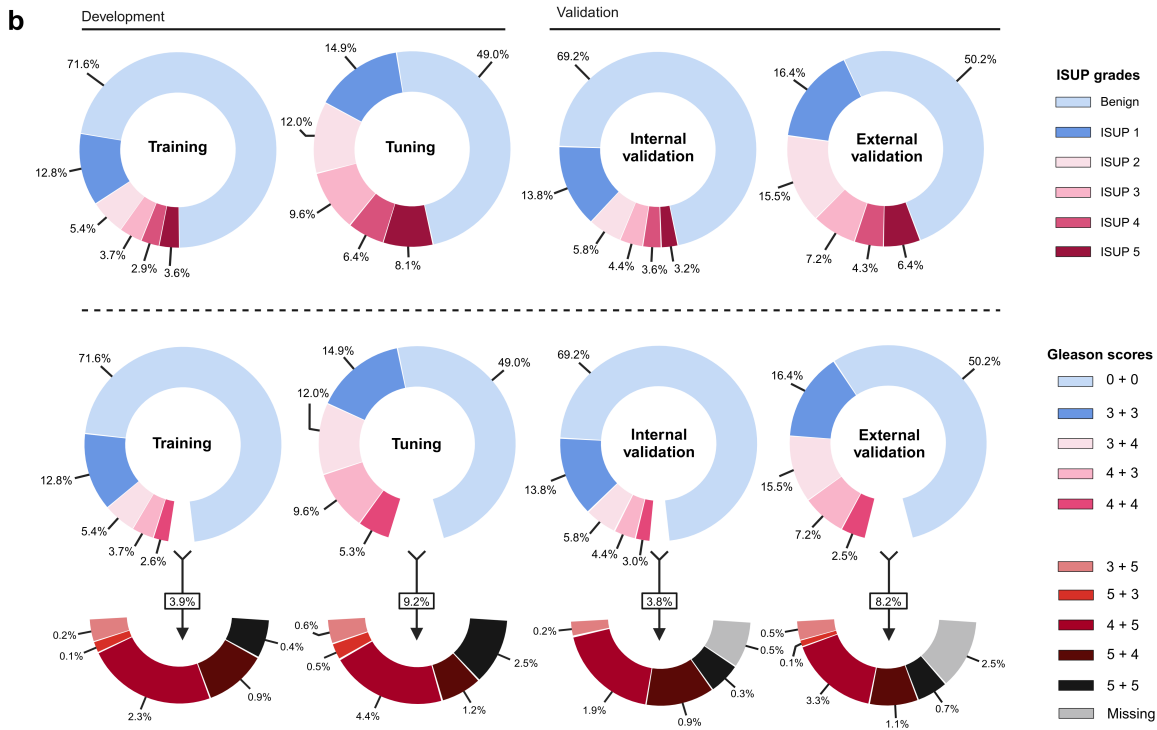

**Figure 1: Overview of patient cohorts. (a)** In total, 7,243 patients who underwent prostate biopsy were included in the study, resulting in approximately 100,000 biopsy cores from 15 clinical sites in 11 countries. The dataset contains 58,744 glass slides, digitized into 82,584 WSIs and tiled into approximately 60 million tissue patches (256 x 256 pixel images). The number of WSIs differs from the number of glass slides due to repeated scanning of subsets of the slides on multiple scanners. The patients were divided into development (training and tuning) and validation (internal and external) partitions, such that there was no overlap between the patients, clinical sites or scanners included in the external validation and the development partitions. See **Supplementary Appendix 1** for detailed patient characteristics, data collection processes, reference standard protocols and CONSORT diagrams for each cohort. **(b)** The ISUP grade and Gleason score distributions of the glass slides included in each partition. For the SPROB20, UKK and WNS cohorts, only ISUP grades were available. **(c)** Geographical distribution of the included clinical sites and the numbers of patients, glass slides and WSIs per cohort. *The approximate total number of biopsy cores was estimated considering the total number of glass slides and the typical number of cores per glass slide for each cohort. ISUP=International Society of Urological Pathology, WSI = Whole slide image.[1]

---

[1] Blilie, A. (2025) https://BioRender.com/e66o459

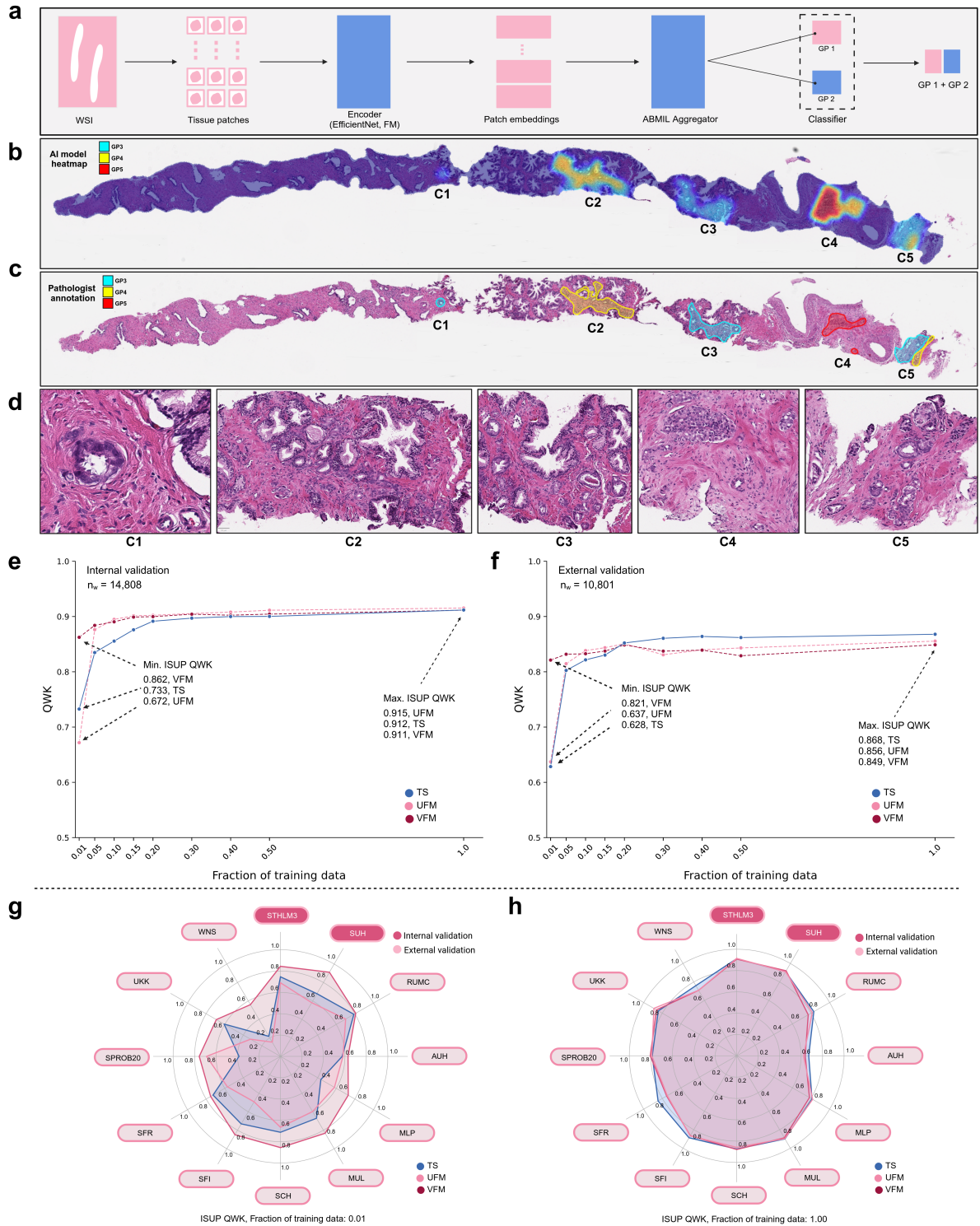

**Figure 2: Model design and overall performance as a function of increased task-specific training. (a)** A schematic of the model design featuring attention-based aggregation of patch-level feature embeddings, followed by a multitask classification layer for predicting the primary and secondary Gleason patterns of the entire WSI. The model is trainable with either foundation model based patch encoders or in a full end-to-end fashion with an EfficientNet encoder. **(b)** Heatmap showing predictions per patch for an example WSI demonstrates that weakly supervised training with WSI-level Gleason score labels has still allowed the model to learn to recognize individual Gleason patterns on the patch level. **(c)** A pathologist's annotations of Gleason patterns (conducted independently and blinded to the AI models). **(d)** Zoomed-in views of the annotated cancer foci (C1-

C5). **(e-f)** Overall ISUP grading concordance between the models (TS, UFM, VFM) and the original reference standard measured in terms of QWK as a function of increasing fractions of downstream task-specific training data used, evaluated on the pooled internal **(e)** and external **(f)** validation cohorts ($n_w$ indicates the number of validation WSIs). **(g-h)** Per-cohort ISUP grading QWK for the models trained with 1% of the training data (n=524 WSIs) (**g**) and 100% of the training data (n=55,789 WSIs) (**h**). ABMIL=Attention-based multiple instance learning, FM=Foundation model, GP=Gleason pattern, ISUP=International Society of Urological Pathology, QWK=Quadratically weighted Cohen's kappa, TS=Task-specific model, UFM=UNI foundation model, VFM=Virchow2 foundation model, WSI=Whole slide image.[2]

---

2 Mulliqi, N. (2025) https://BioRender.com/j00e590

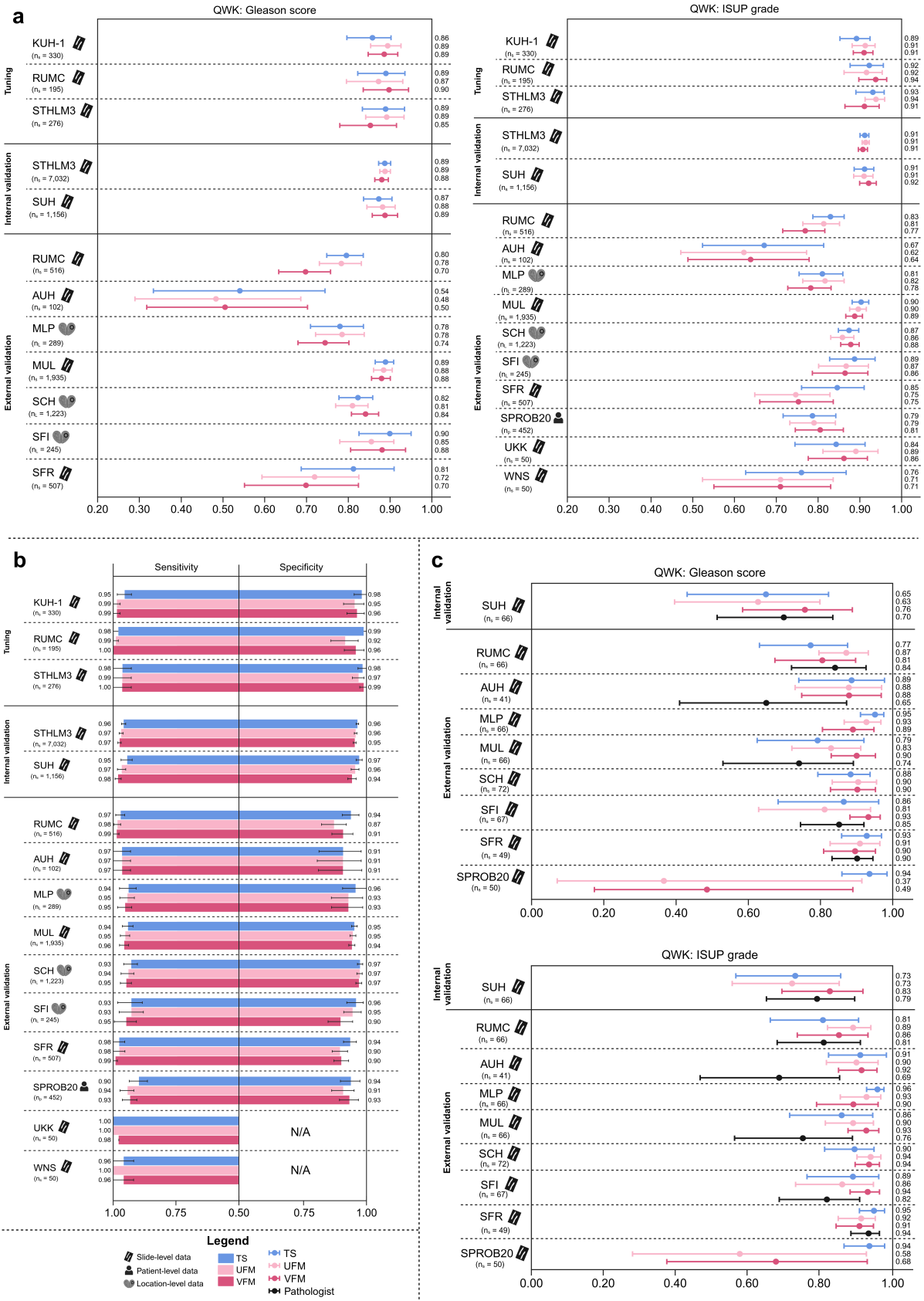

**Figure 3: Gleason grading performance of AI models across international validation cohorts. (a)** Gleason grading concordance between the models (TS, UFM, VFM; all trained with 100% of the task-specific data) and the original reference standards defined by cohort-specific local pathologists,

measured in terms of QWK for Gleason score (left; reference Gleason scores not available for SPROB20, UKK, WNS) and for ISUP grade (right). (**b**) Cancer detection performance of the models measured by sensitivity and specificity (omitted for the UKK and WNS cohorts, which only include malignant cases) relative to the original reference standards. (**c**) The Gleason grading concordance of the models and the local pathologists relative to a uniform slide-level reference standard by the lead study pathologist, measured in terms of QWK for Gleason score (top) and ISUP grade (bottom). For the MLP, SCH and SPROB20 cohorts, the original reference standard was not reported on slide-level and the comparison between the local pathologists and the lead study pathologist is omitted. The granularity of the reporting of the original reference standards varied by cohort. The $n_s$, $n_l$ and $n_p$ indicate the number of glass slides, anatomical locations or patients, respectively, included in each analysis. The values indicated by the plots for QWK, sensitivity and specificity represent point estimates on the full cohorts, while the whiskers and error bars represent 95% confidence intervals estimated by bootstrapping. ISUP=International Society of Urological Pathology, QWK=Quadratically weighted Cohen's kappa, TS=Task-specific model, UFM=UNI foundation model, VFM=Virchow2 foundation model, WSI=Whole slide image.[3]

---

[3] Blilie, A. (2025) https://BioRender.com/w12k266

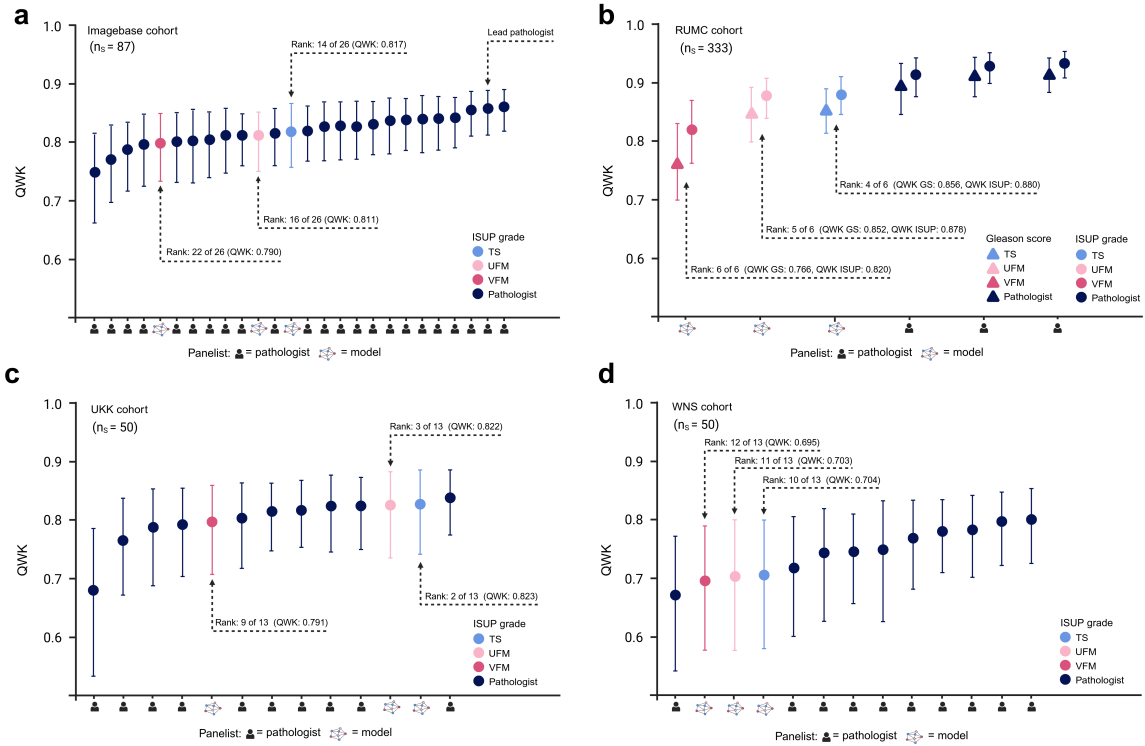

**Figure 4: Inter-observer variability among panels of pathologists and AI models. (a-d)** Mean pairwise concordance between the models (TS, UFM, VFM; all trained with 100% of the task-specific data) and pathologists, compared to mean pairwise concordance between pathologists, measured in terms of QWK for Gleason score (only available for the RUMC cohort) and for ISUP grade, and evaluated in **(a)** ImageBase (part of STHLM3 internal validation cohort) and in the **(b)** RUMC, **(c)** UKK, and **(d)** WNS external validation cohorts. The panelists (pathologists and AI models) are ranked according to their mean pairwise QWK, but the AI models were not included in the calculation of the pathologists' or other models' mean QWK to avoid biasing the panel's consensus towards the models (i.e. for each pathologist or model, the mean QWK against all other pathologists is shown). The number of glass slides included in each analysis is indicated by $n_s$. The dots indicate point estimates and the error bars represent 95% confidence intervals estimated by bootstrapping. ISUP=International Society of Urological Pathology, QWK=Quadratically weighted Cohen's kappa, TS=Task-specific model, UFM=UNI foundation model, VFM=Virchow2 foundation model.[4]

---

[4] Mulliqi, N. (2025) https://BioRender.com/v76j626

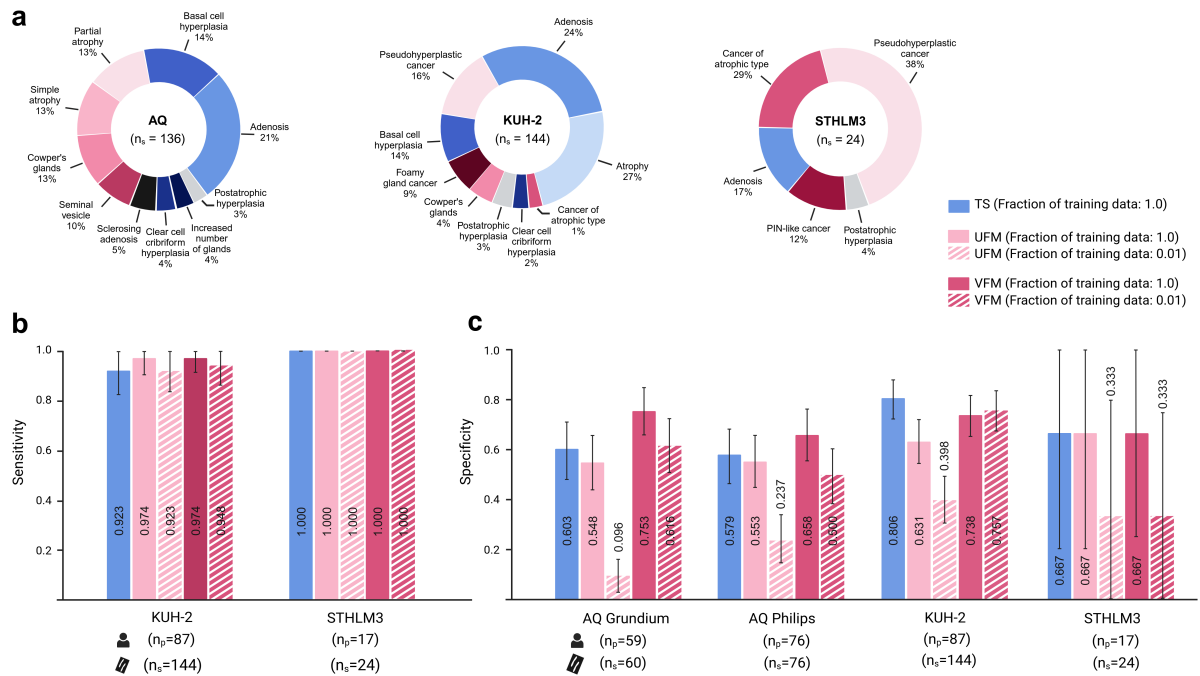

**Figure 5: Diagnostic performance of AI models on challenging morphologies. (a)** The distribution of challenging morphological subtypes, including benign (e.g., atrophy, adenosis, Cowper's glands) and malignant (e.g., pseudohyperplastic cancer, PIN-like cancer) types, in the AQ, KUH-2 and STHLM3 cohorts. **(b)** Diagnostic sensitivity and **(c)** specificity for the models (TS, UFM, VFM; trained with 1% or 100% of the task-specific data), relative to the reference standard by the lead study pathologist. Sensitivity is omitted for the AQ cohort, which only contains benign mimickers of cancer. Subsets of the AQ cohort were scanned on different scanners (Grundium, fully external; Philips, included in development cohorts). The $n_s$ and $n_p$ indicate the number of glass slides and patients, respectively, included in each analysis. The bars indicate point estimates and the error bars represent 95% confidence intervals estimated by bootstrapping. TS=Task-specific model, UFM=UNI foundation model, VFM=Virchow2 foundation model.[5]

---

[5] Blilie, A. (2025) https://BioRender.com/m31o854

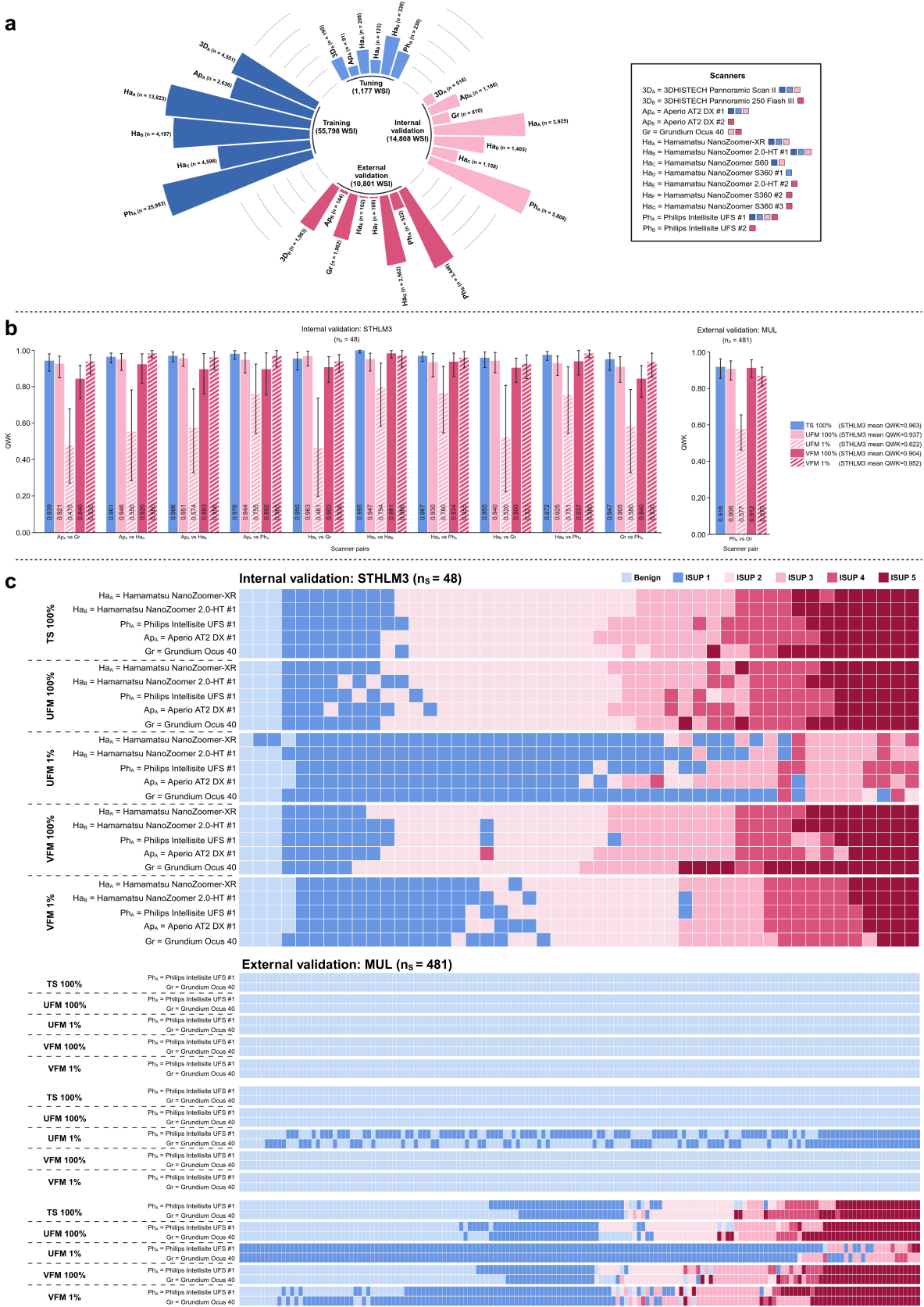

**Figure 6: Cross-scanner performance of AI models. (a)** The distributions of individual slide scanners used to digitize the study cohorts. The scanners used for digitizing external cohorts for the primary validation (see **Figures 2-4**) were not involved in the development, tuning or internal validation cohorts. The Philips Intellisite UFS #1 scanner, used to collect training and internal validation data, was also used to re-scan a subset of the MUL external validation cohort, but these WSIs were used only for the cross-scanner analysis presented here. This scanner was also used to digitize a subset of the AQ cohort for the analysis on challenging morphologies (see **Figure 5**). **(b)** The cross-scanner concordance of each model (TS, UFM, VFM; trained with 1% or 100% of the task-specific data) measured in terms of QWK for ISUP grade, comparing each model's grading for the same slides digitized on pairs of different scanners (incl. 5 scanners for the STHLM3 cohort and 2 scanners for the MUL cohort). In addition, the mean of the 10 pairwise comparisons on the STHLM3 cohort is indicated (see legend). **(c)** Heatmaps showing the ISUP grade predictions by each model across glass slides (columns) digitized on different scanners (rows). For each model, the slides are sorted for visual readability based on the model's average ISUP prediction across scanners i.e. the column ordering is different between models. The number of glass slides included from the internal (STHLM3) and external (MUL) validation cohorts is indicated by $n_s$. The bars indicate point estimates and the error bars represent 95% confidence intervals estimated by bootstrapping. ISUP=International Society of Urological Pathology, QWK=Quadratically weighted Cohen's kappa, TS=Task-specific model, UFM=UNI foundation model, VFM=Virchow2 foundation model.

# Extended Data Figures

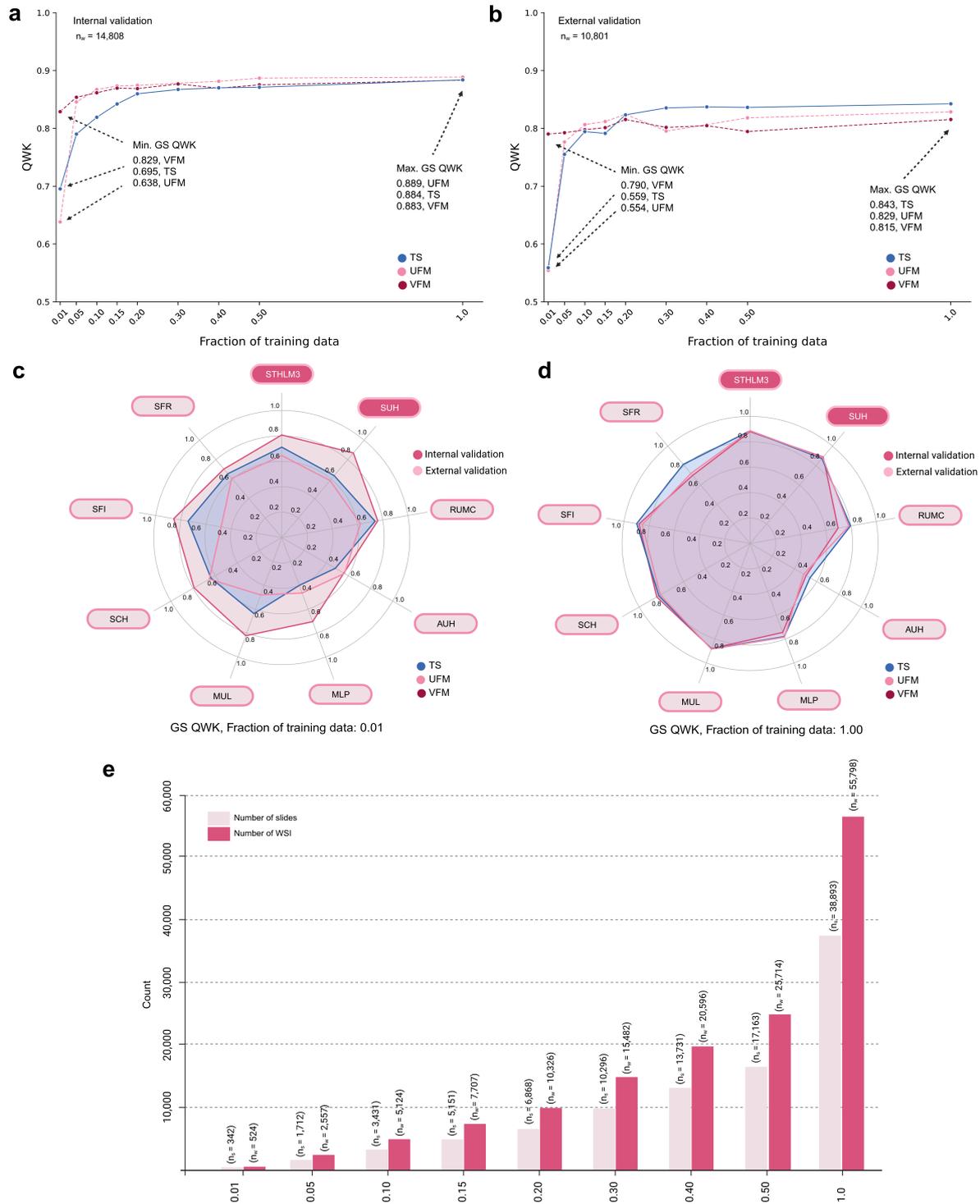

**Extended Data Figure 1: Overall Gleason scoring performance of AI models as a function of increased task-specific training. (a-b)** Overall Gleason score concordance between the models (TS, UFM, VFM) and the original reference standard measured in terms of QWK as a function of increasing fractions of downstream task-specific training data used, evaluated on the pooled internal

**(a)** and external **(b)** validation cohorts. **(c-d)** Per-cohort Gleason scoring QWK for the models trained with 1% of the training data (n=524 WSIs) (**c**) and 100% of the training data (n=55,789 WSIs) (**d**). **(e)** Number of slides and WSIs corresponding to each training data fraction. GS=Gleason score, QWK=Quadratically weighted Cohen's kappa, TS=Task-specific model, UFM=UNI foundation model, VFM=Virchow2 foundation model, WSI=Whole slide image.

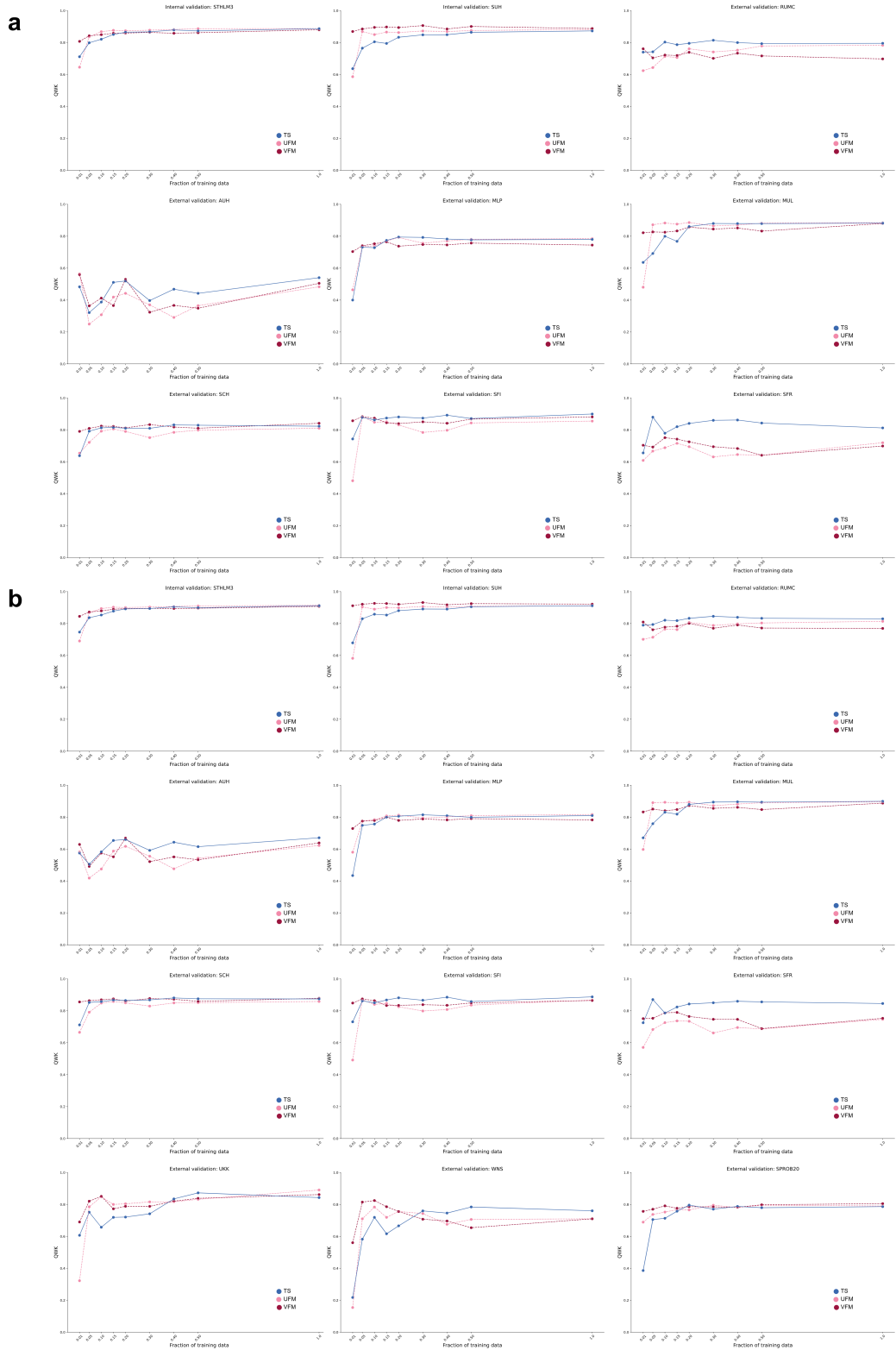

**Extended Data Figure 2: Gleason grading performance of AI models across validation cohorts as a function of increased task-specific training. (a-b)** Per-cohort concordance between the models (TS, UFM, VFM) and the original reference standard measured in terms of QWK as a function of increasing fractions of downstream task-specific training data for Gleason score (**a**) and ISUP grade (**b**). ISUP=International Society of Urological Pathology, QWK=Quadratically weighted Cohen's kappa, TS=Task-specific model, UFM=UNI foundation model, VFM=Virchow2 foundation model.

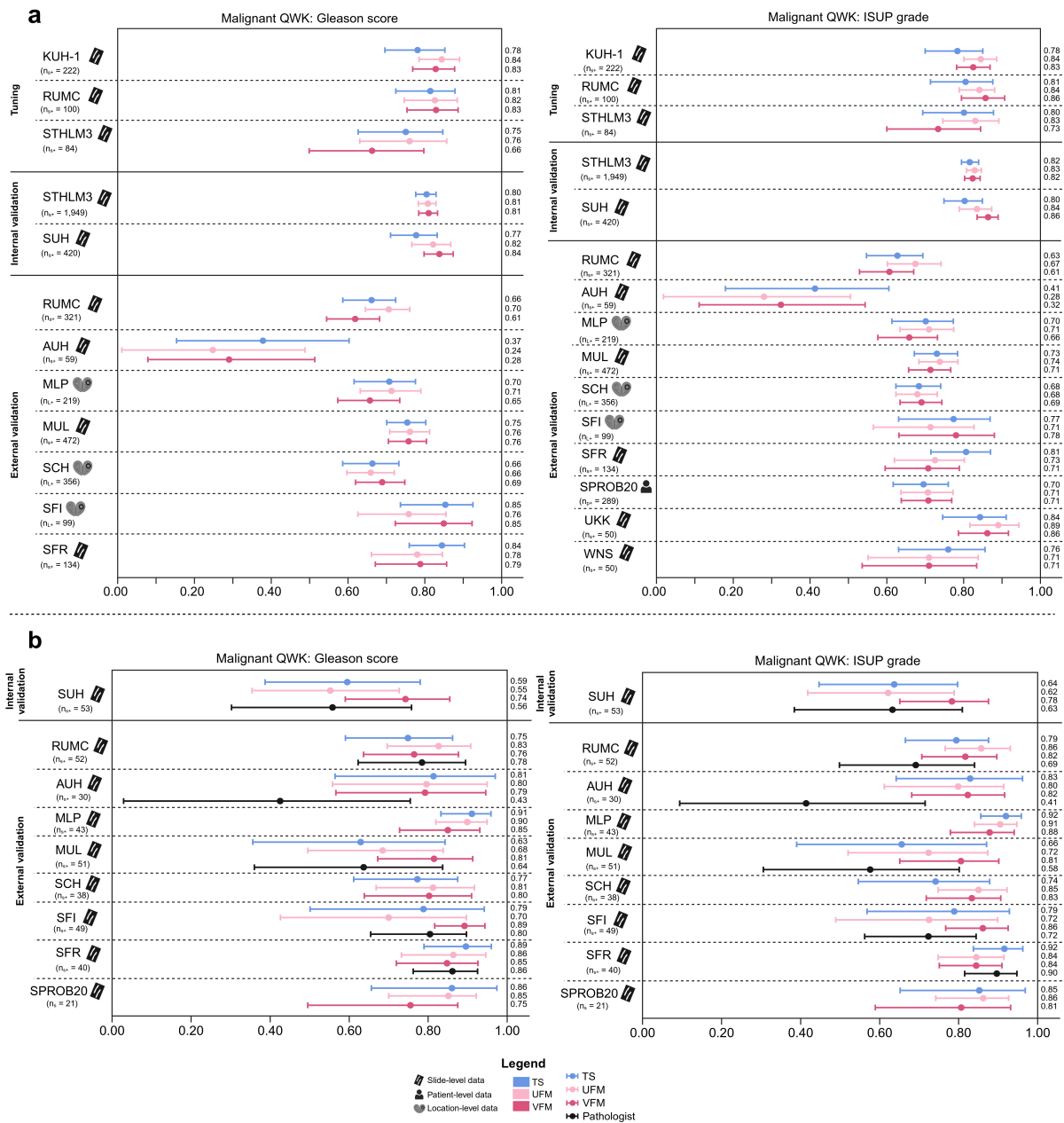

**Extended Data Figure 3. Gleason grading performance of AI models across international validation cohorts excluding benign samples.** (**a**) Gleason grading concordance between the models (TS, UFM, VFM; all trained with 100% of the task-specific data) and the original reference standards defined by cohort-specific local pathologists, measured in terms of QWK for Gleason score (left; reference Gleason scores not available for SPROB20, UKK, WNS) and for ISUP grade (right). (**b**) Gleason grading concordance of the models and the local pathologists relative to a uniform slide-level reference standard by the lead study pathologist, measured in terms of QWK for Gleason score (left) and ISUP grade (right). For the MLP, SCH and SPROB20 cohorts, the original reference standard was not reported on slide-level and the comparison between the local pathologists and the lead study pathologist is omitted. The granularity of the reporting of the original reference standards varied by cohort. The $n_s$, $n_l$ and $n_p$ indicate the number of glass slides, anatomical locations or patients, respectively, included in each analysis. Cases diagnosed as benign in the original reference standard were excluded from these analyses. The values indicated by the plots represent point estimates on the full cohorts, while the whiskers and error bars represent 95% confidence intervals estimated by

bootstrapping. ISUP=International Society of Urological Pathology, QWK=Quadratically weighted Cohen's kappa, TS=Task-specific model, UFM=UNI foundation model, VFM=Virchow2 foundation model, WSI=Whole slide image.[6]

---

[6] Blilie, A. (2025) https://BioRender.com/w87s307

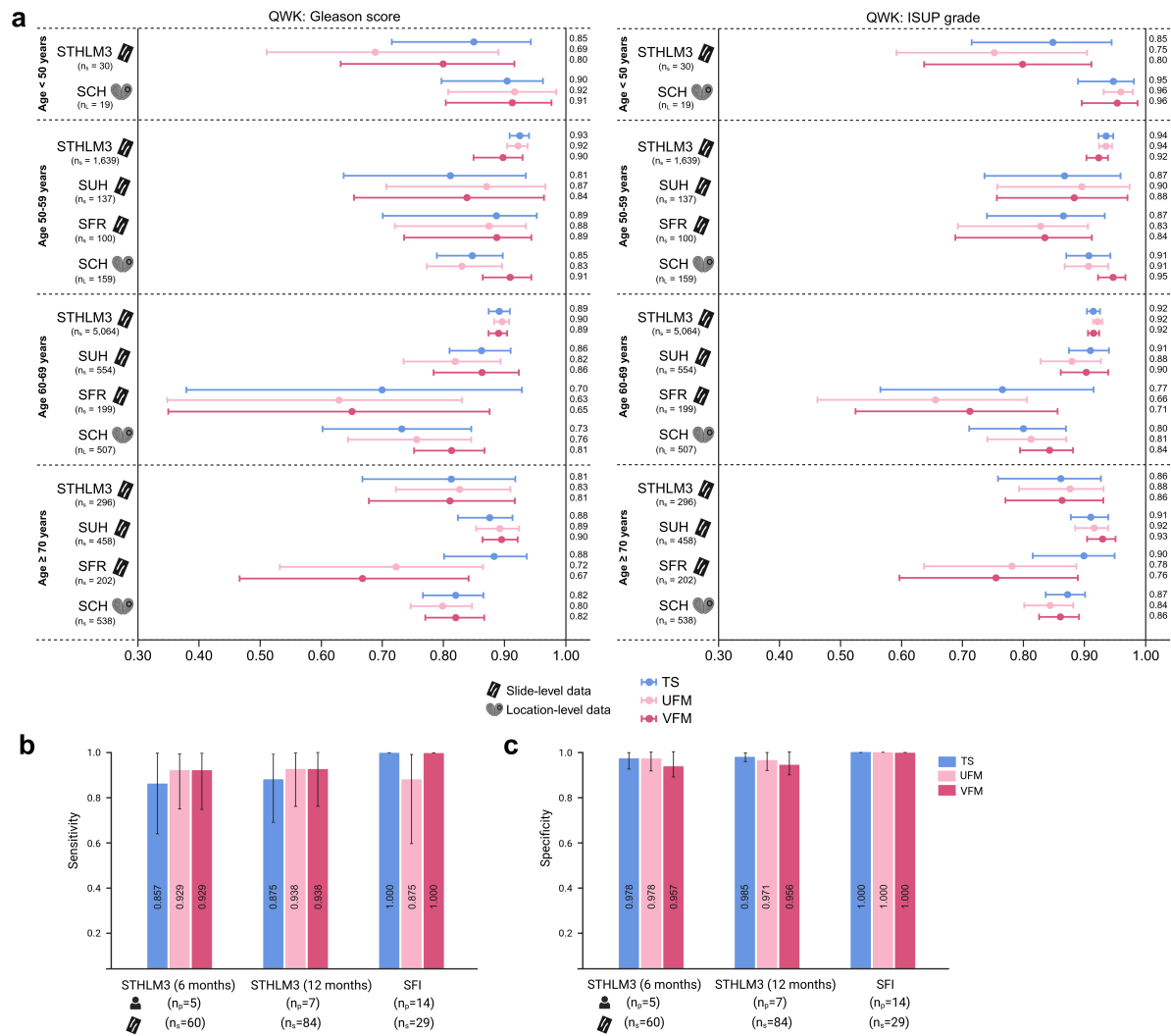

**Extended Data Figure 4: Grading performance of AI models across patient age groups and for patients treated for benign prostatic hyperplasia. (a)** Gleason grading concordance between the models (TS, UFM, VFM; all trained with 100% of the task-specific data) and the original reference standards defined by cohort-specific local pathologists, measured in terms of QWK for Gleason score (left) and for ISUP grade (right) across patient age groups (for cohorts with age information available). The SUH and SFR cohorts had an insufficient number of patients < 50 years to measure the concordance. **(b-c)** Cancer detection performance measured by sensitivity (**b**) and specificity (**c**) against original reference standards on patients treated for benign prostatic hyperplasia with 5-ARI or alpha-blockers prior to biopsy. For the STHLM3 cohort, we included patients who were treated for at least 3 months and who were biopsied within 6 or 12 months after treatment ended. The time of the biopsy in relation to treatment was not reported for SFI patients and we therefore included all patients with a mention of 5-ARI or alpha-blocker treatment in their pathology reports. The $n_s$ and $n_p$ indicate the number of glass slides and patients, respectively, included in each analysis. The values indicated by the plots for QWK, sensitivity and specificity represent point estimates on the full cohorts, while the whiskers and error bars represent 95% confidence intervals estimated by bootstrapping. 5-ARI=5-α-reductase inhibitor, ISUP=International Society of Urological Pathology, QWK=Quadratically

weighted Cohen's kappa, TS=Task-specific model, UFM=UNI foundation model, VFM=Virchow2 foundation model.[7]

---

[7] Blilie, A. (2025) https://BioRender.com/v83h860

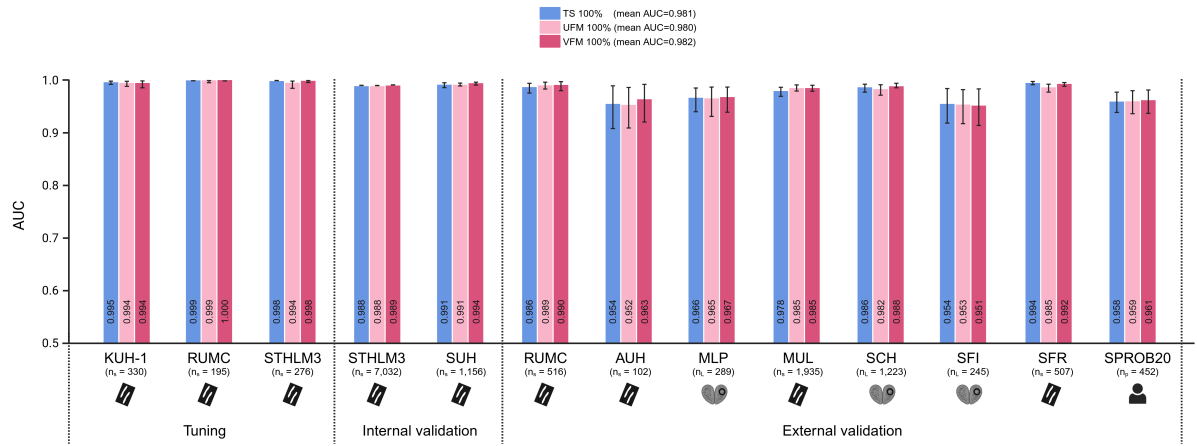

**Extended Data Figure 5: Discriminatory capacity of AI models in cancer detection across international cohorts.** The models' (TS, UFM, VFM; all trained with 100% of the task-specific data) cancer detection performance against the original reference standard on each cohort, measured by AUC using ROC analysis. In addition, the mean AUC over all cohorts is indicated (see legend). The $n_s$, $n_l$ and $n_p$ indicate the number of glass slides, anatomical locations or patients, respectively, included in each cohort. The values indicated by the bars represent point estimates and the error bars represent 95% confidence intervals estimated by bootstrapping. AUC=Area Under the Curve, ROC=Receiver operating characteristic, TS=Task-specific model, UFM=UNI foundation model, VFM=Virchow2 foundation model.[8]

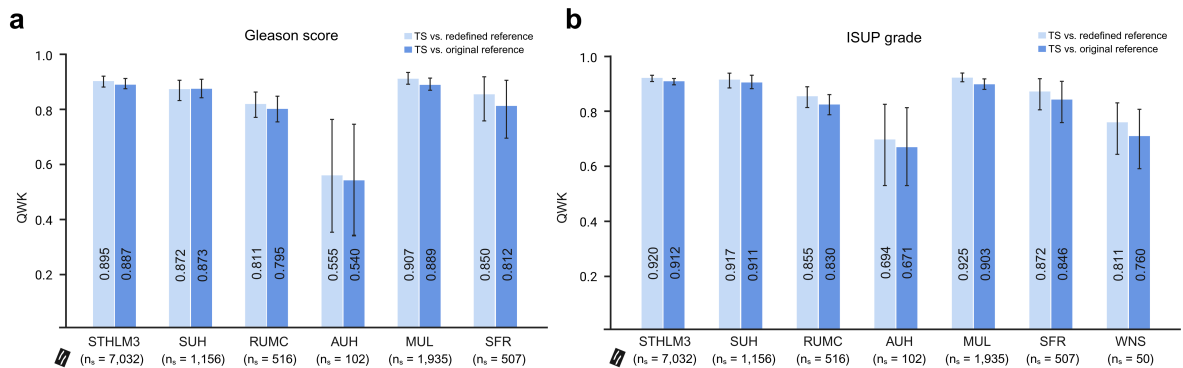

**Extended Data Figure 6: Gleason grading performance of the end-to-end model relative to the refined reference standard. (a-b)** Concordance of the TS model with the original reference standards defined by cohort-specific local pathologists vs. the updated reference standards followed by re-assessment of significant errors by the lead study pathologist quantified in terms of QWK for Gleason score **(a)** and for ISUP grade **(b)**. To allow for a direct comparison, only cohorts with a slide-level original reference standard were included. The $n_s$ indicate the number of glass slides included in each cohort. The values indicated by the bars represent point estimates and the error bars represent 95% confidence intervals estimated by bootstrapping. ISUP=International Society of Urological Pathology, QWK=Quadratically weighted Cohen's kappa, TS=Task-specific model.[9]

---

8  Blilie, A. (2025) https://BioRender.com/n50j237

9  Mulliqi, N. (2025) https://BioRender.com/m66s151

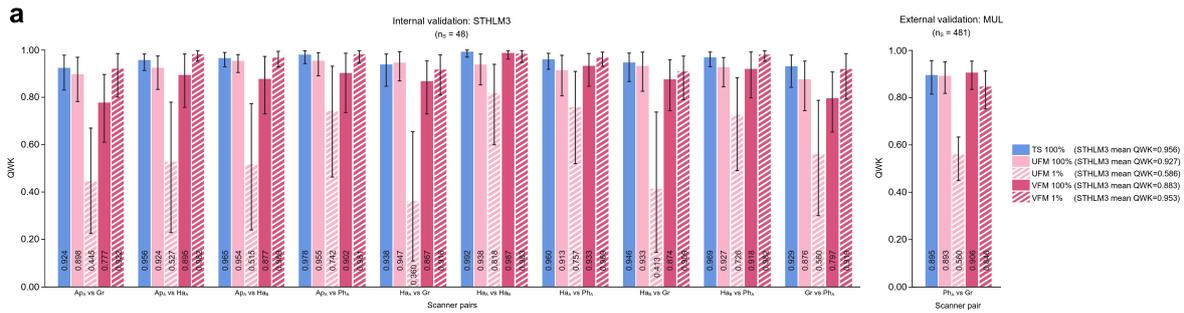
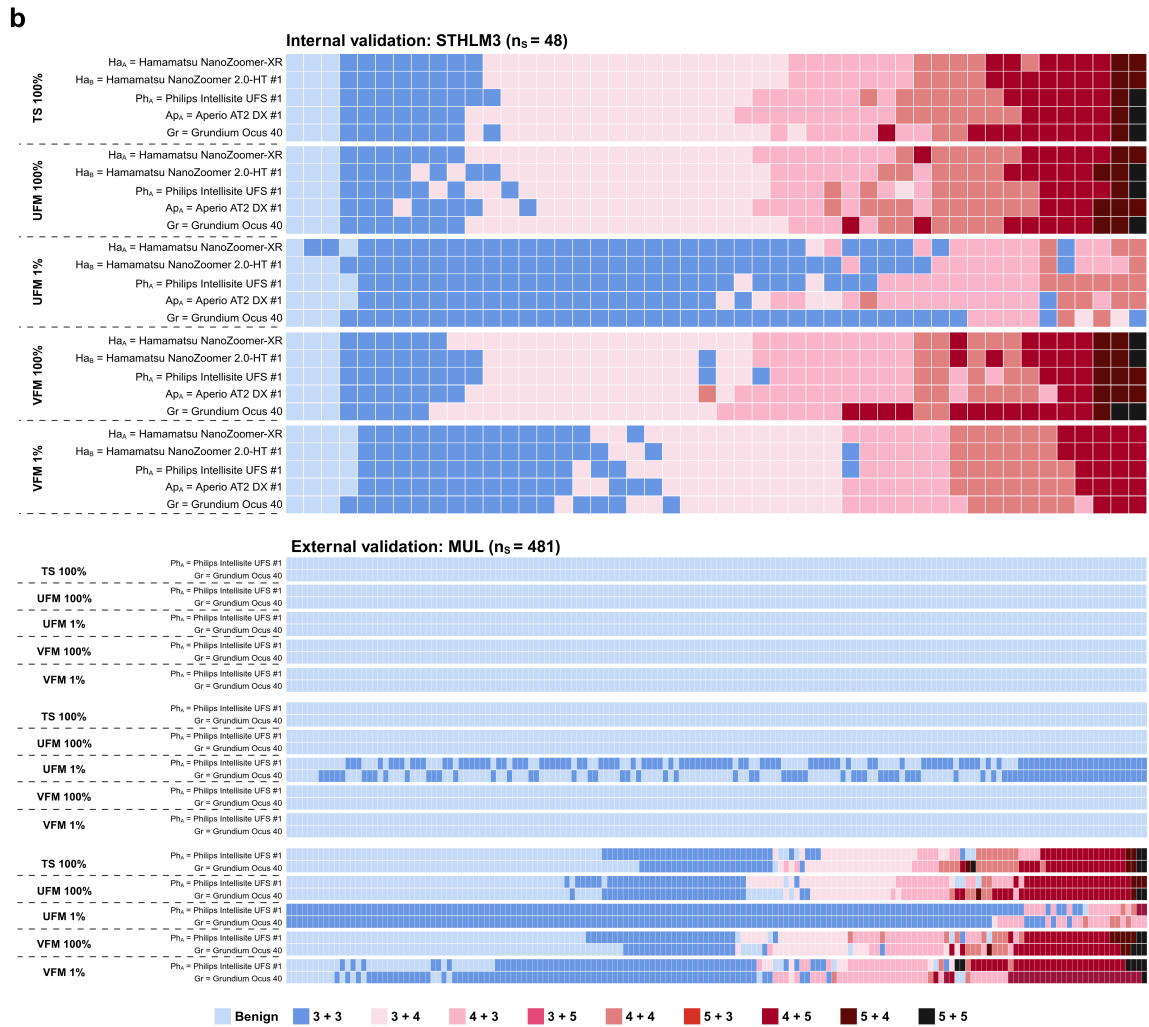

**Extended Data Figure 7: Cross-scanner Gleason scoring performance of AI models. (a)** The cross-scanner concordance of each model (TS, UFM, VFM; trained with 1% or 100% of the task-specific data) measured in terms of QWK for Gleason score, comparing each model's grading for the same slides digitized on pairs of different scanners (incl. 5 scanners for the STHLM3 cohort and 2 scanners for the MUL cohort). In addition, the mean of the 10 pairwise comparisons on the STHLM3 cohort is indicated (see legend). **(b)** Heatmaps showing the Gleason score predictions by each model across glass slides (columns) digitized on different scanners (rows). For each model, the slides are sorted for visual readability based on the model's average Gleason score prediction across scanners i.e. the column ordering is different between models. The number of glass slides included from the internal (STHLM3) and external (MUL) validation cohorts is indicated by $n_s$. The bars indicate point estimates and the error bars represent 95% confidence intervals estimated by bootstrapping.

QWK=Quadratically weighted Cohen's kappa, TS=Task-specific model, UFM=UNI foundation model, VFM=Virchow2 foundation model.

|  | Cohort | Model | GS LWK | Malignant GS LWK | ISUP LWK | Malignant ISUP LWK |
|---|---|---|---|---|---|---|
| Tuning | KUH-1 ($n_s = 330$, $n_{s+} = 222$) | TS | 0.755 (0.709-0.803) | 0.642 (0.574-0.709) | 0.800 (0.758-0.841) | 0.6657 (0.598-0.726) |
|  | KUH-1 ($n_s = 330$, $n_{s+} = 222$) | UFM | 0.797 (0.753-0.838) | 0.711 (0.650-0.768) | 0.829 (0.789-0.863) | 0.7284 (0.668-0.786) |
|  | KUH-1 ($n_s = 330$, $n_{s+} = 222$) | VFM | 0.776 (0.731-0.817) | 0.676 (0.613-0.738) | 0.810 (0.772-0.846) | 0.6862 (0.629-0.746) |
|  | RUMC ($n_s = 195$, $n_{s+} = 100$) | TS | 0.800 (0.740-0.850) | 0.661 (0.566-0.740) | 0.841 (0.791-0.885) | 0.6562 (0.567-0.734) |
|  | RUMC ($n_s = 195$, $n_{s+} = 100$) | UFM | 0.770 (0.705-0.821) | 0.638 (0.547-0.712) | 0.815 (0.766-0.860) | 0.6474 (0.566-0.713) |
|  | RUMC ($n_s = 195$, $n_{s+} = 100$) | VFM | 0.805 (0.752-0.856) | 0.673 (0.591-0.746) | 0.858 (0.816-0.895) | 0.7072 (0.626-0.781) |
|  | STHLM3 ($n_s = 276$, $n_{s+} = 84$) | TS | 0.822 (0.767-0.875) | 0.641 (0.529-0.738) | 0.862 (0.816-0.900) | 0.6837 (0.579-0.769) |
|  | STHLM3 ($n_s = 276$, $n_{s+} = 84$) | UFM | 0.815 (0.759-0.864) | 0.643 (0.523-0.745) | 0.856 (0.815-0.894) | 0.691 (0.593-0.772) |
|  | STHLM3 ($n_s = 276$, $n_{s+} = 84$) | VFM | 0.785 (0.724-0.847) | 0.558 (0.424-0.675) | 0.830 (0.782-0.873) | 0.5969 (0.481-0.695) |
| Internal validation | STHLM3 ($n_s = 7,032$, $n_{s+} = 1,949$) | TS | 0.822 (0.810-0.834) | 0.691 (0.669-0.715) | 0.846 (0.836-0.856) | 0.707 (0.687-0.730) |
|  | STHLM3 ($n_s = 7,032$, $n_{s+} = 1,949$) | UFM | 0.814 (0.802-0.826) | 0.680 (0.656-0.703) | 0.839 (0.830-0.849) | 0.702 (0.680-0.723) |
|  | STHLM3 ($n_s = 7,032$, $n_{s+} = 1,949$) | VFM | 0.813 (0.800-0.826) | 0.689 (0.666-0.712) | 0.836 (0.827-0.848) | 0.704 (0.682-0.726) |
|  | SUH ($n_s = 1,156$, $n_{s+} = 420$) | TS | 0.807 (0.780-0.833) | 0.671 (0.618-0.715) | 0.848 (0.824-0.871) | 0.705 (0.663-0.747) |
|  | SUH ($n_s = 1,156$, $n_{s+} = 420$) | UFM | 0.809 (0.781-0.835) | 0.694 (0.644-0.743) | 0.842 (0.819-0.864) | 0.716 (0.677-0.756) |
|  | SUH ($n_s = 1,156$, $n_{s+} = 420$) | VFM | 0.812 (0.786-0.837) | 0.706 (0.659-0.746) | 0.845 (0.823-0.866) | 0.731 (0.693-0.770) |
| External validation | RUMC ($n_s = 516$, $n_{s+} = 321$) | TS | 0.676 (0.631-0.718) | 0.502 (0.435-0.569) | 0.721 (0.682-0.759) | 0.498 (0.430-0.557) |
|  | RUMC ($n_s = 516$, $n_{s+} = 321$) | UFM | 0.667 (0.624-0.713) | 0.532 (0.472-0.590) | 0.707 (0.663-0.742) | 0.525 (0.463-0.587) |
|  | RUMC $n_s = 516$, $n_{s+} = 321$) | VFM | 0.622 (0.574-0.671) | 0.482 (0.418-0.543) | 0.678 (0.632-0.722) | 0.481 (0.418-0.541) |
|  | AUH ($n_s = 102$, $n_{s+} = 59$) | TS | 0.580 (0.453-0.702) | 0.345 (0.161-0.539) | 0.631 (0.513-0.748) | 0.363 (0.180-0.522) |
|  | AUH ($n_s = 102$, $n_{s+} = 59$) | UFM | 0.530 (0.404-0.663) | 0.254 (0.068-0.432) | 0.583 (0.467-0.700) | 0.264 (0.076-0.439) |
|  | AUH ($n_s = 102$, $n_{s+} = 59$) | VFM | 0.532 (0.415-0.652) | 0.263 (0.101-0.426) | 0.584 (0.475-0.698) | 0.276 (0.126-0.447) |
|  | MLP ($n_L$ 289, $n_{L+} = 219$) | TS | 0.658 (0.591-0.715) | 0.553 (0.472-0.620) | 0.693 (0.639-0.744) | 0.560 (0.485-0.628) |
|  | MLP ($n_L$ 289, $n_{L+} = 219$) | UFM | 0.651 (0.592-0.706) | 0.546 (0.476-0.619) | 0.683 (0.629-0.734) | 0.546 (0.475-0.615) |
|  | MLP ($n_L$ 289, $n_{L+} = 219$) | VFM | 0.613 (0.553-0.669) | 0.497 (0.426-0.570) | 0.648 (0.591-0.702) | 0.501 (0.432-0.572) |
|  | MUL ($n_s = 1,935$, $n_{s+} = 472$) | TS | 0.791 (0.766-0.814) | 0.612 (0.567-0.653) | 0.816 (0.794-0.838) | 0.606 (0.558-0.654) |
|  | MUL ($n_s = 1,935$, $n_{s+} = 472$) | UFM | 0.785 (0.760-0.808) | 0.614 (0.569-0.656) | 0.810 (0.787-0.832) | 0.613 (0.562-0.657) |
|  | MUL ($n_s = 1,935$, $n_{s+} = 472$) | VFM | 0.777 (0.753-0.799) | 0.601 (0.558-0.647) | 0.796 (0.773-0.818) | 0.576 (0.528-0.620) |
|  | SCH ($n_L = 1,223$, $n_{L+} = 356$) | TS | 0.736 (0.704-0.766) | 0.506 (0.448-0.564) | 0.779 (0.753-0.803) | 0.524 (0.473-0.576) |
|  | SCH ($n_L = 1,223$, $n_{L+} = 356$) | UFM | 0.719 (0.687-0.752) | 0.480 (0.428-0.536) | 0.762 (0.737-0.789) | 0.498 (0.449-0.552) |
|  | SCH ($n_L = 1,223$, $n_{L+} = 356$) | VFM | 0.744 (0.714-0.772) | 0.509 (0.447-0.564) | 0.779 (0.754-0.802) | 0.517 (0.462-0.570) |
|  | SFI ($n_L = 245$, $n_{L+} = 99$) | TS | 0.808 (0.743-0.863) | 0.698 (0.586-0.778) | 0.804 (0.751-0.854) | 0.630 (0.522-0.722) |
|  | SFI ($n_L = 245$, $n_{L+} = 99$) | UFM | 0.757 (0.694-0.814) | 0.606 (0.493-0.700) | 0.782 (0.721-0.837) | 0.590 (0.472-0.694) |
|  | SFI ($n_L = 245$, $n_{L+} = 99$) | VFM | 0.782 (0.716-0.838) | 0.700 (0.585-0.783) | 0.778 (0.711-0.832) | 0.644 (0.536-0.734) |
|  | SFR ($n_s = 507$, $n_{s+} = 134$) | TS | 0.746 (0.687-0.803) | 0.635 (0.535-0.722) | 0.767 (0.712-0.817) | 0.614 (0.507-0.702) |
|  | SFR ($n_s = 507$, $n_{s+} = 134$) | UFM | 0.639 (0.576-0.701) | 0.517 (0.403-0.601) | 0.662 (0.598-0.719) | 0.495 (0.388-0.586) |
|  | SFR ($n_s = 507$, $n_{s+} = 134$) | VFM | 0.649 (0.575-0.715) | 0.533 (0.415-0.621) | 0.671 (0.616-0.727) | 0.484 (0.376-0.573) |
|  | UKK ($n_s = 50$, $n_{s+} = 50$) | TS | - | - | 0.705 (0.571-0.810) | 0.705 (0.571-0.810) |
|  | UKK ($n_s = 50$, $n_{s+} = 50$) | UFM | - | - | 0.761 (0.644-0.856) | 0.761 (0.644-0.856) |
|  | UKK ($n_s = 50$, $n_{s+} = 50$) | VFM | - | - | 0.699 (0.588-0.796) | 0.699 (0.588-0.796) |
|  | WNS ($n_s = 50$, $n_{s+} = 50$) | TS | - | - | 0.593 (0.467-0.699) | 0.593 (0.467-0.699) |
|  | WNS ($n_s = 50$, $n_{s+} = 50$) | UFM | - | - | 0.574 (0.420-0.702) | 0.574 (0.420-0.702) |
|  | WNS ($n_s = 50$, $n_{s+} = 50$) | VFM | - | - | 0.564 (0.408-0.693) | 0.564 (0.408-0.693) |
|  | SPROB20 ($n_P = 452$, $n_{P+} = 289$) | TS | - | - | 0.708 (0.657-0.753) | 0.552 (0.485-0.614) |
|  | SPROB20 ($n_P = 452$, $n_{P+} = 289$) | UFM | - | - | 0.716 (0.671-0.758) | 0.575 (0.512-0.637) |
|  | SPROB20 ($n_P = 452$, $n_{P+} = 289$) | VFM | - | - | 0.713 (0.671-0.758) | 0.552 (0.488-0.617) |

**Extended Data Table 1. Gleason grading performance of AI models relative to the original reference standard in terms of Cohen's linearly weighted kappa (LWK).** Models' (TS, UFM, VFM; all trained with 100% of the task-specific data) concordance with the original reference standard measured with LWK for Gleason score and ISUP grade. The UKK, WNS, SPROB20 lack reference Gleason scoring. Additionally, LWK was evaluated on malignant slides only. The $n_s$, $n_l$ and $n_p$ indicate the number of glass slides, anatomical locations or patients, respectively, included in each cohort. Corresponding numbers of malignant cases are indicated with $n_{s+}$, $n_{l+}$ and $n_{p+}$. The LWK values indicate point estimates, with 95% CI estimated by bootstrapping given in parentheses. GS=Gleason score, ISUP=International Society of Urological Pathology, LWK=Linearly weighted Cohen's kappa, TS=Task-specific model, UFM=UNI foundation model, VFM=Virchow2 foundation model.

|  | Cohort | Model | GS LWK | Malignant GS LWK | ISUP LWK | Malignant ISUP LWK |
| --- | --- | --- | --- | --- | --- | --- |
| Internal validation | SUH ($n_s = 66$, $n_{s+} = 53$) | TS | 0.586 (0.437-0.717) | 0.501 (0.350-0.640) | 0.636 (0.507-0.743) | 0.521 (0.380-0.655) |
|  | SUH ($n_s = 66$, $n_{s+} = 53$) | UFM | 0.561 (0.393-0.693) | 0.469 (0.323-0.603) | 0.617 (0.477-0.727) | 0.491 (0.339-0.633) |
|  | SUH ($n_s = 66$, $n_{s+} = 53$) | VFM | 0.669 (0.533-0.785) | 0.603 (0.453-0.731) | 0.718 (0.606-0.815) | 0.622 (0.481-0.744) |
|  | SUH ($n_s = 66$, $n_{s+} = 53$) | Pathologist | 0.592 (0.461-0.704) | 0.464 (0.318-0.606) | 0.672 (0.552-0.779) | 0.520 (0.356-0.654) |
| External validation | RUMC ($n_s = 66$, $n_{s+} = 52$) | TS | 0.678 (0.552-0.798) | 0.625 (0.474-0.749) | 0.719 (0.599-0.821) | 0.650 (0.510-0.760) |
|  | RUMC ($n_s = 66$, $n_{s+} = 52$) | UFM | 0.756 (0.655-0.841) | 0.704 (0.559-0.802) | 0.789 (0.695-0.871) | 0.736 (0.620-0.843) |
|  | RUMC ($n_s = 66$, $n_{s+} = 52$) | VFM | 0.716 (0.592-0.813) | 0.649 (0.510-0.778) | 0.772 (0.667-0.863) | 0.692 (0.553-0.806) |
|  | RUMC ($n_s = 66$, $n_{s+} = 52$) | Pathologist | 0.724 (0.608-0.824) | 0.649 (0.499-0.760) | 0.697 (0.575-0.799) | 0.570 (0.433-0.692) |
|  | AUH ($n_s = 41$, $n_{s+} = 30$) | TS | 0.854 (0.718-0.959) | 0.773 (0.553-0.943) | 0.874 (0.758-0.970) | 0.785 (0.587-0.943) |
|  | AUH ($n_s = 41$, $n_{s+} = 30$) | UFM | 0.803 (0.649-0.923) | 0.689 (0.469-0.867) | 0.814 (0.701-0.915) | 0.676 (0.466-0.834) |
|  | AUH ($n_s = 41$, $n_{s+} = 30$) | VFM | 0.795 (0.656-0.906) | 0.673 (0.464-0.854) | 0.817 (0.709-0.906) | 0.676 (0.484-0.824) |
|  | AUH ($n_s = 41$, $n_{s+} = 30$) | Pathologist | 0.623 (0.417-0.809) | 0.442 (0.138-0.713) | 0.633 (0.436-0.798) | 0.421 (0.165-0.662) |
|  | MLP ($n_L = 66$, $n_{L+} = 43$) | TS | 0.859 (0.793-0.908) | 0.778 (0.665-0.866) | 0.887 (0.825-0.936) | 0.797 (0.684-0.882) |
|  | MLP ($n_L = 66$, $n_{L+} = 43$) | UFM | 0.828 (0.744-0.898) | 0.763 (0.654-0.849) | 0.841 (0.755-0.906) | 0.756 (0.642-0.848) |
|  | MLP ($n_L = 66$, $n_{L+} = 43$) | VFM | 0.790 (0.691-0.875) | 0.719 (0.576-0.834) | 0.820 (0.719-0.900) | 0.750 (0.617-0.855) |
|  | MLP ($n_L = 66$, $n_{L+} = 43$) | Pathologist | - | - | - | - |
|  | MUL ($n_s = 66$, $n_{s+} = 51$) | TS | 0.729 (0.612-0.837) | 0.588 (0.411-0.737) | 0.791 (0.667-0.881) | 0.619 (0.440-0.776) |
|  | MUL ($n_s = 66$, $n_{s+} = 51$) | UFM | 0.734 (0.619-0.827) | 0.587 (0.424-0.735) | 0.815 (0.720-0.895) | 0.654 (0.468-0.807) |
|  | MUL ($n_s = 66$, $n_{s+} = 51$) | VFM | 0.803 (0.711-0.875) | 0.693 (0.545-0.807) | 0.847 (0.765-0.914) | 0.704 (0.549-0.823) |
|  | MUL ($n_s = 66$, $n_{s+} = 51$) | Pathologist | 0.667 (0.519-0.785) | 0.570 (0.404-0.714) | 0.677 (0.532-0.802) | 0.531 (0.352-0.683) |
|  | SCH ($n_L = 72$, $n_{L+} = 38$) | TS | 0.755 (0.657-0.835) | 0.570 (0.413-0.691) | 0.797 (0.706-0.873) | 0.587 (0.411-0.729) |
|  | SCH ($n_L = 72$, $n_{L+} = 38$) | UFM | 0.816 (0.730-0.890) | 0.673 (0.521-0.796) | 0.862 (0.796-0.918) | 0.710 (0.571-0.828) |
|  | SCH ($n_L = 72$, $n_{L+} = 38$) | VFM | 0.802 (0.718-0.876) | 0.643 (0.486-0.773) | 0.848 (0.779-0.911) | 0.674 (0.525-0.792) |
|  | SCH ($n_L = 72$, $n_{L+} = 38$) | Pathologist | - | - | - | - |
|  | SFI ($n_L = 67$, $n_{L+} = 48$) | TS | 0.790 (0.669-0.879) | 0.700 (0.527-0.821) | 0.825 (0.721-0.906) | 0.712 (0.561-0.834) |
|  | SFI ($n_L = 67$, $n_{L+} = 49$) | UFM | 0.754 (0.621-0.859) | 0.641 (0.461-0.782) | 0.799 (0.685-0.890) | 0.662 (0.489-0.815) |
|  | SFI ($n_L = 67$, $n_{L+} = 49$) | VFM | 0.819 (0.736-0.883) | 0.734 (0.623-0.827) | 0.839 (0.754-0.907) | 0.723 (0.591-0.825) |
|  | SFI ($n_L = 67$, $n_{L+} = 49$) | Pathologist | 0.687 (0.574-0.777) | 0.608 (0.461-0.720) | 0.676 (0.549-0.776) | 0.555 (0.418-0.677) |
|  | SFR ($n_s = 49$, $n_{s+} = 40$) | TS | 0.832 (0.732-0.906) | 0.775 (0.656-0.877) | 0.876 (0.801-0.946) | 0.815 (0.698-0.909) |
|  | SFR ($n_s = 49$, $n_{s+} = 40$) | UFM | 0.797 (0.683-0.887) | 0.723 (0.572-0.847) | 0.792 (0.684-0.876) | 0.682 (0.540-0.794) |
|  | SFR ($n_s = 49$, $n_{s+} = 40$) | VFM | 0.762 (0.644-0.847) | 0.686 (0.540-0.795) | 0.774 (0.663-0.856) | 0.673 (0.542-0.781) |
|  | SFR ($n_s = 49$, $n_{s+} = 40$) | Pathologist | 0.762 (0.668-0.836) | 0.693 (0.573-0.786) | 0.832 (0.737-0.906) | 0.764 (0.638-0.863) |
|  | SPROB20 ($n_P = 50$, $n_{P+} = 21$) | TS | 0.876 (0.766-0.961) | 0.769 (0.546-0.936) | 0.882 (0.791-0.960) | 0.768 (0.551-0.929) |
|  | SPROB20 ($n_P = 50$, $n_{P+} = 21$) | UFM | 0.565 (0.336-0.810) | 0.684 (0.513-0.812) | 0.650 (0.455-0.834) | 0.695 (0.541-0.823) |
|  | SPROB20 ($n_P = 50$, $n_{P+} = 21$) | VFM | 0.603 (0.393-0.810) | 0.612 (0.410-0.774) | 0.696 (0.504-0.863) | 0.679 (0.483-0.838) |
|  | SPROB20 ($n_P = 50$, $n_{P+} = 21$) | Pathologist | - | - | - | - |

**Extended Data Table 2. Gleason grading performance of AI models relative to uniform reference standard in terms of Cohen's linearly weighted kappa (LWK).** Models' (TS, UFM, VFM; all trained with 100% of the task-specific data) and local cohort-specific pathologists' ("Pathologist") concordance with the uniform reference standard by the lead study pathologist, measured with LWK for Gleason score and ISUP grade. For the MLP, SCH and SPROB20 cohorts, the original reference standard was not reported on slide-level and the comparison between the local pathologists and the lead study pathologist is omitted. Additionally, LWK was evaluated on malignant slides only. The $n_s$, $n_l$ and $n_p$ indicate the number of glass slides, anatomical locations or patients, respectively, included in each cohort. Corresponding numbers of malignant cases are indicated with $n_{s+}$, $n_{l+}$ and $n_{p+}$. The LWK values indicate point estimates, with 95% CI estimated by bootstrapping given in parentheses. GS=Gleason score, ISUP=International Society of Urological Pathology, LWK=Linearly weighted Cohen's kappa, TS=Task-specific model, UFM=UNI foundation model, VFM=Virchow2 foundation model.

# Methods

## Study design and datasets

The study followed a pre-specified design (see **Supplementary Appendix 1** for the study protocol). Details on the included patient cohorts are provided in the study protocol and an overview is shown in **Figure 1**. All patients from the participating clinical sites who underwent prostate core needle biopsy were initially eligible for inclusion. The exclusion criteria were based on problems with information retrieval (e.g. missing, mismatched, or ambiguous identifiers or pathology information), staining and slide preparation (e.g., non-prostate tissue, slides not stained with hematoxylin and eosin (HE), pen markings on tissue), or slide digitization (e.g. corrupt files).

The international prostate cancer digital pathology dataset includes 7,342 patients who underwent prostate biopsies between 2012 and 2023, resulting in approximately 100,000 core needle biopsies and 82,000 WSIs. Samples originate from 15 clinical sites or clinical trials across 11 countries (Austria, Australia, Denmark, Finland, France, Germany, the Netherlands, Norway, Poland, Sweden and Switzerland). All slides represent formalin-fixed, paraffin-embedded (FFPE) HE-stained prostate core needle biopsy specimens with a varying number of cores and/or tissue sections per slide. The data were partitioned into a model development set containing training (n=55,798 WSIs) and tuning (n=1,177 WSIs) cohorts and into a validation set containing internal (n=14,808 WSIs) and external (n=10,801 WSIs) validation cohorts. The data were split at patient-level, i.e. all WSIs representing a given patient were randomized together to avoid information leakage between development and validation data[70,71].

The study involved two phases: a development phase and a validation phase. During the development phase, 10-fold cross-validation on the development data and separate evaluations on the tuning cohort were used to evaluate the effects of different design choices on model performance. After the model design was completed, the fixed AI models were evaluated in the validation phase. Validation data were held-out and not analyzed in any way prior to the design freeze. The internal validation data represent laboratories and/or scanners that are also present in the development set, whereas the external validation data are fully independent of the development or tuning cohorts in terms of patients, laboratories and scanners. There are three deviations from these rules: 1) The AQ validation cohort is not fully external due to a subset of it being scanned on the same Philips scanner that was involved in the development set (we present results separately for the partly external and fully external subsets of AQ), 2) A subset of the MUL validation cohort was re-scanned using the same Philips scanner that was involved in the development set (we present results on these WSIs only in a cross-scanner consistency experiment, and used a fully external Grundium scanner for the main analysis of this cohort), 3) The RUMC cohort was allocated for development and internal validation in the study protocol, but we chose not to use it for model training (that is, the RUMC validation cohort is fully external to the model, but RUMC development data were accessible prior to design freeze).

## Whole slide scanning

Slides were scanned using 14 WSI scanner instruments (9 different models from 5 vendors) including Philips IntelliSite UFS, Hamamatsu NanoZoomer (XR, 2.0-HT, S60, S360), Aperio (AT2 DX), 3DHISTECH Pannoramic (250 Flash III, Scan ll), and Grundium Ocus40. The distribution of scanned WSIs across scanners is presented in **Figure 6A**. In the STHLM3 and MUL cohorts, subsets of slides were re-scanned using 5 and 2 scanners, respectively. During the model design phase, multiple WSIs per slide in the STHLM3 development cohort were used as a data augmentation technique. During the model validation phase, one WSI per slide was picked randomly from the STHLM3 internal validation cohort for evaluation, and only WSIs captured with the fully external Grundium scanner were used from the MUL external validation cohort. The other WSIs of these slides were reserved only for cross-scanner reproducibility analysis. Importantly—all external validation cohorts (except for a subset of AQ, see "Study design and datasets") consist of slides digitized with scanners different from those used during model development. For details, see **Supplementary Appendix 1: Table 2**.

## Data management

Our data collection, management and verification process generally followed the process: patient identifiers were pseudonymized at extraction at each site, slides were scanned and identifiers were stored in filenames or metadata. Linking slide data to clinical/pathology information involved parsing filenames, resolving inconsistencies, and employing semi-automated OCR systems to extract identifiers. To ensure unique labeling and an added layer of centralized pseudonymization, MD5 hashes were generated based on filenames, scanner serial numbers, scanning time and the original patient/slide identifiers. The clinical and pathology data for the cohorts were retrieved through various methods: directly from registries, provided in tabular form by data providers, or manually tabulated from scanned pathology reports. Comprehensive unit testing was implemented with Python's *unittest* framework to verify dataset integrity. For details, see **Supplementary Appendix 1.**

## Reference standard protocols

Reference standards in the form of a pathologist's Gleason grading were provided by the lead study pathologist (L.E.) for the KUH-1, KUH-2, STG, and STHLM3 cohorts. Local pathologists at each site graded the AQ, AUH, MLP, MUL, SCH, SFI, SFR, SPROB20 and SUH cohorts. Panels of pathologists graded the following validation cohorts: the ImageBase[45] subset of STHLM3 (23 pathologists), the subsets of RUMC serving as test sets in PANDA[26] (4 pathologists), UKK (10 pathologists), and WNS (11 pathologists). The granularity of reference standards differed across sites, including assessment per slide (AQ, AUH, KUH-1, KUH-2, MUL, RUMC, SFR, STG, STHLM3, SUH, UKK, WNS), per anatomical prostate location (MLP, SCH, SFI), and per patient (SPROB20). Only the ISUP grade was provided for the SPROB20, UKK, and WNS cohorts, whereas all other cohorts have both ISUP grade and Gleason score reported. Reference standards across all cohorts were obtained conventionally using a microscope, except for digital assessment for the UKK and WNS cohorts. For details, see **Supplementary Appendix 1: Table 3**.

A uniform reference standard was additionally established by the lead study pathologist (L.E.) on subsets of slides from all internal and external validation cohorts originally assessed by other pathologists (except for cohorts with a panelist reference standard: UKK, WNS). Slides for re-assessment were selected randomly, and stratified on ISUP grade based on the original grading. Additionally, the lead study pathologist re-assessed cases with clinically significant errors. All re-assessments were conducted blinded to the original grading and AI predictions, with grading reported per slide using the Cytomine platform[72]. For details, see **Supplementary Appendix 1.**

## Tissue detection and tiling

Tissue was detected from WSIs using an in-house developed tissue segmentation model based on a UNet++ architecture[73] with a ResNeXt-101 (32 x 4d)[74] encoder. Patches of 512×512 pixels were first extracted across the entire WSI area at a resolution of 8.0 µm/px with an overlap of 128 px, segmented pixel-wise to detect tissue, and then stitched into a single binary segmentation mask per WSI. Subsequently, high-resolution tissue patches of 256×256 px were extracted at 1.0 µm/px from the WSIs using the segmented tissue masks to select only patches with minimally 10% of pixels detected as tissue. Tissue patches were extracted either without overlap (for model training, to limit GPU memory footprint) or with 128 px overlap (for predictions, to improve diagnostic performance). Patches were downsampled from the closest, higher resolution in the resolution pyramid to 1.0 µm/px using Lancsoz resampling. Patches were stored in the disk-friendly TFRecord data format[75], with one file per WSI.

## Model architectures

All models were built using an attention-based multiple instance learning (ABMIL)[20] architecture with weak slide-level supervision, using either an end-to-end trained task-specific patch encoder or frozen foundation model encoders. The encoders extract feature embeddings from patches, which are then attention-aggregated into slide-level feature vectors for further classification. The TS model architecture used an EfficientNet-V2-S encoder[42] to extract 1280-dimensional patch-level embeddings. The FMs were trained by their developers using the DINO v2 self-supervised learning algorithm[76] which employs a student–teacher paradigm with both student and teacher networks utilizing the same architecture (ViT-L/16 for UFM, and ViT-H/14 for VFM)[9,41] and produce 1024-dimensional (UNI) and 1280-dimensional (VFM) patch embeddings.

After the patch encoder, the feature vectors first undergo average pooling and a fully connected (FC) layer to reduce them to 1x1000 dimensions. Subsequently, they are passed through a gated-variant ABMIL aggregator, first transformed into 512-dimensional representations through a linear layer and then into hidden 384-dimensional representations in intermediate layers. Outputs from these layers are combined through element-wise multiplication, and a final linear layer is used to compute attention weights for each patch. The attention weights are normalized using the softmax function and used as weights to the weighted mean of all patch embeddings that produce a final 512-dimensional slide-level feature vector. The slide-level feature vector is processed with classification layers for predicting primary and secondary Gleason patterns for the set of input patches passed through the model (for slide-level grading: all patches from a WSI). The final classification layers contain FC layers and rectified linear unit (ReLU) activations, finally outputting two vectors (the primary and secondary Gleason patterns) of 4-class classification logits (i.e. benign, Gleason pattern 3, pattern 4 or pattern 5), followed by softmax. We applied dropout (p=0.2) to the input embeddings as well as after each intermediate layer in the aggregator and classifier networks for regularisation.

## Model training

The TS model was trained with end-to-end learning, where all model parameters (incl. the encoder) were jointly optimized with regard to a single loss function. The training of the FM-based models consisted in keeping the FM encoders frozen with the weights from their self-supervised pre-training, and training only the attention-aggregator and classification layers. The EfficientNet-V2-S encoder was initialized with ImageNet-pretrained weights, and the attention module and classification network were initialized using Xavier initialisation[77]. All models were trained using cross-entropy loss and AdamW optimizer[78] with a base learning rate (lr=0.0001).

At each iteration, up to 1,800 patches were randomly sampled without replacement per WSI to serve as a single minibatch associated with a WSI-level label. Training was run on multiple GPUs using the distributed data-parallel (DDP) PyTorch framework[79] with the NVIDIA collective communication library (nccl) backend. Gradient accumulation was performed by averaging gradients over 4 iterations across 8 GPUs to obtain an effective minibatch size of 32 WSIs. The number of tissue patches per minibatch varied, as not all WSIs had 1,800 patches. Each minibatch was sampled by selecting one WSI per GPU process, accumulating gradients independently on each GPU, and averaging them across processes before the optimizer step. We used PyTorch's automatic mixed precision (AMP)[80] and gradient scaler for optimized memory use. In addition, to reduce the GPU memory footprint of end-to-end training, we applied activation checkpointing (i.e. recomputing activations during the backward passes instead of storing them in GPU memory from the forward pass) for the TS encoder. Additionally, memory pre-allocation was performed at the start of training to decrease GPU memory fragmentation due to variable-sized input WSIs having different patch numbers.

The model has two output heads both using cross-entropy losses for predicting the primary and secondary Gleason patterns of a WSI. The overall loss is the summed loss of the two heads, normalized by the gradient accumulation interval. The labels (i.e. 0 for benign slides and Gleason patterns 3, 4 and 5 for cancer) are mapped to an ordinal evenly spaced scale (i.e. 0, 1, 2, 3) before loss calculations. After having obtained the raw output values from the models' classification heads, we apply the argmax rule to get Gleason patterns (i.e. 0, 1, 2, or 3). Given that there is no in-built regularization in the model to avoid invalid combinations of zero and non-zero Gleason patterns, we duplicate the values of single non-zero Gleason patterns (e.g. 0+3 will be corrected to 3+3). To compute metrics during training, the Gleason scores are encoded onto an ordinal scale: benign (0), 3+3 (1), 3+4 (2), 4+3 (3), 3+5 (4), 4+4 (5), 5+3 (6), 4+5 (7), 5+4 (8), 5+5 (9) as defined in other studies[81–83]

To improve the robustness of the models to scanner and staining variation, we employ several types of data augmentations. First, we use the subsets of training slides re-scanned on multiple instruments for scanner augmentation where on each epoch, one WSI per slide is randomly picked. To simulate slide-level staining and scanning variation, we apply stain augmentation by generating simulated variability in stain intensity and distribution[84] together with Gaussian blur, unsharp masking and color augmentations to all patches of a WSI. Furthermore, we use Sierra color calibration[59] as an augmentation technique, where color calibration is applied to all patches of a WSI during training with a probability of 50%. Each patch is additionally independently processed with simple geometrical transformations (i.e. random flipping and 90° rotations) and noise-simulating augmentations (such as Gaussian noise, ISO noise, decreasing image quality by JPEG and WebP compression).

To reduce spurious correlations between image characteristics and the target labels during training, namely variation of ISUP grade distribution across laboratories and/or scanners, we apply a sampling scheme. At the beginning of each epoch, we sample slides such that a uniform distribution of ISUP grades is obtained for each scanner (that is, the ISUP grade distributions *between* scanners will also be identical). The models were trained using 10-fold cross-validation, with folds stratified by patients, ISUP grade and cohorts (STHLM3, SUH, STG). That is, each model was trained on 90% of the training data, and the remaining 10% in each CV fold was used to measure performance for early stopping. After every epoch, QWK for ISUP was measured on these test data, and training was stopped if no improvement took place in 200 epochs. From each CV fold, the best model was kept, resulting in an ensemble of 10 models.

## Model prediction

We applied test-time augmentation (TTA) with three iterations per model, using simple geometric transformations (i.e. flips and 90° rotations) applied randomly and independently to each patch. Due to some validation cohorts having labels assigned per prostate location or patient, we grouped the respective tissue patches and obtained predictions for the pooled set of WSIs with a shared reference standard label. All patches of each WSI were included to obtain predictions and processed stepwise in batches of 64 to limit memory usage in the patch encoder part of the models. The final model predictions are a slide-level classification predicting the final Gleason score and ISUP grade for a WSI (or multiple WSIs per location or patient) and patch-level classifications predicting the Gleason patterns per patch. Encoding the raw output values from the model to Gleason scores was done in the same way as during training. Model ensembling and TTA was done by obtaining the majority vote of the 30 predicted Gleason scores (10 models x 3 TTA runs), and further translated into an ISUP grade. The tile-level classification was performed by bypassing the ABMIL module and performing a classification task iteratively on each patch obtaining the Gleason pattern probabilities per patch. This approach allows for e.g. visualizing Gleason patterns in the WSI.

## Computational efficiency measurements

Computational efficiency in terms of time and energy consumption was measured by running predictions on the tuning set (n=801 slides) for the three models while logging power usage and runtime. To ensure that any observed differences in energy consumption were solely attributed to model complexity, each model was evaluated using identical hardware configurations (e.g., GPU models, memory allocation) and standard input settings. Experiments were run on 1 x NVIDIA A100 40GB GPU for one trained model and one TTA run. The final values were multiplied by 30 to obtain the total consumption for the 10-model ensemble with 3xTTA. The GPU power consumption was logged using NVIDIA System Management Interface (nvidia-smi) monitoring real-time GPU power draw at a frequency of 1 sample per second in Watts. The cumulative energy usage (kWh) was obtained by adding power draw values over the total runtime (GPUh) for each model. Additionally, per-biopsy energy consumption (Wh/biopsy) was calculated by normalizing total energy usage by the number of slides processed.

## Statistical analysis

To quantify the concordance between the models and the reference standards in terms of ISUP grade and Gleason score we used QWK and LWK, as implemented in scikit-learn[85,86]. The average concordance across models and pathologists was computed with the mean kappa values. To quantify the concordance of negative/positive diagnosis for prostate cancer with the reference standard we used sensitivity (true positive rate), specificity (true negative rate), and the AUROC. We calculated the 95% confidence intervals for the models' and pathologists' performance using bootstrapping over 1000 replicates. We addressed statistical confounders in the training and validation data which can very often cause models to exploit unintended correlations[57,87,88], leading to unrealistically optimistic estimates of performance as long as such correlations are available to the model in validation sets. For example, variations of the Gleason score and ISUP grade distributions across subsets of the data digitized with different scanners (and/or prepared in different labs) can introduce such bias linking scanner type or clinical site to diagnostic outcomes. To mitigate this, we use external validation cohorts and ISUP sampling strategies during training. Another potential confounder is pen marks placed by pathologists on the slides during diagnosis, which may lead models to associate pen markings with malignancy. We tackled this by segmenting only tissue for analysis and washing affected slides before scanning or excluding them if washing was not possible.

## Computing hardware and software

We used Python (v3.8.10), PyTorch (v2.0.0, CUDA 12.2) (https://pytorch.org) and PyTorch DDP for multi-GPU training for all experiments across all models. We used the pre-trained weights for UNI and Virchow2 FMs from their official releases on the HuggingFace hub (https://huggingface.co/MahmoodLab/UNI; https://huggingface.co/paige-ai/Virchow2) and integrated them with the ViT implementations provided by timm library (v0.9.8). All experiments were done on two high-performance clusters: Alvis (part of the National Academic Infrastructure for Supercomputing in Sweden) and Berzelius (part of the National Supercomputer Centre). On Alvis, training was done on 4 x 80GB NVIDIA A100 GPUs (256 GB system memory, 16 CPU cores per GPU). On Berzelius, training was done on 8 x 80 GB NVIDIA A100 GPUs (127 GB system memory, 16 CPU cores per GPU). Predictions were run on the clusters on a single 40 GB A100 NVIDIA GPU. Initial model development and prototyping were done locally on 2 x 24 GB NVIDIA GeForce RTX 3090 GPUs (127 GB system memory, 32 CPU cores). Docker (v20.10.21) was used locally, Singularity and Apptainer were used on the computing clusters. OpenSlide (v4.0.0), openslide-python (v1.3.1), and OpenPhi (v2.1.0) were used to access WSIs. DareBlopy (v0.0.5) was used for compatibility between the TFRecord data format (.tfrecord) and PyTorch. Albumentations (v1.3.1) and Stainlib (v0.6.1) were used for image augmentations. For implementing the tissue segmentation model PyTorch segmentation_models_pytorch library (v0.3.3) was used. NumPy (v1.24.0), scikit-learn (v1.2.2), and Pandas (v1.5.3) were used for numerical operations, model evaluation, and data management. Pillow (v9.4.0) and OpenCV-python were used for basic image processing tasks. Matplotlib (v3.7.1) and Seaborn (v0.12.2) were used for plots and figures and Biorender was used to assemble figure panels.

# Supplementary Appendix 1

# Study Protocol: Development and Retrospective Validation of an Artificial Intelligence System for Diagnostic Assessment of Prostate Biopsies

Version 1.0

Nita Mulliqi[1], Anders Blilie[2,3], Xiaoyi Ji[1], Kelvin Szolnoky[1], Henrik Olsson[1], Matteo Titus[1], Geraldine Martinez Gonzalez[1], Sol Erika Boman[1,4], Masi Valkonen[5], Einar Gudlaugsson[2], Svein R. Kjosavik[3,6], José Asenjo[7], Marcello Gambacorta[8], Paolo Libretti[8], Marcin Braun[9], Radzislaw Kordek[9], Roman Łowicki[10], Kristina Hotakainen[11,12], Päivi Väre[13], Bodil Ginnerup Pedersen[14,15], Karina Dalsgaard Sørensen[15,16], Benedicte Parm Ulhøi[17], Mattias Rantalainen[1], Pekka Ruusuvuori[4,18], Brett Delahunt[19], Hemamali Samaratunga[20], Toyonori Tsuzuki[21], Emilius A.M. Janssen[2,22], Lars Egevad[23], Kimmo Kartasalo[1], Martin Eklund[1]

1. Department of Medical Epidemiology and Biostatistics, Karolinska Institutet, Stockholm, Sweden
2. Department of Pathology, Stavanger University Hospital, Stavanger, Norway
3. Faculty of Health Sciences, University of Stavanger, Stavanger, Norway
4. Department of Molecular Medicine and Surgery, Karolinska Institutet, Stockholm, Sweden
5. Institute of Biomedicine, University of Turku, Turku, Finland
6. The General Practice and Care Coordination Research Group, Stavanger University Hospital, Norway
7. Department of Pathology, Synlab, Madrid, Spain
8. Department of Pathology, Synlab, Brescia, Italy
9. Department of Pathology, Chair of Oncology, Medical University of Lodz, Lodz, Poland
10. 1st Department of Urology, Medical University of Lodz, Lodz, Poland
11. Department of Clinical Chemistry, University of Helsinki, Helsinki, Finland
12. Laboratory Services, Mehiläinen Oy, Helsinki, Finland
13. Mehiläinen Länsi-Pohja Hospital, Kemi, Finland
14. Department of Radiology, Aarhus University Hospital, Aarhus, Denmark
15. Department of Clinical Medicine, Aarhus University, Aarhus, Denmark
16. Department of Molecular Medicine, Aarhus University Hospital, Aarhus, Denmark
17. Department of Pathology, Aarhus University Hospital, Aarhus, Denmark
18. Faculty of Medicine and Health Technology, Tampere University, Tampere, Finland
19. Department of Pathology and Molecular Medicine, Wellington School of Medicine and Health Sciences, University of Otago, Wellington, New Zealand
20. Aquesta Uropathology and University of Queensland, QLD, Brisbane, Australia
21. Department of Surgical Pathology, School of Medicine, Aichi Medical University, Nagoya, Japan
22. Faculty of Science and Technology, University of Stavanger, Stavanger, Norway
23. Department of Oncology and Pathology, Karolinska Institutet, Stockholm, Sweden

Corresponding author: Martin Eklund, martin.eklund@ki.se.

# Revision history

| Version and date | Status | Main changes |
|---|---|---|
| 1.0 (2024-07-04) | Initial publication. | Initial publication. |

# Abstract

Histopathological evaluation of prostate biopsies using the Gleason scoring system is critical for prostate cancer diagnosis and treatment selection. However, grading variability among pathologists can lead to inconsistent assessments, risking inappropriate treatment. Similar challenges complicate the assessment of other prognostic features like cribriform cancer morphology and perineural invasion. Many pathology departments are also facing an increasingly unsustainable workload due to rising prostate cancer incidence and a decreasing pathologist workforce coinciding with increasing requirements for more complex assessments and reporting.

Digital pathology and artificial intelligence (AI) algorithms for analysing whole slide images (WSI) show promise in improving the accuracy and efficiency of histopathological assessments. Studies have demonstrated AI's capability to diagnose and grade prostate cancer comparably to expert pathologists. However, external validations on diverse data sets have been limited and often show reduced performance. Historically, there have been no well-established guidelines for AI study designs and validation methods. Diagnostic assessments of AI systems often lack pre-registered protocols and rigorous external cohort sampling, essential for reliable evidence of their safety and accuracy.

This study protocol covers the retrospective validation of an AI system for prostate biopsy assessment. The primary objective of the study is to develop a high-performing and robust AI model for diagnosis and Gleason scoring of prostate cancer in core needle biopsies, and at scale evaluate whether it can generalise to fully external data from independent patients, pathology laboratories, and digitalisation platforms. The secondary objectives cover AI performance in estimating cancer extent and in detecting cribriform prostate cancer and perineural invasion. This protocol outlines the steps for data collection, predefined partitioning of data cohorts for AI model training and validation, model development, and predetermined statistical analyses, ensuring systematic development and comprehensive validation of the system. The protocol adheres to TRIPOD+AI, PIECES, CLAIM, and other relevant best practices.

# Table of contents

# 1. Introduction

Histopathological evaluation of prostate core needle biopsies is an important factor for prostate cancer diagnosis and treatment. Pathologists examine biopsies using the Gleason scoring system (Gleason, 1992) assigning primary and secondary grades based on the relative quantities of tissue representing different Gleason patterns (e.g. a Gleason score of 3 + 4 = 7 indicating primary Gleason pattern 3 and secondary Gleason pattern 4) (Epstein *et al.*, 2005). Grading is however inherently subjective and associated with high intra- and inter-pathologist variability placing patients at risk of inappropriate treatment selection (Melia *et al.*, 2006; Egevad *et al.*, 2013; Ozkan *et al.*, 2016). With the aim of standardisation, the International Society of Urological Pathology (ISUP) updated grading guidelines such that Gleason scores (GS) are pooled into five ordinal categories (i.e. 1 to 5) referred to as the ISUP grades (also called grade groups or WHO grade) (Ji, 2005; Epstein *et al.*, 2016; WHO Classification of Tumours Editorial Board and International Agency for Research on Cancer, 2022). Besides Gleason scoring, similar issues also hamper the reliable and repeatable assessment of other histopathological entities relevant to the clinical management of prostate cancer, such as cribriform cancer morphology (Egevad *et al.*, 2023) or perineural invasion (PNI) (Egevad *et al.*, 2021), both of which are associated with a poor prognosis.

Digital pathology (Pantanowitz *et al.*, 2018) and the application of artificial intelligence (AI) algorithms to analyse whole slide images (WSIs) hold promise for reducing variability and improving the accuracy of histopathological assessments. Many previous studies have demonstrated that AI can diagnose and grade prostate cancer on par with expert pathologists (Campanella *et al.*, 2019; Bulten *et al.*, 2020, 2022; Ström *et al.*, 2020). However, external validations demonstrating the generalisation capacity of these models on data spanning across scanning devices, laboratories, and patient populations not involved in the model development have been limited. Moreover, results from the validation studies have often shown deteriorated performance on the external data (Campanella *et al.*, 2019; Swiderska-Chadaj *et al.*, 2020; Ji *et al.*, 2023). These complications are not specific to prostate pathology, as there are several examples of scanner-induced variability and bias posing challenges for AI models across different tasks and tissue types (Howard *et al.*, 2021; Schmitt *et al.*, 2021; Duenweg *et al.*, 2023).

The unresolved issues with generalisation limit the widespread application of AI in clinical practice, including histopathology. The field has historically lacked well-established guidelines on AI study designs and standardised methods for the proper evaluation and reporting of AI validation studies. Generally, diagnostic assessments of AI systems lack pre-registered study protocols with predefined analysis plans and rigorous sampling of external cohorts, which are key factors for generating reliable evidence of the safety and diagnostic accuracy of these systems in view of further prospective evaluations in clinical trials (Nagendran *et al.*, 2020; McGenity, Bossuyt and Treanor, 2022). Here, we present a comprehensive study protocol for retrospective validation of an AI system for diagnostic assessment of prostate biopsies. This protocol outlines study objectives, analysis and experimental pipelines, as well as data cohorts for evaluating the generalisability and robustness of the AI system. The AI system is ultimately intended to be used as part of computer-aided diagnosis (CAD) software to provide decision-making support for pathologists, but the focus of the current study is on the standalone diagnostic performance of the system. Aspects relating to the clinical implementation of the system, user interaction, and analysis of the diagnostic performance of the system in combination with the supervision of a human pathologist are outside of the scope of this protocol.

Several guidelines have recently been proposed or are under development for reporting clinical validation studies of AI-based methods e.g. SPIRIT-AI (Standard Protocol Items: Recommendations for Interventional Trials) and its companion statement CONSORT-AI (Consolidated Standards of Reporting Trials), which are intended for protocols and reporting of randomised clinical trials involving an AI intervention component (Cruz Rivera *et al.*, 2020; Liu *et al.*, 2020), or the DECIDE-AI (Developmental and Exploratory Clinical Investigations of DEcision support systems driven by Artificial Intelligence) guideline which applies specifically to early, small-scale evaluation of AI interventions, with a focus on clinical utility, safety and human factors (Vasey *et al.*, 2022).

In terms of guidelines applicable to pre-clinical and offline evaluation of AI prediction models, the TRIPOD+AI (Transparent Reporting of a multivariable prediction model of Individual Prognosis Or Diagnosis) (Collins *et al.*, 2021) guideline on developing or reporting performance of AI prediction models has recently been released (Collins *et al.*, 2024), while the STARD-AI (Standards for Reporting of Diagnostic Accuracy Study) (Sounderajah *et al.*, 2021) guideline is still under development. This protocol incorporates guidelines by the TRIPOD+AI (Collins *et al.*, 2024), applicable parts of the best practice checklists proposed in PIECES (Protocol Items for External Cohort Evaluation of a Deep Learning System in Cancer Diagnostics) (Kleppe *et al.*, 2021), CLAIM (Checklist for AI in Medical Imaging) (Mongan, Moy and Kahn, 2020; Tejani *et al.*, 2023) and methodological checklists with a focus on radiology due to absence of such guidelines in the field of pathology (Park and Han, 2018). This AI study protocol covers the steps of data collection, prespecified partitioning of data cohorts, model development, and prespecified statistical analyses, ensuring systematic development and thorough validation of the system.

## 2. Study objectives

The objective of the study is to develop a high-performing and robust AI model for diagnosis and Gleason scoring of prostate cancer in core needle biopsies, and at scale demonstrate that it can generalise to fully external data from independent patients, pathology laboratories, and digitalisation platforms.

### 2.1. Primary objective

The primary objective is to assess the concordance between the AI model and pathologists in diagnosing and Gleason scoring prostate cancer in core needle biopsies.

### 2.2. Secondary objectives

There are three secondary objectives which this study accommodates:
- Assess the concordance between the AI model and pathologists in measuring cancer extent (in millimetres) in prostate core needle biopsies.
- Assess the concordance between the AI model and pathologists in detecting perineural invasion in prostate core needle biopsies.
- Assess the concordance between the AI model and pathologists in detecting cribriform cancer in prostate core needle biopsies.

## 3. Artificial intelligence system

The AI system to be developed and evaluated in this study is intended for the histopathological assessment of digitised prostate core needle biopsies. The system will be based on deep neural networks and its specific design (e.g. image preprocessing steps, model architecture and training approach) will be optimised during the study (see Section 4 for further description of the design choices and hyperparameters that will be evaluated). This study comprises multiple AI models, each tailored for the specific objectives i.e. grading, perineural invasion, cribriform cancer and cancer length and together these models integrate to form an AI system.

**System input:** A WSI stored in a supported vendor-specific format, depicting a formalin-fixed, paraffin-embedded (FFPE) haematoxylin & eosin (HE) stained prostate core needle biopsy specimen with one or several tissue cuts of one or several biopsy cores.

**System output:**
- Gleason score: the system will output GS, such as 4 + 3 = 7, indicating the primary and secondary patterns observed within the input WSI. The GS ranges from 3 + 3 = 6 to 5 + 5 = 10, with lower scores representing less aggressive cancer and higher scores indicating more aggressive cancer. Benign samples are encoded as 0 + 0.
- ISUP grade: the system will output an ISUP grade which groups GS into ordinal categories, ranging from 1 to 5. The GS are expressed as ISUP grades as follows: ISUP 1 (GS 6), ISUP 2 (GS 3 + 4 = 7), ISUP 3 (GS 4 + 3 = 7), ISUP 4 (GS 8), ISUP 5 (GS 9 - 10). Benign samples are encoded as 0.
- Cancer extent: the system will quantify the extent of cancer within the provided WSI in millimetres. This measurement indicates the size of the cancerous area within the tissue specimen.
- Cribriform cancer: the system will output the predicted probability of cribriform prostate cancer morphology being present within the input WSI.
- Perineural invasion: the system will output the predicted probability of perineural invasion being present within the input WSI.
- Visualisation: the system will provide a visualisation of its predictions including areas of different Gleason patterns, PNI and cribriform cancer, which can be examined in a WSI viewer software overlaid on the digital slide. The exact format of the visualisation will vary depending on the viewer software.

# 4. Study design

In this study, the aim is to develop the AI system described above and validate its diagnostic performance on retrospectively collected cohorts. To carry out the study, historical data, including medical records, pathology reports, and digitised images have been collected for cases where both the AI system and human pathologists make diagnostic assessments. The study design involves two independent phases: AI system development and AI system validation as shown in Fig. 1. The development phase involves an iterative cycle of refining the model design and hyperparameters using predefined development and tuning cohorts for model training and estimation of the effects of design choices on diagnostic performance. Once the overall performance on the development and tuning sets is deemed to have reached a plateau and further changes to the model design no longer yield meaningful improvements, a design freeze will take place and the final AI model will be graduated to the validation phase. This design achieves complete isolation between the model development and the retrospective validation to avoid any information leakage, which could lead to overly optimistic validation results. All model parameters and hyperparameters, including selection of any classifier thresholds, will be set based on the development and tuning cohorts and no adjustments or tweaking will be conducted on the validation cohorts, which will remain entirely untouched during the development phase.

The development cohorts provide a wide representation of tissue morphologies, scanning devices, laboratories, and clinical characteristics of patients, allowing for the training of a robust model. The tuning cohorts enable assessing model generalisation (i.e. performance on data from different laboratories and scanners than the development cohorts) on each development iteration, and direct performance comparison with state-of-the-art models evaluated on these same datasets in earlier studies (Ström *et al.*, 2020; Bulten *et al.*, 2022). Sequential experiments will be conducted one modification at a time to evaluate e.g. different preprocessing approaches for extracting image data from the WSI, deep neural network architectures, optimiser hyperparameters etc. (see Supplementary Appendix Section 1). Model performance at each step will be measured using cross-validation on the development cohorts and independent evaluation on the tuning cohorts. To accelerate the development process by reducing runtime for early model designs and to simplify troubleshooting, we will initially only use one of the development cohorts for model training and gradually introduce the other development cohorts one by one. This approach to model development allows:

- Effective troubleshooting: systematic experiments facilitate easier debugging and identification of error root causes.
- Traceability and accountability: transparency and traceability of how the model evolved during development, and accountability in cases of improvements or issues.
- Isolation of changes: the impact of each modification is assessed independently without the confounding effects of simultaneous changes (e.g. changing multiple hyperparameters at once).
- Optimal model tuning: controlled and sequential modifications allow for optimal tuning of the model and achieving the best possible model performance.

The validation phase will employ a blinded approach, wherein neither the pathologists nor the AI model have access to each other's assessments. The validation cohorts consist of samples representing a range of heterogeneous clinical settings and were collected from patients not included in the development or tuning cohorts. They are categorised as internal (scanner and laboratory included in the development set), partly external (scanner included in the development set) or fully external (neither scanner nor laboratory included in the development set) depending on the slide scanners and clinical laboratories involved. Internal validation can be expected to provide an optimistic estimate of the diagnostic performance of the AI model in the absence of laboratory or scanner variation. The generalisation performance of the model will ultimately be evaluated on the external validation cohorts, which avoids any optimistic bias. The design also allows for additional validation cohorts to be added at any point after the development phase.

Due to inter-observer variability among pathologists, reference standards established by pathologists vary across different validation cohorts. This complicates the assessment of the AI model for generalisation across cohorts, as any differences in observed performance can be partly attributed to differences in reference standards and partly attributed to imperfect AI generalisation to data originating from different clinical sites. In the case of the primary study objective of Gleason scoring, we have addressed this issue by having a representative subset of slides from each cohort be re-assessed by the lead pathologist (L.E.). The lead pathologist is highly experienced in urological pathology and has shown high concordance relative to other experienced uropathologists in several studies (Kweldam *et al.*, 2016; Egevad *et al.*, 2017; Bulten *et al.*, 2022). For the secondary study objectives of cribriform cancer and perineural invasion detection, the assessments were conducted either by the lead pathologist or by other experienced (>25 years of clinical experience after residency) uropathologists (B.D., H.S.) whose concordance with the lead pathologist has been quantified in earlier studies (Egevad *et al.*, 2021, 2023). This provides a consistent reference standard which will allow us to assess the technical generalisation performance of the model (without complete confounding between laboratory, scanner, and pathologist reference standards), in addition to large-scale evaluation relying on the varying reference standards provided by the local pathologists for each cohort.

Clinical and pathological characteristics of the included patients are summarised in Table 1 and detailed information on the slide scanning is provided in Table 2. Details on reference standards for each cohort with respect to grading are given in Table 3, and with respect to cribriform cancer and PNI are given in Table 4. Information on slides representing morphological subtypes is given in Table 5, and the number of slides for which immunohistochemistry (IHC) staining was performed in order to confirm the diagnosis is tabulated in Table 6. See Supplementary Appendix Section 3 for CONSORT diagrams summarising the data cohorts.

## 5. Inclusion and exclusion criteria

Provided below are the detailed criteria used to assess the eligibility of patients, individual biopsy slides, or WSIs for inclusion in this study.

### 5.1. Inclusion criterion
- Patients who underwent a prostate core needle biopsy were eligible.

## 5.2. Exclusion criteria

- Clinical information:
    a) Patients with either slides or associated pathology information unavailable.
    b) Slides lacking identifiers (IDs) preventing linkage to the pathology data.
    c) Slides with identical IDs preventing unambiguous linkage to the pathology data.
    d) Slides with mismatching GS and ISUP grade information.
    e) Slides with mismatching information concerning malignancy and GS or ISUP grade (e.g. indicated benign but a GS is provided).
    f) Slides with partial or erroneous GS reporting (e.g. <6, 4 + 0 or 1 + 1 etc.).
- Staining and slide preparation:
    a) Samples not containing prostate tissue e.g. bladder biopsies, testicular biopsies.
    b) Samples not stained with HE (e.g. IHC stains).
    c) Initial cuts of tissue blocks deemed unsuitable by the pathologist for providing a diagnosis and requiring a recut.
    d) Empty biopsy slides with no tissue on the glass.
- Slide integrity and annotation:
    a) Slides with pen mark annotations that cover a vast amount of the tissue, obscuring the underlying morphology.
    b) Slides with pen mark annotations conflicting with the pathology diagnosis (e.g. there exists a pen mark annotation on the slide, but the slide is diagnosed as benign or vice versa). This only applies to the STHLM3 cohort (see Section 7.1), where the pen mark annotation process is known to be consistent for all samples.
    c) Slides with pen mark annotations that result in the majority of the tissue being out of focus during scanning.
- Slide digitisation:
    a) Earlier scans of the same slide on the same scanner instrument, assuming the latest WSI represents a successful rescanning due to e.g. earlier focus issues.
    b) Corrupt WSI files which cannot be accessed with Openslide (Goode *et al.*, 2013) or OpenPhi (Mulliqi *et al.*, 2021).

# 6. Data partitions

## 6.1. Requirements for data partition

We established a number of requirements to guide the inclusion, exclusion and partitioning of data into development, tuning and validation sets to account for several sources of potential bias in the training and validation of the model. We followed available guidelines and criteria for balanced and representative data partitions (Park and Han, 2018; Mongan, Moy and Kahn, 2020; Willemink *et al.*, 2020; Varoquaux and Cheplygina, 2022) and arrived at the following set of requirements:

1. Representative sample selection: Ensure the data are representative of the diversity encountered in clinical practice by including multi-site cohorts with variations in scanning equipment (e.g. vendors, models, image formats), biopsy preparation (e.g. staining, tissue cutting), morphological heterogeneity (e.g. different Gleason scores and rare cancer subtypes) and patient demographics.
2. Representative sample size: Include a sufficiently large sample for development and validation to increase the probability of generalisability in the larger population.
3. Mitigate overfitting due to observer bias: Alleviate the possibility of overfitting or "over tweaking" of the model, which may be caused by excessive refinement of the model design aimed at maximising cross-validation performance in development data, since that can jeopardise generalisation outside the development cohorts. The issue can be mitigated by additional (external) tuning data cohorts serving as a less biased performance indicator during model development. It should be further ensured that the tuning cohorts are independent of model training (for example, criteria for early stopping of model training should be assessed only on the development data).
4. Ensure independence of specimens between data partitions: Each data partition (development, tuning, internal or external validation sets) should be independent of the others with no overlap of biopsies or patients.
5. Ensure independence of sample preparation process between data partitions: Sample external cohorts such that there is no overlap with respect to the clinical laboratories that prepared these cohorts and the development cohorts.
6. Ensure independence of the digitisation process between data partitions: Sample external cohorts such that there is no overlap with respect to the scanning device used for these cohorts and the development cohorts.

## 6.2. Predefined data partitions

The process of splitting the data cohorts into development, tuning, and internal and external validation sets was conducted adhering to the requirements for data partitions and is described below (see Fig. 1 for an overview). The characteristics of the data cohorts included in this study are summarised in Table 1 and described in detail in Section 7.

***The development set*** was sampled from the following cohorts: Capio S:t Göran Hospital (STG), Radboud University Medical Center (RUMC), Stavanger University Hospital (SUH) and Stockholm3 (STHLM3). From the RUMC, STHLM3 and SUH cohorts, the patients who were not allocated to tuning or validation sets (see below) were assigned to the development set (approximately 80% of patients). Given the limited size and skewed grade distribution of the STG cohort, it was fully allocated into the development set. The development set covers several clinical laboratories and scanner devices as well as a large degree of variation in tissue morphology and the clinical characteristics of patients, in part due to the largest cohort, STHLM3, originating from a population-based screening trial (Requirements 1-2). Each of the development cohorts was further split into 10 cross-validation folds by randomly allocating patients to folds, stratified by the maximum slide level ISUP grade of each patient.

***The tuning set*** was sampled from the following cohorts: Karolinska University Hospital (KUH-1), RUMC and STHLM3. The entire KUH-1 cohort was assigned to tuning and represents a fully external cohort relative to the development set (i.e. different patients, laboratory and scanner). This set also corresponds to the European external validation cohort of the PANDA challenge (Bulten *et al.*, 2022). The subsets of the RUMC and STHLM3 cohorts assigned to the tuning set represent internal data relative to the development set (i.e. different patients but the same laboratories and scanners) and correspond to the PANDA public test sets in Kaggle (i.e. the PANDA tuning sets). The tuning sets allow for evaluating the effects of model design changes on data that is independent of the development set, direct comparison with state-of-the-art models from PANDA, and in the case of KUH-1, assessing the generalisation performance of the model prior to design freeze (i.e. performance on data coming from different patients, laboratories, and scanners compared to the development data) (Requirement 3). A subset of slides belonging to the PANDA Swedish tuning set was allocated to the internal validation set for reasons related to patient stratification and the inclusion of specific subsets of interest in the internal validation (see below).

*The internal validation set* was sampled from the following cohorts: RUMC, STHLM3 and SUH, consisting of patients who were not part of the development or tuning sets but whose biopsies were obtained from the same clinical laboratories and scanned with the same scanners as the development and tuning set samples. The STHLM3 internal validation set includes the following subsets, supplemented with randomly sampled patients to achieve a total 20% fraction of patients assigned to tuning and validation: ImageBase (Egevad *et al.*, 2017), Swedish private test set in Kaggle (i.e. PANDA Swedish internal validation set) (Bulten *et al.*, 2022), perineural invasion multi-observer validation set (Kartasalo *et al.*, 2022), and rare morphological subtypes set (Olsson *et al.*, 2022). Including these samples as subsets of the internal validation set will facilitate (internal) comparisons with results obtained in the papers referenced in the preceding sentence. The SUH internal validation set includes the following subsets, supplemented with randomly sampled patients to achieve a 20% fraction of patients assigned to validation: all patients (n=25) with multiple recuts of their biopsy tissue blocks, and patients (n=81) corresponding to a random selection of 119 slides stratified on ISUP grade (to be rescanned repeatedly over time for an AI temporal stability study). The STHLM3 subsets allocated into the internal validation set were selected based on being particularly valuable for the evaluation phase of the study, while the SUH subsets will be used as validation sets in upcoming follow-up studies involving the AI model developed here, hence cannot be assigned to the development set. The RUMC internal validation set includes the RUMC private test set in Kaggle (i.e. PANDA RUMC internal validation set) (Bulten *et al.*, 2022), supplemented with randomly sampled patients to achieve a total 20% fraction of patients assigned to tuning and validation.

*External validation cohorts* are fully external relative to the development set (no overlap with respect to patients, laboratory, or scanner) or partly external (no overlap with respect to patients or laboratory, but digitisation performed using a scanner that is present also in the development set). Fully external validation set cohorts include Aarhus University Hospital (AUH), Karolinska University Hospital morphological subtypes (KUH-2), Mehiläinen Länsi-Pohja (MLP), Medical University of Lodz (MUL), Synlab Switzerland (SCH), Synlab Finland (SFI), Synlab France (SFR), Spear Prostate Biopsy 2020 (SPROB20), University Hospital Cologne (UKK), Hospital Wiener Neustadt (WNS). Partly external validation set cohorts include: Aquesta Uropathology morphological subtypes (AQ), partially scanned on a scanner present in the development set and partially scanned on an external scanner. The external nature of the validation set cohorts fulfils Requirements 4-6.

All data splits were performed on patient level, that is, all slides and resulting WSIs from a given patient were allocated to the same data partition in order to avoid information leakage between development and validation sets. If a patient was biopsied on several occasions, all biopsies were included and allocated together. Any samples lacking patient identifiers were assigned to development data to avoid the risk of slides from any patients ending up in both development and evaluation cohorts.

Subsets of the slides included in this study have been scanned multiple times. If the same slide had been rescanned multiple times on the same individual scanner (i.e. the same physical device), we only kept the WSI with the latest scanning date, assuming the rescanning was due to e.g. initially poor focus or other scanning issues. Subsets of the STG, STHLM3 and MUL cohorts were rescanned with multiple different scanners (see Table 2). To avoid biasing the evaluation towards these slides that appear in the dataset multiple times, we will only include one WSI per slide in the validation sets. For STHLM3, we will randomly select one WSI for each slide to be evaluated, and for MUL, we will utilise WSIs from the Grundium Ocus40 scanner, excluding those on the Philips UFS scanner. This ensures that the MUL cohort remains entirely external relative to the development data, considering that the STHLM3 cohort was partly digitised on the same Philips UFS instrument. The repeated scans will, however, be used during AI model development as an augmentation technique (except for the Grundium Ocus40 which is kept as an external scanner for validation), and for a separate cross-scanner reproducibility analysis (see Section 8).

# 7. Data cohorts

## 7.1. Development, tuning and internal validation data cohorts

### 7.1.1. Karolinska University Hospital (KUH-1)

The KUH-1 samples were collected at the Department of Pathology, Karolinska University Hospital in Solna, Sweden in 2018. Among the cases assessed by L.E. during 2018, we included all positive slides of all patients diagnosed with ISUP grade 2-5 cancer, all positive slides from a random selection of patients diagnosed with ISUP grade 1 cancer, and all slides from a random selection of patients with a negative diagnosis. Patients underwent systematic transrectal biopsies in approximately 1/3 of the cases, and magnetic resonance imaging (MRI) targeted or combined biopsies in approximately 2/3 of cases. Slides typically contain one core, sectioned at two levels. This cohort has been used as an external validation set in previous studies (Ström *et al.*, 2020; Bulten *et al.*, 2022).

#### 7.1.1.1. Reference standard protocol

All cases were assessed by the lead pathologist (L.E.) using a microscope to determine the GS and cancer extent per slide, as well as the ISUP grade per slide and per patient. The linear cancer extent was generally measured from end to end in cases with discontinuous cancer and it was reported on a per-cut level.

### 7.1.2. Radboud University Medical Center (RUMC)

The RUMC samples were collected at the Radboud University Medical Center in Nijmegen, the Netherlands from January 2012 to December 2017 (Bulten *et al.*, 2020). Patients were sampled randomly, stratified by the highest reported GS in the pathology reports, and the slide with the most aggressive part of the tumour was included for each patient. Additionally, a group of patients with only benign biopsies were randomly sampled. Patients generally underwent MRI-targeted transrectal biopsy. The data underwent additional refinement in preparation for the PANDA Kaggle challenge (Bulten *et al.*, 2022): only one core, sectioned at one level was retained per WSI, the background was masked to hide most of the markings made on the glass, and the images were converted into .tiff format (JPEG compression, quality 70). For the purposes of PANDA, the cohort was partitioned into three sets—development, tuning, and internal validation, stratified by patient and the highest Gleason pattern in the biopsy.

#### 7.1.2.1. Reference standard protocol

The reference standard for all cases on the RUMC development set was determined based on the original pathology reports. Due to each slide containing multiple biopsy cores, trained non-experts digitally outlined the individual cores, allowing them to be partitioned into separate WSIs, and assigned core-level GS based on the pathology reports. Inconclusive pathology reports were assigned for a second review, and if no match could be made these cases were discarded (Bulten *et al.*, 2020).

Subsets of the cohort underwent additional re-assessments as follows:
- The PANDA RUMC tuning set (n=195, corresponds to our RUMC tuning set) and the PANDA RUMC internal validation set (n=333, part of our RUMC internal validation set) were assessed in three rounds. In the first round, three uropathologists individually graded the cases digitally, providing a GS per slide. A majority vote was taken for cases where an agreement was reached on the ISUP grade but there was a discrepancy in the Gleason patterns, and cases where two uropathologists agreed and the third one had a maximum deviation of one ISUP grade. In the second round, all the cases that did not achieve consensus were re-graded by the uropathologist whose grade differed from the others, followed by pooling of all the assessments and discussion in a consensus meeting in the third round. The GS was reported per slide.
- A subset of slides (n=66) from the RUMC internal validation cohort was randomly selected, stratified by the ISUP grade, for re-assessment by the lead pathologist (L.E.). This re-assessment was conducted digitally on Cytomine (Marée *et al.*, 2016) using 3DHISTECH WSIs (.mrxs converted to .tiff) to report the GS per slide.

### 7.1.3. Capio S:t Göran Hospital (STG)

The STG samples were collected at Capio S:t Göran Hospital in Stockholm, Sweden from 2016 to 2017. We included a random selection of slides with an enrichment for high-grade cancer. Patients underwent transrectal biopsy, and slides typically contain one core, sectioned at two levels. This cohort was also part of the development set in a previous study (Ström *et al.*, 2020).

#### 7.1.3.1. Reference standard protocol

All cases were assessed by the lead pathologist (L.E.) using a microscope to provide GS, ISUP grade, and cancer extent on a per-slide level. The linear cancer extent was generally measured from end to end in cases with discontinuous cancer and it was reported on a per-cut level.

### 7.1.4. Stockholm3 (STHLM3)

The STHLM3 samples were collected in a population-based clinical trial (ISRCTN84445406) (Grönberg *et al.*, 2015) from 2012 to 2015 in Stockholm, Sweden. Histological sample preparation was performed at Histocenter, Gothenburg, Sweden, and the samples were assessed at the Department of Pathology, Karolinska University Hospital in Stockholm. Patients underwent 10-12 core systematic transrectal biopsies and slides usually contain one core, sectioned at two levels. Subsets of the digitised samples have been used as development and internal validation sets in previous studies (Ström *et al.*, 2020; Bulten *et al.*, 2022; Ji *et al.*, 2022; Kartasalo *et al.*, 2022; Olsson *et al.*, 2022). Patient and slide selection, retrieval and digitisation took place on five occasions between 2014 and 2023 (see Table 2), as below:
- 2014: All cores from the first 500 patients diagnosed with prostate cancer in the STHLM3 trial were scanned on a Hamamatsu NanoZoomer 2.0-HT.
- 2017-2019: All patients with at least one core graded as GS 4 + 4 or 5 + 5 and 497 randomly selected patients with at least one core graded as 3 + 3 were considered. From each of these patients, we included all positive cores and a randomly selected negative core. Finally, we randomly selected 139 cancer-free patients from whom we included one randomly selected core. Additionally, we added all cores which were indicated to have PNI and had not been scanned earlier. The cores were scanned on an Aperio AT2.

- 2018-2019: The cores of a random selection of patients were scanned on a Hamamatsu NanoZoomer XR.
- 2019-2020: The cores of a random selection of patients were scanned on the Philips IntelliSite Ultra Fast Scanner (UFS).
- 2023: Patients belonging to the PANDA challenge Swedish public and private validation sets were scanned on the Grundium Ocus40.
- 2023: Initially, cores with < 4 millimetres of cancer were excluded to have sufficient cancer tissue for future molecular profiling of the samples. Among the remaining patients, 50% of those with ISUP 1 or ISUP 2 (patient level ISUP) were randomly selected for inclusion, while all patients with ISUP 3-5 were included for scanning on the Grundium Ocus40.

7.1.4.1.  Reference standard protocol

All cases were assessed by the lead pathologist (L.E.) using a microscope to obtain the GS, the ISUP grade, cancer extent and PNI on a per-slide level. The linear cancer extent was generally measured from end to end in cases with discontinuous cancer and reported on a per-cut level. However, in cases with 1 or 2 cores infiltrated by low-grade discontinuous cancer with a benign gap exceeding 3 millimetres, the benign tissue was subtracted in the reporting of total cancer extent.

Subsets of the cohort underwent additional re-assessments as follows:
- A subset of slides (n=212) from the STHLM3 internal validation cohort underwent a second review to construct a reference standard for the PANDA Swedish internal validation set. Slides initially indicated as benign according to the original reference standard were not re-reviewed, while cases indicated as malignant were divided between two uropathologists (B.D. and H.S.), each reviewing 100 slides blinded to the original review. In the case of agreement between the initial and the second review, the consensus ISUP grade was assigned to the case. In case of disagreement, a third uropathologist (T.T.) reviewed the case. For cases that were indicated as malignant by all pathologists, the final ISUP grade was assigned according to 2/3 consensus. If all three reviews were in disagreement, the case was excluded from the internal validation set. Any cases indicated as benign in the second or third review were excluded from the PANDA Swedish internal validation set. The re-assessment was conducted digitally on Cytomine using Hamamatsu and Aperio WSIs (.ndpi and .svs converted to .tiff) as described in (Bulten *et al.*, 2022).
- A subset of slides (n=24) from the STHLM3 internal validation cohort was additionally assessed by the lead pathologist (L.E.) for specific rare morphologies (see Table 5) using a microscope. This set has been used as validation data in a previous study (Olsson *et al.*, 2022).

- A subset of slides (n=87) from the STHLM3 internal validation cohort, representing the ImageBase set (Egevad *et al.*, 2017) was additionally assessed by an expert panel of uropathologists (n=23). The assessment was conducted using digital micrographs. This set has been previously used in the study (Ström *et al.*, 2020) as an internal validation set.
- A subset of slides (n=702) from the STHLM3 development and internal validation cohorts was digitally assessed for cribriform cancer as described in (Egevad *et al.*, 2023). To arrive at this selection, we first enriched Gleason pattern 4 tissue by randomly selecting one core per combination of patient and ISUP grade among all cores with ISUP grades 3-5. To maintain some representation of GS 3+4 biopsies, we randomly selected 86 additional cores with one core per patient from the set of all cores with ISUP grade 2. The slides were assessed by the lead pathologist (L.E.) on Cytomine using Hamamatsu (.ndpi) and Aperio (.svs) WSIs to create pixel-wise annotations of areas with cribriform cancer. The pathologist could also indicate uncertain cases with a borderline category.
- A subset of slides positive for cribriform cancer (n=152) and a random selection of slides negative for cribriform cancer (n=152) according to the assessment by L.E. were additionally assessed by an expert panel of uropathologists (n=9) as described in (Egevad *et al.*, 2023). The pathologists assessed the presence of cribriform cancer on slide level on Cytomine using Hamamatsu (.ndpi) and Aperio (.svs) WSIs. The pathologists were blinded to the distribution of positive or negative slides and to each other's assessments.
- All slides positive for PNI (n=485) in the STHLM3 development and internal validation cohorts were digitally re-assessed as described in (Kartasalo *et al.*, 2022). The slides were assessed by the lead pathologist (L.E.) in QuPath (Bankhead *et al.*, 2017) using Hamamatsu (.ndpi) and Aperio (.svs) WSIs to create pixel-wise annotations of areas of PNI.
- A subset of slides positive for PNI (n=106) and a random selection of slides negative for PNI (n=106) according to the assessment by L.E. was additionally assessed by an expert panel of uropathologists (n=4) as described in (Egevad *et al.*, 2021). The pathologists assessed the presence of PNI on slide level on Cytomine using Hamamatsu (.ndpi) and Aperio (.svs) WSIs. The pathologists were blinded to the distribution of positive or negative slides and to each other's assessments. The pathologists could also indicate uncertain cases with borderline categories.

### 7.1.5. Stavanger University Hospital (SUH)

The SUH samples represent consecutive cases collected from routine diagnostics at the Department of Pathology, Stavanger University Hospital in Stavanger, Norway from December 2016 to March 2018. Biopsies were taken at the Department of Urology in Stavanger University Hospital and other private urological clinics at the Stavanger Urological Center. Patients primarily underwent systematic transrectal biopsies, although some received MRI-targeted biopsies, either alone or combined with systematic biopsy. Slides typically contain two cores from the same anatomical location, sectioned at two levels. A subset of the SUH cohort has been used as an external validation set in previous studies (Ji *et al.*, 2022; Olsson *et al.*, 2022).

7.1.5.1.   Reference standard protocol

The reference standard was obtained from the original pathology reports from the clinical routine. Seven uropathologists and seven general pathologists assessed the slides microscopically reporting the GS, ISUP grade, Gleason pattern 4 percentage, cancer extent, biopsy length, PNI, fatty tissue infiltration (FTI), and additional stainings (e.g. IHC) on the slide level. The linear cancer extent was generally measured from end to end in cases with discontinuous cancer and it was reported on a per-cut level.

Subsets of the SUH cohort underwent additional re-assessments as follows:
- A subset of slides (n=66) from the SUH internal validation cohort was randomly selected and stratified by ISUP grade for re-assessment by the lead pathologist (L.E.). This re-assessment was conducted digitally on Cytomine using Hamamatsu WSIs (.ndpi) to report the GS per slide.
- A subset of slides (n=332) with Gleason pattern 4 tissue from the SUH development and internal validation cohorts was initially assessed by a uropathologist (A.B.) for potential cribriform cancer using QuPath. We then randomly selected at most 90 positive, 30 borderline and 30 negative slides from the development cohort and at most 30 positive, 10 borderline and 10 negative slides from the internal validation cohort to be re-assessed by the lead pathologist (L.E.), resulting in 200 slides. This re-assessment was conducted digitally on Cytomine using Hamamatsu (.ndpi) WSIs to report cribriform cancer per slide. The pathologist could also indicate uncertain cases with a borderline category.
- All slides from cases reported as positive for PNI in the SUH development and internal validation cohorts were initially assessed by a uropathologist (A.B.) for potential PNI using a microscope. We then randomly selected at most 25 positive and 5 negative slides per ISUP grade from the development cohort, and at most 8 positive and 2 negative slides per ISUP grade from the internal validation cohort to be re-assessed by the lead pathologist (L.E.), resulting in 185 slides. This re-assessment was conducted digitally on Cytomine using Hamamatsu (.ndpi) WSIs to report PNI per slide. The pathologist could also indicate uncertain cases with a borderline category.

## 7.2. External validation cohorts

### 7.2.1. Aichi Medical University (AMU)

The AMU samples were collected at the Aichi Medical University in Nagakute, Japan from 2020 to 2023. Samples were selected to include cribriform prostate cancer cases and non-cribriform cases. Cribriform cases were chosen sequentially, while non-cribriform cases were selected among cases containing Gleason pattern 4 and age-adjusted to match the cribriform cases. Patients generally underwent systematic transrectal biopsy, with only a few undergoing MRI-targeted biopsy. Slides typically contain several cores, sectioned at several levels.

#### 7.2.1.1. Reference standard protocol

All cases were assessed by a uropathologist (T.T.) initially using a microscope and then confirmed digitally with the NDP.View software using Hamamatsu WSIs (.ndpi). The presence or absence of cribriform prostate cancer was reported on slide level and GS was reported on patient level.

### 7.2.2. Aquesta Uropathology morphological subtypes (AQ)

The AQ cases were collected at the Aquesta Specialised Uropathology laboratory in Toowong, Australia from 2009 to 2023. The biopsies were performed in private hospitals and urology clinics in Queensland state, Australia. Slides were specifically selected to represent rare morphologies such as benign mimickers of prostate cancer which are typically hard to diagnose in routine pathology. Patients generally underwent MRI-targeted transrectal biopsies, and each slide has two cores, sectioned at two levels.

#### 7.2.2.1. Reference standard protocol

A uropathologist (H.S.) assessed the slides microscopically and reported the GS, ISUP grade, additional stainings (e.g. IHC), and the presence or absence of specific morphological subtype categories on slide level (see Table 5). Slides representing benign mimickers were microscopically re-assessed by the lead pathologist (L.E.).

### 7.2.3. Aarhus University Hospital (AUH)

The AUH samples were part of the PRIMA clinical trial conducted at the Aarhus University Hospital in Aarhus, Denmark from January 2018 to December 2021 (Fredsøe *et al.*, 2023). Histopathology assessment was conducted at the Department of Pathology, Aarhus University Hospital, Aarhus, Denmark. In this trial, men aged 50-59 years with elevated prostate-specific antigen (PSA) (3-10 ng/ml) and/or positive STHLM3 test (defined as STHLM3 score equal to or above 11%) and MRI of PIRADS 3-5 were referred to MRI-targeted transrectal biopsy. Out of 117 patients who underwent the biopsy procedure, the pathologist selected slides based on histopathological features with the aim of a uniform distribution of ISUP grades. Slides typically contain two cores, sectioned at three levels. This cohort was used as an external validation set in a previous study (Ji *et al.*, 2022).

#### 7.2.3.1. Reference standard protocol
All cases were assessed by a uropathologist (B.P.U.) microscopically and the GS, the ISUP grade, cancer extent and biopsy length were reported on the slide level.

Subsets of the AUH cohort underwent additional re-assessments as follows:
- A subset of slides (n=41) was randomly selected, stratified by the ISUP grade, for re-assessment by the lead pathologist (L.E.). This re-assessment was conducted digitally on Cytomine using Hamamatsu WSIs (.ndpi) to report the GS per slide.

### 7.2.4. Karolinska University Hospital morphological subtypes (KUH-2)
The KUH-2 samples were collected at the Department of Pathology, Karolinska University Hospital in Solna, Sweden in 2022. The biopsy procedure and number of tissue sections per slide adhere to the KUH-1 cohort. Similarly to the AQ cohort, these samples were specifically selected to represent cases that are typically challenging to diagnose in clinical practice, such as rare disease morphologies and benign mimickers. This cohort was used as an external validation set in a previous study (Olsson *et al.*, 2022).

#### 7.2.4.1. Reference standard protocol
The reference standard protocol for the KUH-2 cohort adheres to KUH-1, except for additional reporting of the presence or absence of specific morphological subtype categories, assessed by the lead pathologist (L.E.) on slide level (see Table 5).

### 7.2.5. Mehiläinen Länsi-Pohja (MLP)
The MLP samples represent consecutive cases from routine pathology at the Mehiläinen Länsi-Pohja Hospital in Kemi, Finland from 2016 to 2019. Patients underwent systematic transrectal biopsies, and biopsies were sampled based on anatomical location: left and right typically consisting of six cores per location. Slides typically contain one core, sectioned at two to three levels.

#### 7.2.5.1. Reference standard protocol

The reference standard was obtained from routine assessments done by several pathologists using a microscope to determine the GS, the ISUP grade, cancer extent and biopsy length per patient or per anatomical location (i.e. a set of biopsy cores assessed together).

Subsets of the MLP cohort underwent additional re-assessments as follows:
- A subset of slides (n=66) was randomly selected, stratified by the ISUP grade, for re-assessment by the lead pathologist (L.E.). The patient level ISUP grade was used for stratification, due to missing slide level grading. This re-assessment was conducted digitally on Cytomine using 3DHISTECH WSIs (.mrxs) to report the GS per slide.

### 7.2.6. Medical University of Lodz (MUL)

The MUL samples represent consecutive cases from routine pathology at the 1st Department of Urology, University Clinical Hospital of the Military Academy of Medicine - Central Veterans Hospital, Medical University of Lodz, Lodz, Poland from January 2018 to March 2019. Histopathological assessment was conducted at the Department of Pathology, Department of Oncology, Medical University of Lodz, Lodz, Poland. Patients underwent systematic transrectal biopsy and slides typically contain one core, sectioned at four to seven levels.

#### 7.2.6.1. Reference standard protocol

The reference standard was determined based on an initial assessment by a single pathologist (M.B.) and a second review by a more experienced pathologist (R.K.). Both pathologists have a specialisation in surgical pathology and are currently specialising in uropathology. The pathologists assessed the cases using a microscope and reported the GS, the ISUP grade, total cancer percentage and Gleason pattern 4 and 5 percentages on the slide level.

Subsets of the MUL cohort underwent additional re-assessments as follows:
- A subset of slides (n=66) was randomly selected, stratified by ISUP grade, for re-assessment by the lead pathologist (L.E.). This re-assessment was conducted digitally on Cytomine using Grundium WSIs (.svs) to report the GS per slide.
- All slides containing Gleason pattern 4 (n=276) were initially assessed for potential cribriform cancer by a uropathologist (A.B.). The assessment was conducted digitally on Cytomine using Grundium WSIs (.svs) to report cribriform cancer per slide and mark the positive and borderline foci. All foci were then re-assessed on Cytomine by the lead pathologist (L.E.).

- The slides (n=276) assessed for cribriform cancer were also initially assessed for potential PNI by a uropathologist (A.B.). The assessment was conducted on Cytomine using Grundium WSIs (.svs) to report PNI per slide and mark the positive and borderline foci. All foci were then re-assessed on Cytomine by the lead pathologist (L.E.).

### 7.2.7. Synlab Switzerland (SCH)

The SCH samples represent consecutive cases from routine diagnoses at the Argot Laboratory in Lausanne, Switzerland from January 2020 to December 2020. Patients underwent systematic, MRI-targeted or combined transrectal biopsies. Slides typically contain one core, sectioned at two levels. A varying number of cores were typically obtained from a varying number of anatomical locations.

#### 7.2.7.1. Reference standard protocol

The reference standard was determined based on the pathology reports from routine diagnostics. Using the microscope the pathologists reported the GS, the ISUP grade, cancer extent, biopsy length, Gleason pattern 4 percentage, cribriform cancer, PNI, high-grade prostatic intraepithelial neoplasia (HGPIN) and possible IHC staining per anatomical location (i.e. a set of biopsy cores assessed together) and per patient.

Subsets of the SCH cohort underwent additional re-assessments as follows:
- A subset of slides (n=72) was randomly selected, stratified by the ISUP grade and anatomical location for re-assessment by the lead pathologist (L.E.). This re-assessment was conducted digitally on Cytomine using Philips WSIs (.isyntax converted to .tiff) to report the GS per slide.
- A subset of slides (n=56) were digitally re-assessed for cribriform cancer by a uropathologist (H.S.). We selected all positive anatomical locations and a random selection of 6 negative anatomical locations with Gleason pattern 4 tissue and included all slides from these locations. This re-assessment was conducted digitally on Cytomine using Philips WSIs (.isyntax converted to .tiff) to report cribriform cancer per slide. The pathologist could also indicate uncertain cases with a borderline category.
- A subset of slides (n=94) were digitally re-assessed for PNI by a uropathologist (B.D.). We randomly selected 12 positive and 5 negative anatomical locations per ISUP grade and included all slides from these locations. This re-assessment was conducted digitally on Cytomine using Philips WSIs (.isyntax converted to .tiff) to report PNI per slide. The pathologist could also indicate uncertain cases with a borderline category.

### 7.2.8. Synlab Finland (SFI)

The SFI samples represent consecutive cases from routine diagnostics at the Synlab Laboratory in Helsinki, Finland from January 2020 to February 2021. Patients underwent systematic, MRI-targeted or combined transrectal biopsies. Slides typically contain two cores, sectioned at five to six levels. A varying number of cores were typically obtained from a varying number of anatomical locations.

#### 7.2.8.1. Reference standard protocol

The reference standard was determined based on the pathology reports from routine diagnostics. Using the microscope the pathologists reported the GS, the ISUP grade, cancer extent, biopsy length, Gleason pattern 4 percentage, cribriform cancer, PNI, HGPIN and possible IHC staining per anatomical location (i.e. a set of biopsy cores assessed together) and in some cases per patient.

Subsets of the SFI cohort underwent additional re-assessments as follows:
- A subset of slides (n=67) was randomly selected, stratified by the ISUP grade and anatomical location for re-assessment by the lead pathologist (L.E.). This re-assessment was conducted digitally on Cytomine using Philips WSIs (.isyntax converted to .tiff) to report the GS per slide.

### 7.2.9. Synlab France (SFR)

The SFR samples represent consecutive cases from routine diagnostics at the Technipath-Synlab Medical Laboratory in Dommartin, Rhône, France from September 2020 to December 2020. Patients underwent systematic, MRI-targeted or combined transrectal biopsies. Slides usually contain two to three cores from the same anatomical location, sectioned at two levels.

#### 7.2.9.1. Reference standard protocol

The reference standard was determined based on the pathology reports from routine diagnostics. Pathologists using a microscope reported the GS, the ISUP grade, cancer extent, biopsy length, Gleason pattern 4 percentage, cribriform cancer, PNI, HGPIN and possible IHC staining per anatomical location (i.e. slide) and in some cases per patient.

Subsets of the SFR cohort underwent additional re-assessments as follows:
- A subset of slides (n=49) was randomly selected, stratified by the ISUP grade and anatomical location for re-assessment by the lead pathologist (L.E.). This re-assessment was conducted digitally on Cytomine using Philips WSIs (.isyntax converted to .tiff) to report the GS per slide.

### 7.2.10. Spear Prostate Biopsy 2020 (SPROB20)

The SPROB20 samples were collected at Uppsala University Hospital, Uppsala, Sweden from 2015 to 2018. Patients underwent targeted transrectal biopsies. Slides typically contain one core, sectioned at one level. This cohort is publicly available at the AIDA Data Hub (Walhagen *et al.*, 2020).

#### 7.2.10.1. Reference standard protocol

The reference standard was obtained from the clinical routine. The pathologists assessed the slides microscopically and reported the ISUP grade at the patient level in two ways: as the maximum and as the average of the slide level ISUP grades. The underlying slide-level ISUP grades were not provided on the AIDA Data Hub.

Subsets of the SPROB20 cohort underwent additional re-assessments as follows:
- A subset of slides (n=50) was randomly selected, stratified by ISUP grade and patient, for re-assessment by the lead pathologist (L.E.). This re-assessment was conducted digitally on Cytomine using Hamamatsu WSIs (.ndpi converted to .tiff) to report the GS per slide.

### 7.2.11. University Hospital Cologne (UKK)

The UKK samples represent consecutive cases from the Institute of Pathology at the University Hospital Cologne in Cologne, Germany. Patients underwent combined systematic and MRI-targeted transrectal biopsies. Slides typically contain one core, sectioned at three levels. The publicly available subset of samples was randomly selected and stratified by the ISUP grade, including ten samples per ISUP grade. This cohort was obtained from a publicly available dataset which was part of the development and validation sets in an earlier study (Tolkach *et al.*, 2023). The WSIs were converted from JPEG2000 compressed OME-TIFF format via an intermediate raw Zarr format to JPEG compressed (quality 80) generic pyramidal TIFF format for OpenSlide compatibility using the *bioformats2raw* (v. 0.9.3), *raw2ometiff* (v. 0.7.1) and *libvips* (v. 8.9.1) converters.

#### 7.2.11.1. Reference standard protocol

The reference standard was determined digitally by a panel of 10 different pathologists from Austria, Germany, Israel, Japan, the Netherlands, Russia and the United States. All pathologists reported the ISUP grade per slide and the final grade was obtained as the majority vote. A consensus was considered reached in cases where the majority ISUP grade had at least six votes.

### 7.2.12. Hospital Wiener Neustadt (WNS)

The WNS samples represent consecutive cases from the Hospital Wiener Neustadt in Wiener Neustadt, Austria. Patients underwent combined systematic and MRI-targeted transrectal biopsies. Slides typically contain one core, sectioned at one level. The publicly available subset of samples was randomly selected and stratified by the ISUP grade, including ten samples per ISUP grade. This cohort was obtained from a publicly available dataset which was part of the development and validation sets in an earlier study (Tolkach *et al.*, 2023). The WSIs were converted from JPEG2000 compressed OME-TIFF format via an intermediate raw Zarr format to JPEG compressed (quality 80) generic pyramidal TIFF format for OpenSlide compatibility using the *bioformats2raw* (v. 0.9.3), *raw2ometiff* (v. 0.7.1) and *libvips* (v. 8.9.1) converters.

#### 7.2.12.1. Reference standard protocol

The reference standard was determined digitally by a panel of 11 different pathologists from Austria, Germany, Israel, Japan, the Netherlands, Russia and the United States. All pathologists reported the ISUP grade per slide and the final grade was obtained as the majority vote. A consensus was considered reached in cases where the majority ISUP grade had at least six votes.

## 8. Statistical analyses

### 8.1. Overview of statistical analyses

#### 8.1.1. Primary analysis: Diagnosis and Gleason scoring

I. Internal and external validation against the original cohort-specific reference standard
II. Subgroup analyses
    A. Evaluate performance across different age groups
    B. Evaluate performance on systematic vs. targeted biopsies
    C. Evaluate performance on non-treated patients vs. patients treated for benign prostatic hyperplasia prior to biopsy
    D. Evaluate performance on morphological subtypes
    E. Evaluate performance on cases requiring vs. not requiring IHC staining
    F. Evaluate performance compared to the current state-of-the-art AI systems
III. Sensitivity analyses
    A. Cross-scanner consistency analyses
    B. Compare the AI system vs. individual pathologist panel members

  C. Internal and external validation against uniform reference standard by the lead pathologist
  D. Blinded re-assessment of slides with marked errors

### 8.1.2. Secondary analysis: Cancer extent prediction
I. Internal and external validation against the original cohort-specific reference standards
II. Subgroup analyses
  A. Evaluate performance across different age groups
  B. Evaluate performance on systematic vs. targeted biopsies
  C. Evaluate performance on non-treated patients vs. patients treated for benign prostatic hyperplasia prior to biopsy
III. Sensitivity analyses
  A. Cross-scanner consistency analyses

### 8.1.3. Secondary analysis: Cribriform cancer detection
I. Internal and external validation against the original cohort-specific reference standards
II. Subgroup analyses
  A. Evaluate performance across different age groups
  B. Evaluate performance on systematic vs. targeted biopsies
  C. Evaluate performance on non-treated patients vs. patients treated for benign prostatic hyperplasia prior to biopsy
III. Sensitivity analyses
  A. Cross-scanner consistency analyses
  B. Compare the AI system vs. individual pathologist panel members
  E. Re-assessment excluding borderline slides

### 8.1.4. Secondary analysis: Perineural invasion detection
I. Internal and external validation against the original cohort-specific reference standards
II. Subgroup analyses
  A. Evaluate performance across different age groups
  B. Evaluate performance on systematic vs. targeted biopsies
  C. Evaluate performance on non-treated patients vs. patients treated for benign prostatic hyperplasia prior to biopsy
III. Sensitivity analyses
  A. Cross-scanner consistency analyses
  B. Compare the AI system vs. individual pathologist panel members

E. Re-assessment excluding borderline slides

### 8.1.5. Exploratory analyses
I. Evaluate visualisations of the AI output
II. Evaluate the impact of tissue segmentation algorithms
III. Evaluate end-to-end vs. transfer-learning-based models
IV. Evaluate the impact of physical colour calibration

## 8.2. Details of statistical analyses

**Primary analysis: Diagnosis and Gleason scoring**

We will quantify the concordance of the AI system's cancer diagnosis (positive/negative), Gleason score and ISUP grade with the reference standards in the tuning, internal validation and external validation cohorts using the metrics described below. The analysis will be conducted on slide level (AQ, AUH, KUH-1, KUH-2, MUL, RUMC, SFR, STHLM3, SUH, UKK, WNS), anatomical location level (MLP, SFI, SCH) and/or patient level (KUH-1, SCH, SFI, SFR, SPROB20) depending on the granularity of the available reference standards.

**Cancer diagnosis:** Sensitivity (true positive rate) and specificity (true negative rate) will be used to quantify the agreement of negative/positive diagnosis for prostate cancer with the reference standard. Confidence intervals for sensitivity and specificity will be computed using the non-parametric bootstrap over cases. We will additionally report the Area Under the Receiver Operating Characteristics Curve (AUROC) and confusion matrices.

**Gleason score:** Quadratically weighted Cohen's kappa (QWK) will be used to quantify the agreement of Gleason scoring with the reference standard. In addition, we will also report linearly weighted Cohen's kappa (LWK) and confusion matrices. To allow calculating weighted kappas, Gleason patterns (e.g. 3+4) will be encoded into ordinal variables following earlier studies (Jung *et al.*, 2022; Egevad, Micoli, Delahunt, *et al.*, 2024; Egevad, Micoli, Samaratunga, *et al.*, 2024) as follows: benign (0), 3+3 (1), 3+4 (2), 4+3 (3), 3+5 (4), 4+4 (5), 5+3 (6), 4+5 (7), 5+4 (8), 5+5 (9). Confidence intervals will be computed using the non-parametric bootstrap over cases.

**ISUP grade:** Quadratically weighted Cohen's kappa (QWK) will be used to quantify the agreement of the ISUP grade with the reference standard. In addition, we will also report linearly weighted Cohen's kappa (LWK) and confusion matrices. To allow calculating weighted kappas,

ISUP grades will be treated as ordinal variables (0-5), with benign encoded as 0. Confidence intervals will be computed using the non-parametric bootstrap over cases.

### Secondary analysis: Cancer extent prediction

We will quantify the concordance of the AI system's prediction of linear cancer extent expressed in millimetres with the reference standards in those tuning, internal validation and external validation cohorts where a reference standard is available (AUH, KUH-1, STHLM3, SUH, STG, MLP, SCH, SFI, SFR). The concordance will be quantified using root mean squared error (RMSE). In addition, we will also report Pearson's linear correlation coefficient, and show scatter plots of predicted millimetre cancer length vs. millimetre cancer length reported by the reference standard. The analysis will be conducted on slide level (AUH, KUH-1, STHLM3, SUH, STG, SFR), anatomical location level (MLP, SFI, SCH) and/or patient level (MLP, SCH, SFI, SFR) depending on the granularity of the available reference standards (see Table 3). Confidence intervals will be computed using the non-parametric bootstrap over cases.

### Secondary analysis: Cribriform cancer detection

We will quantify the concordance of the AI system's prediction of the presence of cribriform cancer with the reference standards in those internal and external validation cohorts where a reference standard is available (MUL, SCH, STHLM3, SUH). The tuning set has an insufficient number of cribriform samples for evaluation and will be included in the training. The concordance will be quantified using unweighted Cohen's kappa. In addition, we will also report AUROC, sensitivity (true positive rate), specificity (true negative rate) and confusion matrices. Slides reported as borderline for cribriform cancer will be considered negative. The analysis will be conducted on slide level. Confidence intervals will be computed using the non-parametric bootstrap over cases.

### Secondary analysis: Perineural invasion detection

We will quantify the concordance of the AI system's prediction of the presence of perineural invasion with the reference standards in those internal and external validation cohorts where a reference standard is available (MUL, SCH, STHLM3, SUH). The tuning set has an insufficient number of PNI samples for evaluation and will be included in the training. The concordance will be quantified using unweighted Cohen's kappa. In addition, we will also report AUROC, sensitivity (true positive rate), specificity (true negative rate) and confusion matrices. Slides reported as borderline for perineural invasion will be considered negative. The analysis will be conducted on slide level. Confidence intervals will be computed using the non-parametric bootstrap over cases.

## Subgroup analyses

**Subgroup analysis A:** We will measure the performance of the AI system in terms of the primary and secondary objectives across subgroups of patients divided by age. Analysis will be conducted on the cohorts where age information can be retrieved (see Table 1) according to the age groups: <50, 50 - 59, 60 - 69, and ≥ 70.

**Subgroup analysis B:** We will measure the performance of the AI system in terms of the primary and secondary objectives across subgroups of patients divided by biopsy sampling technique (systematic vs. targeted vs. combined). The analysis will be conducted on the cohorts where biopsy sampling technique information can be retrieved.

**Subgroup analysis C:** We will measure the performance of the AI system in terms of the primary and secondary objectives across subgroups of patients who were treatment-naive or had received treatment for benign prostatic hyperplasia (BPH) (using e.g. 5-alpha reductase inhibitors) before the biopsy procedure. The analysis will be conducted on the cohorts where treatment information can be retrieved. Some (very few) individuals included in the patient cohorts may also have undergone prior prostate cancer treatment (e.g. radiation therapy), but the number of cases is insufficient for a subgroup analysis.

**Subgroup analysis D:** We will measure the performance of the AI system in terms of the primary objective on subgroups of slides representing morphological subtypes of benign and malignant tissue that are usually hard for pathologists to diagnose. We evaluate the performance of the AI system in the STHLM3 morphological subtypes internal validation cohort, the KUH-2 external validation cohort and the AQ external and partly external validation cohorts. See Table 5 for the distribution of morphological subtypes reported in each cohort. We will evaluate performance in terms of cancer diagnosis and additionally, Gleason scoring, where applicable to the subtype.

**Subgroup analysis E:** We will measure the performance of the AI system in terms of the primary objective across subgroups of slides which required IHC staining for confirming the diagnosis and slides which the pathologists could assess without IHC. The analysis will be conducted on the cohorts where information on IHC can be retrieved (see Table 6).

**Subgroup analysis F:** We will measure the performance of the AI system in terms of the primary objective in comparison to the state-of-the-art algorithms developed in the PANDA challenge (Bulten *et al.*, 2022). The analysis will be conducted on the subgroups of the KUH-1,

RUMC and STHLM3 cohorts representing the internal and external validation sets of PANDA. For a fair comparison, we will apply the AI system on the WSIs provided to the challenge participants, which differ in terms of preprocessing and file format from the underlying original WSIs of the KUH-1 and STHLM3 cohorts, which are used in our primary analysis.

A. We evaluate the performance in the tuning cohort KUH-1 (i.e. PANDA European external validation set) and compare the AI system with the PANDA challenge algorithms.

B. We evaluate the performance in the combined PANDA subset of the RUMC and STHLM3 internal validation cohorts (i.e. PANDA internal validation set) and compare the AI system with the PANDA challenge algorithms.

## Sensitivity analyses

**Sensitivity analysis A:** We will evaluate the reproducibility of the AI system's output in terms of the primary and secondary objectives on WSIs obtained from the same slides on multiple scanners. The analysis will be conducted on the STHLM3 tuning and internal validation cohorts and the MUL external validation cohort, which contain WSIs rescanned on different scanners (see Table 2). In the STHLM3 cohort, a subset of slides (n=287) have been rescanned on five scanners: Aperio AT2 DX, Grundium Ocus40, Hamamatsu NanoZoomer 2.0-HT C9600-12, Hamamatsu NanoZoomer XR C12000-02 and Philips IntelliSite UFS. In the MUL cohort, a subset of slides (n=503) have been rescanned on two scanners: Grundium Ocus40 and Philips IntelliSite UFS. We will quantify the reproducibility of the AI predictions across scanners using QWK, and LWK and the percentage of slides with discordant predictions for each objective and each pair of scanners. We will additionally report confusion matrices.

**Sensitivity analysis B:** To put the discrepancies between the AI system and the reference standards in the context of inter-observer variation between pathologists, we will quantify all-against-all pairwise agreements in panels consisting of pathologists and the AI system.

For the primary objective, the analysis will be conducted on subsets of the STHLM3 (ImageBase) and RUMC (PANDA Radboud) internal validation cohorts and on the full UKK and WNS external validation cohorts, which were assessed by a panel of pathologists and have per-pathologist grades available in addition to their consensus (see Table 3). For the secondary objectives of cribriform cancer and PNI detection, the analysis will be conducted on subsets of the STHLM3 internal validation cohort, assessed by panels of pathologists (see Table 4).

We will calculate the average pairwise agreement (QWK and LWK for the primary objective, unweighted Cohen's kappa for the secondary objectives) for all the pathologists in the panels, including the AI system, and compare the average AI-pathologist agreement to the average pathologist-pathologist agreement. Confidence intervals will be computed using bootstrapping, as detailed before (Egevad *et al.*, 2018).

**Sensitivity analysis C:** To assess the sensitivity of the results to different pathologists providing the cohort-specific reference standards and to isolate differences in observed AI performance due to varying reference standards from those due to imperfect generalisation to different labs and scanners, we will repeat the primary analysis using a consistent reference standard. We will measure the agreement between the AI system and the uniform reference standard set by the lead pathologist (L.E.) on subsets of the SUH and RUMC internal validation cohorts and the AUH, MLP, MUL, SCH, SFI, SFR, and SPROB20 external validation cohorts (see Table 3 for a summary of the re-assessed subsets and Section 7 for details on the case selection for each cohort). While the original reference standards were varyingly reported either on the level of slides, anatomical locations, or patients, L.E.'s re-assessments are consistently reported on slide level.

Furthermore, we will measure the agreement in ISUP grades (QWK and LWK) between the original reference standards and the lead pathologist on the re-assessed subsets of each cohort. To facilitate this comparison for cohorts with original reference standards provided on anatomical location or patient level (whereas the grading by L.E. is on slide level), the location or patient level grading by L.E. will be obtained as the maximum ISUP grade over all slides belonging to a location or patient.

**Sensitivity analysis D:** We will perform a sensitivity analysis that involves a re-assessment of slides where the AI system committed clinically significant errors by repeating the primary analysis against the updated reference standard. This analysis aims to evaluate what portion of clinically significant errors can be attributed to data quality issues, such as mistyped information in the reference standard tables, mixed-up slide identifiers, or WSI scanning issues in cases where the original reference standard was set using a microscope. Significant errors are defined as cases where the AI model predicts a slide as benign, but the reference standard indicates ISUP grade $\geq 2$, or conversely the AI predicts a slide as ISUP grade $\geq 2$, but the reference standard indicates benign. These slides will be re-assessed by the lead pathologist (L.E.) and/or other experienced uropathologists, blinded to the original reference standard and the AI output. If a slide cannot be assessed due to e.g. poor focus, it will be excluded. The evaluation will be conducted on the internal and external validation cohorts, on both the full cohorts after updating the reference standards, and on only the updated subsets. Additionally, during this analysis, pathologists will report whether any of the cases with clinically significant errors represent ductal adenocarcinoma (DAC). Despite being the second most common subtype of prostate cancer after acinar adenocarcinoma, DAC only accounts for 0.17% of prostate cancers (Ranasinha *et al.*, 2021) and may therefore be challenging for AI to detect due to the limited amount of training data.

**Sensitivity analysis E:** We will perform a sensitivity analysis that involves the exclusion of samples reported by the pathologists as "borderline" for cribriform cancer or PNI, followed by repeating the secondary analyses concerning these objectives. Conducting the analysis only on samples indicated as negative or positive will provide an estimate of the AI system's performance in detecting cribriform cancer and PNI less affected by the uncertainty and subjectivity in the definition of these entities. We will additionally quantify the prevalence of borderline diagnoses among slides initially classified as false positives vs. true negatives to quantify whether borderline cases are overrepresented among false positives. This would indicate that false positives mainly arise due to uncertainty of the reference standard.

## Exploratory analysis: Evaluate visualisations of the AI output

We will output visualisations of the AI system's predictions to highlight areas on each slide containing different Gleason patterns, cribriform cancer or PNI. The visualisations will be assessed qualitatively by the lead pathologist (L.E.) and/or other experienced uropathologists for concordance with their assessments. We may additionally quantify the rate of agreement between the AI system and the pathologists by collecting region annotations to serve as a reference standard, and by calculating the pixel-wise sensitivity, specificity, intersection over union or other suitable metrics.

## Exploratory analysis: Evaluate the impact of tissue segmentation algorithms

Detecting tissue from the background to only apply the rest of the analysis on tissue pixels is a common preprocessing step for most computational pathology algorithms. While this task of tissue segmentation may seem trivial, many modern AI algorithms reach such low error rates in their main task, that any errors in tissue detection can contribute to the overall model performance in a considerable way. In particular, missed tissue poses a risk of false negative diagnoses, if this leads to the exclusion of malignant tissue from the analysis. We will evaluate the effect of tissue segmentation on the overall performance of the AI system in terms of the primary and secondary objectives by comparing two different tissue segmentation algorithms. One of the algorithms represents classical image processing and relies on filtering and thresholding the image (Ström *et al.*, 2020). The other algorithm is a trained deep learning based segmentation model. We will apply both algorithms to perform the tissue segmentation during model training and validation and compare the results on the internal and external validation cohorts.

**Exploratory analysis: Evaluate end-to-end vs. transfer-learning-based models**

Recently, so-called foundation models trained in a self-supervised manner on large and heterogeneous datasets, have been proposed as generally applicable solutions to diverse tasks in computational pathology as an alternative to tissue type or task specific models (Chen *et al.*, 2024). We aim to compare our end-to-end trained prostate cancer specific model to transfer-learning-based models relying on state-of-the-art foundation models for histopathology. We will apply a suitable foundation model as a feature extractor and train an additional classifier to adapt the model to the task of diagnosis and Gleason scoring of prostate biopsies. For this transfer learning step, we will use the same development cohorts as for the end-to-end trained model. We will then evaluate the model on the same internal and external validation cohorts as the end-to-end trained model for a direct comparison.

**Exploratory analysis: Evaluate the impact of physical colour calibration**

Variations in the reproduction of colour across different digital pathology scanners may pose a problem for AI, leading to inconsistent model outputs depending on the scanner used for slide digitisation. A physical calibrant in the form of a spectrophotometrically characterised slide has been proposed as a means for standardising the colour characteristics of WSIs acquired with different scanners (Clarke *et al.*, 2018). We will evaluate the impact of applying physical colour calibration on the performance of the AI model on those internal and external validation cohorts where the calibrant slide could be scanned on the same scanner as the prostate biopsies to allow calibration.

## 8.3. Confounding factors

Statistical confounding, or spurious correlations, in the training and validation data of predictive models, may lead to "shortcut learning" or so-called "Clever Hans predictors" (Lapuschkin *et al.*, 2019), where overly optimistic performance on validation data is seen as the result of the model taking advantage of unintended correlations between some attributes of the data and the correct labels. Such biases are also common in digital pathology datasets (Howard *et al.*, 2021; Schmitt *et al.*, 2021). We have carefully considered the potential presence of such biases in our cohorts and taken the steps described below to mitigate the issue.

An important confounding factor is the scanner instruments used for digitising various subsets of our data cohorts. Patients in different cohorts and subsets of cohorts have been sampled in varying ways, leading to differences in the compositions of these groups in terms of GS and ISUP grade distribution. These correlations between specific clinical sites or scanner instruments and the target labels can create biases during training since the model could learn to associate the appearance of WSIs obtained from a specific site or with a specific scanner with a higher or lower likelihood of a particular diagnostic or grading outcome. If the same bias is present in validation data, this will lead to overly optimistic results. Conversely, if the bias present in training data is not present in the validation data, a model relying on these spurious correlations will perform poorly. The main approach we have taken to mitigate the risk of overly optimistic validation results is relying on fully external validation data. The external validation cohorts represent patients, clinical sites, laboratories and scanners not present in the training data. This minimises the risk of the same spurious correlations appearing in both training and external validation data. When it comes to discouraging the model from learning any spurious correlations between laboratories or scanners and the target labels, which could result in suboptimal performance in the absence of these correlations, we will apply a sampling scheme which removes the correlations between these variables during model training.

Another common confounding factor we have identified is markings on the slides. Pathologists often place pen marks on the glass slides to indicate cancerous regions. These can lead the AI model to directly associate the presence of markings with the presence of cancer, or indirectly to associate image quality artefacts such as poor focus caused by the pen marks with a higher likelihood of cancer being present. We have mitigated these issues by 1) Applying tissue detection and masking of background pixels as an image preprocessing step, ensuring that pen markings adjacent to tissue will not be shown to the model, 2) Washing and rescanning of slides where pen markings are placed on top of tissue or caused focusing issues, or 3) Excluding slides where neither of the first two options was possible. The first approach of background masking is applied to all the WSIs included in the study. The second approach of washing slides was applied to the development cohorts where we had control over the scanning process, namely STHLM3 and SUH. In the RUMC cohort, we excluded slides with pen marks on the tissue based on the findings of the participants in the PANDA challenge.

## 8.4. Representative sampling

A key issue in the evaluation of diagnostic tests is how disease prevalence influences estimates of statistical measures used to assess the diagnostic performance of the tests. Prevalence is generally defined as the proportion of individuals in a population who have a particular disease at a given time. However, more specifically, the prevalence relates to the datasets used for evaluating a diagnostic test.

The positive predictive value (PPV; i.e. the probability that individuals with a positive test result truly have the disease), negative predictive value (NPV; i.e. the probability that individuals with a negative test result truly do not have the disease), and the Cohen's kappa statistics are influenced by the disease prevalence in the datasets used for evaluating the performance of diagnostic tests. As prevalence increases, the PPV of a test also increases; and conversely, NPV decreases with increasing prevalence. This relationship means that in datasets where a disease (or disease subtype) is more common, the test's ability to identify true positives increases and true negatives decreases. Similarly, the disease prevalence and case mix will impact estimates of Cohen's kappa.

In contrast to PPV, NPV and Cohen's kappa, sensitivity (also known as true positive rate i.e. the ability of a test to correctly identify patients with the disease) and specificity (also known as true negative rate i.e. the ability to correctly identify those without the disease) are not affected by changes in prevalence. These measures are intrinsic properties of the test and do not depend on how common the disease is in a population or dataset.

The sampling scheme or experimental design impacts the estimated prevalence in a study, thereby affecting the diagnostic performance statistics that are sensitive to prevalence. For example, in case-control studies, the prevalence is artificially set by the researcher. In datasets collected for the development of diagnostic AI systems (such as the one described in this protocol), it is common to upsample patients with a disease or disease subtype. If a consecutive case series were used for training an AI system to perform Gleason scoring, a very large set would be required in order to ensure a sufficiently large subsample of e.g. Gleason score 9 and 10 samples for efficient training. Similarly, convenience sampling, where subjects are selected based on their availability rather than at random or according to a defined study design, can lead to a sample with a prevalence rate that does not match the general population. These types of experimental designs and sampling schemes can lead to assessments of PPV, NPV, and Cohen's kappa that do not reflect estimates that would be obtained in a consecutive case series in the general population.

The impact of prevalence on performance estimates underlines the importance of carefully considering the design of diagnostic studies. When prevalence is expected to differ, adjustments or different interpretations of PPV and NPV may be necessary to avoid misinformative conclusions. The data we use for training and evaluation of the AI system is a mixture of convenience samples (AMU, AQ, KUH-2, RUMC, SPROB20, STG) and data representing consecutive clinical cases or another well defined and controlled sampling scheme (AUH, KUH-1, MLP, MUL, SCH, SFI, SFR, STHLM3, SUH, UKK, WNS). For the datasets with a known sampling scheme and experimental design, we can use prior probability shift corrections to achieve estimates of PPV, NPV, and Cohen's kappa on a well defined base population (Schölkopf *et al.*, 2012; Heiser, Allikivi and Kull, 2020).

## 8.5. Power

We have not performed formal power (or sample size) calculations. The reason for this is as follows:

- The central objective of this study is to calculate point estimates of performance (using statistical measures as described above) and their confidence intervals, rather than emphasising power to detect a specific effect size (which is more relevant when comparing interventions or diagnoses).
- This is a retrospective evaluation of AI for prostate pathology. This means that the sample size is fixed based on the datasets at hand.

## 8.6. Data quality and label noise

Collecting and pseudonymising or anonymising clinical and pathology data and associating these records with the correct WSIs requires a number of steps, each introducing potential sources for error. Our data collection, management and verification process generally followed these steps:

**Retrieval and digitisation of clinical/pathology data:** Depending on the data cohort, the clinical and pathology data were extracted from existing databases/registries (STHLM3) in tabular form, provided in tabular form by the data providing sites (AMU, AQ, AUH, MLP, MUL, RUMC, SPROB20, SUH, UKK, WNS) or tabulated manually in-house from pathology reports scanned into PDF files (KUH-1, KUH-2, SCH, SFI, SFR, STG). The manual tabulation in-house involved human translation of the reports from Finnish (SFI), French (SCH, SFR) and Swedish (KUH-1, KUH-2, STG) by trained non-experts fluent in the respective languages. Patient identifiers were pseudonymised during the data extraction or tabulation process by each data provider.

**Retrieval and digitisation of slides:** Slides were retrieved from the respective archives at each site and scanned with the instruments tabulated in Table 2. Each slide had a label with an identifier and depending on the scanning site, the identifiers were stored either in the form of macro/label images as part of the WSI metadata, automatically detected from QR codes and stored as WSI metadata, or manually typed in by the scanner operator when naming the resulting WSI files.

**Linking slides to clinical/pathology data:** Depending on the manner in which slide identifiers were stored for each WSI, the linking step involved one of the following approaches. For WSIs, where the identifier was manually typed into the filename, customised scripts were written in Python for each data cohort to parse the filename strings. This involved comparing the parsed identifiers to those present in the clinical/pathology data, and iterative refinements to rectify issues such as missing or additional zeros, missing or additional whitespace or other delimiters, and discrepancies with the representation of characters not belonging to the Basic Latin (standard ASCII) set, e.g. Ä or Ö. For WSIs, where the identifier was stored in the form of WSI metadata, we used an in-house developed optical character recognition (OCR) system to extract identifiers in a semi-automated manner from the QR-code based metadata items and the macro/label images embedded in the WSIs. The system first extracted the QR-code based identifier, if available, or performed OCR using the *pytesseract* (version 0.3.2) implementation of the Tesseract OCR engine (Smith, 2007). The system featured a simple user interface, which presented the automatically detected identifier pre-filled into a text box, alongside the macro/label image of the slide. The human operator then had the option of accepting the proposed identifier or correcting it manually based on the label image. All identifiers were assessed by trained non-experts using this semi-automated approach.

**Relabeling:** Slides and patients were initially labelled independently by each data provider using pseudonymised identifiers. This poses a risk that the same identifier (e.g. Patient_01) is used by multiple data providers, which would cause ambiguous matches in the final combined dataset. In order to minimise this risk and to obtain unique identifiers for each WSI, each slide and each patient, we calculated unique MD5 hashes based on the variables below. This step additionally provided another round of pseudonymisation to minimise the risk of any non-pseudonymised identifiers being accidentally used by the data providing sites.
- WSI ID: Filename + scanner serial number + scanning time stamp
- Slide ID: Cohort name + original slide ID
- Patient ID: Cohort name + original patient ID

**Verification:** The final dataset covering all the data cohorts is managed internally as a CSV spreadsheet, generated and maintained using scripts written in Python relying on *pandas (Creators The pandas development team; McKinney, 2010)*. Upon generation and any modifications, the dataset undergoes comprehensive unit testing to ensure correctness, implemented in Python using the *unittest* framework. A version history of the dataset is retained to allow tracing back errors. In summary, the tests used for verification cover the following aspects. The uniqueness and unambiguity of matches based on the identifiers described above are verified. Patient-level variables are tested for consistency across all slides and WSIs from the same patient, and slide level variables are checked for consistency across multiple WSIs representing the same slide. We verify that all variables have valid values, with specific tests for categorical, quantitative, and Boolean variables and test for logical mismatches between variables (e.g. a slide negative for cancer cannot be positive for PNI). We ensure there is no overlap between patients in different development vs. validation splits or between cross-validation folds in the development data. Please refer to the Supplementary Appendix Section 2 for an extensive list of all the tests.

## 9. Discussion

This study protocol underscores our dedication to transparency and scientific rigour in developing AI systems for medical diagnostics. The protocol outlines data cohorts, development-validation partitions, performance metrics and an experimental pipeline prespecified before any investigations or experiments on the validation datasets have taken place. For each data cohort, we report information on patient characteristics and selection, biopsy acquisition, histopathological sample preparation, digitisation, and previous utilisation of the cohorts in earlier studies on other AI systems. Furthermore, we report reference standard protocols detailing the variables assessed by pathologists, the level of assessment (pixels, slides, anatomical locations or patients), and any additional re-assessments. This comprehensive documentation of data cohorts facilitates transparency and reproducibility of the research, interpretation of data diversity and representativeness, as well as reliability and integrity of developing and validating the AI system. The study results will be submitted for publication regardless of whether they are positive, negative or inconclusive in relation to the study hypothesis.

Despite the rigorous design, the study has a number of limitations, which we aim to address in future revisions of the protocol and in follow-up studies. Firstly, many AI systems, including those developed for diagnostic purposes, often suffer from the under-representation of certain demographic groups in the data used for their development and validation (Garin *et al.*, 2023). In this study as well, we recognise potential biases in patient demographic representation and are committed to addressing them through additional data collection and subsequent validation processes. Importantly, while all data cohorts and partitions are predefined, the protocol is designed to accommodate the addition of new cohorts for development (up until the model design freeze and initiation of the validation phase) or for validation without altering the initial partitions. For example, we are currently collecting validation data from ethnically diverse North American (Vigneswaran *et al.*, 2024) and Middle Eastern cohorts. This protocol will be extended accordingly to support additional retrospective evaluation of the AI system across these and other patient populations on a global scale.

Secondly, reproducible AI performance across different digital pathology scanners would greatly facilitate scalable clinical deployment of AI systems, and we address this question in a prespecified cross-scanner consistency analysis, which currently has some limitations. The majority of the scanners used for rescanning slides from the STHLM3 and MUL validation cohorts for this analysis were also involved in the digitisation of the development data (except the Grundium Ocus40 scanner). This can potentially lead to optimistic results due to the AI model having been exposed to the variation seen between these scanners during training. Nevertheless, it should be noted that all the external validation cohorts described in this protocol have been digitised on scanners not involved in the collection of the AI development data, which will allow us to assess cross-scanner generalisation indirectly. For a direct comparison using the exact same set of slides digitised on multiple external scanners (i.e. corresponding to a paired study design), we are in the process of rescanning slides on additional scanners. This will allow us to repeat the analysis using scanners fully external to the AI system in a follow-up study.

Thirdly, the criteria for distinguishing between uropathologists and general pathologists are often vague and lack standardised definitions across different countries and hospitals. This may introduce differences when comparing agreement rates between general pathologists and uropathologists across different cohorts. Furthermore, there are varying practices in the reporting of prostate pathology, for example in terms of measuring cancer extent and summarising Gleason scoring results on the patient level. This might introduce additional systematic differences when evaluating the performance across cohorts, which we have mitigated by additional re-assessments performed in a consistent manner by the lead pathologist (L.E.). Still, prostate pathology assessment remains a subjective process and inter- and intra-observer variability cannot be fully eliminated from the reference standards.

This protocol covers retrospective validation of an AI system for assessing prostate core needle biopsies for four main objectives i.e. prostate cancer diagnosis and grading, cancer extent, cribriform cancer and perineural invasion. These objectives are crucial for predicting disease prognosis and guiding treatment for prostate cancer patients. However, additional objectives of our work on AI for prostate cancer will be added. For example, the diagnostic AI system described in this protocol can serve as a foundation model for developing models for direct prognostication (based on relevant oncological outcomes, such as time to biochemical recurrence (BCR), metastatic disease or prostate cancer death) and treatment prediction, and with further refinements can be adapted to predict additional objectives based on prostate morphology or other use cases in prostate pathology, such as reducing the need for IHC staining (Table 6). Moreover, we will perform molecular characterisation (genomic and transcriptomic profiling) of tissue samples from diagnostic biopsies, following the same protocol as we use in the ProBio trial for metastatic prostate cancer (Crippa *et al.*, 2020; De Laere *et al.*, 2022). Linked imaging and genomic data will be used to develop models to predict clinically important genomic alterations and mutations from the morphological data in the WSIs. For example, we will develop AI models for the prediction of alterations in the BRCA genes; patients with alterations in these genes tend to respond well to poly ADP-ribose polymerase (PARP) inhibitors (de Bono *et al.*, 2020; Chi *et al.*, 2023; Fizazi *et al.*, 2023). Such AI models could in a clinical setting help to triage tissue samples for genomic analysis to verify AI predictions, which would reduce costs and improve chances of detecting clinically actionable genetic information. We will also use the data presented in this protocol to further develop conformal predictors to detect unreliable AI predictions (Olsson *et al.*, 2022). Additional information regarding these objectives will be added in future revisions of this protocol (and then noted in the revision history of the document).

The importance of relating performance to a well defined population (see Section 8.4) motivates prospective evaluation in a clinical trial, which we are currently planning. (The prospective trial will be described and detailed in its own protocol.) Prospective evaluation also enables assessing aspects relevant to the clinical implementation of AI systems that are not possible to evaluate on retrospective data, e.g. user interaction, pathologist-in-the-loop approaches, etc. This planned clinical trial will thus evaluate the AI system performance in a real-world clinical setting against gold-standard diagnostic practices and provide evidence of its efficacy and reliability for guiding clinical decision-making in prostate cancer diagnosis.

## 10. Ethical considerations

The study is conducted in agreement with the Helsinki Declaration. The collection of patient samples was approved by the Stockholm regional ethics committee (permits 2012/572-31/1, 2012/438-31/3, and 2018/845-32), the Swedish Ethical Review Authority (permit 2019-05220), and the Regional Committee for Medical and Health Research Ethics (REC) in Western Norway (permits REC/Vest 80924, REK 2017/71). Informed consent was provided by the participants in the Swedish dataset. For the other datasets, informed consent was waived by the institutional review board due to the usage of de-identified prostate specimens in a retrospective setting.

## 11. Acknowledgements

A.B. received a grant from the Health Faculty at the University of Stavanger, Norway. B.G.P and K.D.S received funding from Innovation Fund Denmark (Grant no. 8114-00014B) for the Danish branch of the NordCaP project. M.R. received funding from the Swedish Research Council and the Swedish Cancer Society. P.R. received funding from the Research Council of Finland (Grant no. 341967) and the Cancer Foundation Finland. M.E. received funding from the Swedish Research Council, Swedish Cancer Society, Swedish Prostate Cancer Society, Nordic Cancer Union, Karolinska Institutet, and Region Stockholm. K.K. received funding from the David and Astrid Hägelen Foundation, Instrumentarium Science Foundation, KAUTE Foundation, Karolinska Institute Research Foundation, Orion Research Foundation and Oskar Huttunen Foundation.

We want to thank Carin Cavalli-Björkman, Astrid Björklund and Britt-Marie Hune for assistance with scanning and database support. We would also like to thank Simone Weiss for assistance with scanning in Aarhus, and Silja Kavlie Fykse and Desmond Mfua Abono for scanning in Stavanger. We would like to acknowledge the patients who participated in the STHLM3 diagnostic study and the OncoWatch and NordCaP projects and contributed the clinical information that made this study possible.

The computations are possible through the National Academic Infrastructure for Supercomputing in Sweden (NAISS) and the Swedish National Infrastructure for Computing (SNIC) at C3SE partially funded by the Swedish Research Council through grant agreement no. 2022-06725 and no. 2018-05973, by the supercomputing resource Berzelius provided by the National Supercomputer Centre at Linköping University and the Knut and Alice Wallenberg Foundation, and by CSC - IT Center for Science, Finland.

## 12. Competing interests

N.M., L.E., K.K. and M.E. are shareholders of Clinsight AB, and M.R. is a co-founder and shareholder of Stratipath AB.

medRxiv preprint doi: https://doi.org/10.1101/2024.07.04.24309948; this version posted July 7, 2024. The copyright holder for this preprint **(which was not certified by peer review)** is the author/funder, who has granted medRxiv a license to display the preprint in perpetuity. It is made available under a CC-BY 4.0 International license .http://arxiv.org/abs/2307.05519.

35. Jung, M. *et al.* (2022) 'Artificial intelligence system shows performance at the level of uropathologists for the detection and grading of prostate cancer in core needle biopsy: an independent external validation study', *Modern pathology: an official journal of the United States and Canadian Academy of Pathology, Inc*, 35(10), pp. 1449–1457.

36. Kartasalo, K. *et al.* (2022) 'Detection of perineural invasion in prostate needle biopsies with deep neural networks', *Virchows Archiv: an international journal of pathology*, 481(1), pp. 73–82.

37. Kleppe, A. *et al.* (2021) 'Designing deep learning studies in cancer diagnostics', *Nature reviews. Cancer*, 21(3), pp. 199–211.

38. Kweldam, C.F. *et al.* (2016) 'Gleason grade 4 prostate adenocarcinoma patterns: an interobserver agreement study among genitourinary pathologists', *Histopathology*, 69(3), pp. 441–449.

39. Lapuschkin, S. *et al.* (2019) 'Unmasking Clever Hans predictors and assessing what machines really learn', *Nature communications*, 10(1), p. 1096.

40. Liu, X. *et al.* (2020) 'Reporting guidelines for clinical trial reports for interventions involving artificial intelligence: the CONSORT-AI extension', *The Lancet. Digital health*, 2(10), pp. e537–e548.

41. Marée, R. *et al.* (2016) 'Collaborative analysis of multi-gigapixel imaging data using Cytomine', *Bioinformatics* , 32(9), pp. 1395–1401.

42. McGenity, C., Bossuyt, P. and Treanor, D. (2022) 'Reporting of Artificial Intelligence Diagnostic Accuracy Studies in Pathology Abstracts: Compliance with STARD for Abstracts Guidelines', *Journal of pathology informatics*, 13, p. 100091.

43. McKinney, W. (2010) 'Data Structures for Statistical Computing in Python', in *Proceedings of the 9th Python in Science Conference*. *Python in Science Conference*, SciPy. Available at: https://doi.org/10.25080/majora-92bf1922-00a.

44. Melia, J. *et al.* (2006) 'A UK-based investigation of inter- and intra-observer reproducibility of Gleason grading of prostatic biopsies', *Histopathology*, 48(6), pp. 644–654.

45. Mongan, J., Moy, L. and Kahn, C.E., Jr (2020) 'Checklist for Artificial Intelligence in Medical Imaging (CLAIM): A Guide for Authors and Reviewers', *Radiology. Artificial intelligence*, 2(2), p. e200029.

46. Mulliqi, N. *et al.* (2021) 'OpenPhi: An interface to access Philips iSyntax whole slide images for computational pathology', *Bioinformatics* [Preprint]. Available at: https://doi.org/10.1093/bioinformatics/btab578.

47. Nagendran, M. *et al.* (2020) 'Artificial intelligence versus clinicians: systematic review of design, reporting standards, and claims of deep learning studies', *BMJ* , 368, p. m689.

48. Olsson, H. *et al.* (2022) 'Estimating diagnostic uncertainty in artificial intelligence assisted pathology using conformal prediction', *Nature communications*, 13(1), p. 7761.

49. Ozkan, T.A. *et al.* (2016) 'Interobserver variability in Gleason histological grading of prostate cancer', *Scandinavian journal of urology*, 50(6), pp. 420–424.

50. Pantanowitz, L. *et al.* (2018) 'Twenty years of digital pathology: An overview of the road travelled, what is on the horizon, and the emergence of vendor-neutral archives', *Journal*
47

# 14. Figures and tables

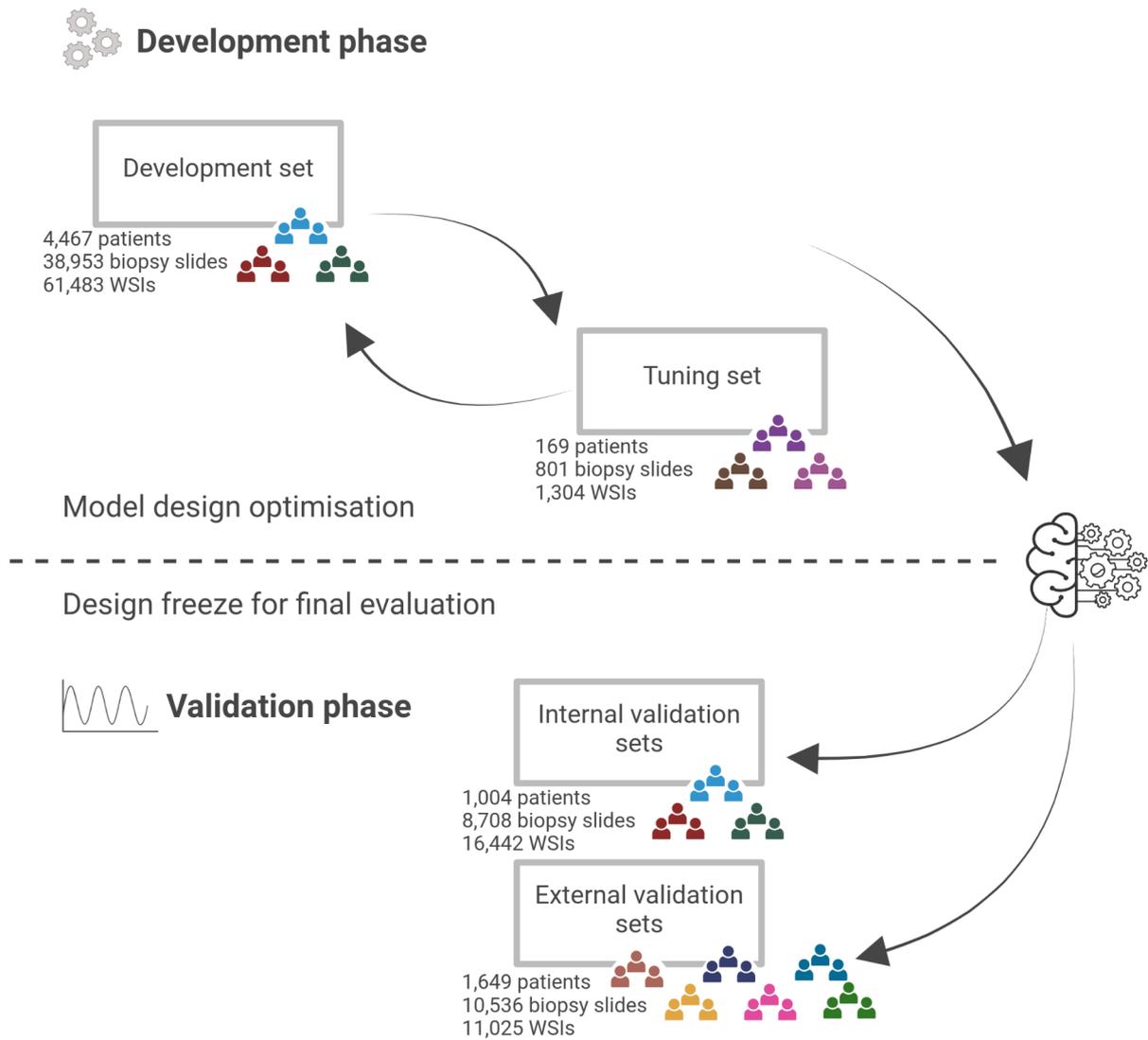

**Figure 1. Overview of the study design.** The study design has two main steps: (top) The development phase involves model design optimisation through an iterative process of experiments. In each experiment, the model is trained and its performance is evaluated on the development set using cross-validation and on a separate tuning set. (bottom) The validation phase is initiated with a design freeze, after which no further changes to the model take place. Validation comprises the assessment on the internal data (i.e. collected from the same laboratory and/or using the same scanner as development data) and the external data (i.e. collected from other laboratories using other scanners than any of the development data). This figure was created with BioRender.

**Table 1. Patient clinical and pathological characteristics.** Patient and slide level information for the development, tuning, internal, and external validation cohorts including age, PSA, ISUP grade and cancer length distributions. Averaged age and PSA are shown for patients who underwent multiple biopsies. The ISUP distributions are based on the initial, original grading excluding any re-assessments. For the AMU, MLP, SCH, SFI, SFR and SPROB20 cohorts, where pathology reporting was performed on anatomical location or patient level, the total summed numbers of slides associated with a given ISUP grade or cancer length are shown. The AUH cohort has an age range of 50.4 to 69.9 yrs (mean 63.2 yrs, median 64.0 yrs) and a PSA range of 1.5 ng/mL to 9.8 ng/mL (mean 4.6 ng/mL, median 4.2 ng/mL). The SPROB20 cohort has an age range of 39 to 79 yrs (median 67 yrs). Slides in the AQ, KUH-2 and SUH cohorts missing ISUP grade information represent non-gradable morphological variants. PSA=prostate-specific antigen, ISUP=International Society of Urological Pathology, STHLM3=Stockholm3, SUH=Stavanger University Hospital, RUMC=Radboud University Medical Center, STG=Capio S:t Göran Hospital, KUH-1=Karolinska University Hospital, AMU=Aichi Medical University, AQ=Aquesta Uropathology, AUH=Aarhus University Hospital, KUH-2=Karolinska University Hospital morphological subtypes, MLP=Mehiläinen Länsi-Pohja, MUL=Medical University of Lodz, SCH=Synlab Switzerland, SFI=Synlab Finland, SFR=Synlab France, SPROB20=Spear Prostate Biopsy 2020, UKK=University Hospital Cologne, WNS=Hospital Wiener Neustadt.

| Development | STHLM3 | SUH | RUMC | STG |
|---|---|---|---|---|
| **No. participants (%)** | **n=2,711** | **n=710** | **n=976** | **n=70** |
| **Age, years** | | | | |
| <=49 yrs | 4 (0.14) | 13 (1.83) | | 0 (0.0) |
| 50 - 54 yrs | 216 (7.96) | 35 (4.92) | | 1 (1.42) |
| 55 - 59 yrs | 429 (15.82) | 94 (13.23) | | 2 (2.85) |
| 60 - 64 yrs | 702 (25.89) | 137 (19.29) | / | 4 (5.71) |
| 65 - 69 yrs | 1,207 (44.52) | 191 (26.90) | | 6 (8.57) |
| >= 70 yrs | 153 (5.64) | 240 (33.80) | | 37 (52.85) |
| Missing | 0 (0.0) | 0 (0.0) | | 20 (28.6) |
| **Prostate-specific antigen** | | | | |
| <3 ng/mL | 611 (22.53) | 60 (8.45) | | 2 (2.85) |
| 3 - <5 ng/mL | 1,306 (48.17) | 135 (19.01) | | 1 (1.42) |
| 5 - <10 ng/mL | 592 (21.83) | 350 (49.29) | / | 6 (8.57) |
| >= 10 ng/mL | 202 (7.45) | 163 (22.95) | | 38 (54.28) |
| Missing | 0 (0.0) | 2 (0.28) | | 23 (32.85) |
| **No. slides (%)** | **n=29,536** | **n=4,606** | **n=4,564** | **n=247** |
| **Cancer length** | | | | |
| No cancer | 23,530 (79.67) | 3,435 (74.57) | / | 1 (0.40) |
| >0 - 1 mm | 2,021 (6.84) | 238 (5.16) | | 7 (2.83) |
| >1 - 5 mm | 2,577 (8.72) | 405 (8.78) | | 42 (17.00) |
| >5 - 10 mm | 1,054 (3.56) | 226 (4.90) | | 86 (34.81) |
| >10 mm | 354 (1.19) | 300 (6.51) | | 111 (44.93) |
| Missing | 0 (0.0) | 2 (0.04) | | 0 (0.0) |
| **Cancer grade** | | | | |
| Benign | 23,530 (79.67) | 3,435 (74.57) | 912 (19.98) | 1 (0.40) |
| ISUP 1 (3+3) | 3,571 (12.09) | 683 (14.82) | 731 (16.01) | 1 (0.40) |
| ISUP 2 (3+4) | 1,265 (4.28) | 240 (5.20) | 594 (13.01) | 1 (0.40) |
| ISUP 3 (4+3) | 494 (1.67) | 129 (2.79) | 800 (17.52) | 2 (0.80) |
| ISUP 4 (4+4, 3+5, 5+3) | 377 (1.28) | 54 (1.17) | 668 (14.63) | 32 (12.95) |
| ISUP 5 (4+5, 5+4, 5+5) | 299 (1.01) | 63 (1.36) | 859 (18.82) | 210 (85.02) |
| Missing | 0 (0.0) | 2 (0.04) | 0 (0.0) | 0 (0.0) |

| Internal validation | STHLM3 | SUH | RUMC |
|---|---|---|---|
| **No. participants (%)** | **n=654** | **n=178** | **n=172** |
| **Age, years** | | | |
| <=49 yrs | 3 (0.45) | 1 (0.56) | / |
| 50 - 54 yrs | 58 (8.86) | 6 (3.37) | |
| 55 - 59 yrs | 96 (14.67) | 15 (8.42) | |
| 60 - 64 yrs | 182 (27.82) | 39 (21.91) | |
| 65 - 69 yrs | 289 (44.18) | 46 (25.84) | |
| >= 70 yrs | 26 (3.97) | 71 (39.88) | |
| Missing | 0 (0.0) | 0 (0.0) | |
| **Prostate-specific antigen** | | | |
| <3 ng/mL | 123 (18.80) | 12 (6.74) | / |
| 3 - <5 ng/mL | 321 (49.08) | 23 (12.92) | |
| 5 - <10 ng/mL | 153 (23.39) | 91 (51.12) |  |
| >= 10 ng/mL | 57 (8.71) | 52 (29.21) |  |
| Missing | 0 (0.0) | 0 (0.0) |  |
| **No. slides (%)** | **n=7,036** | **n=1,156** | **n=516** |
| **Cancer length** |  |  |  |
| No cancer | 5,098 (72.45) | 736 (63.70) |  |
| >0 - 1 mm | 583 (8.28) | 52 (4.48) |  |
| >1 - 5 mm | 767 (10.90) | 109 (9.48) | / |
| >5 - 10 mm | 434 (6.16) | 87 (7.50) |  |
| >10 mm | 154 (2.18) | 172 (14.82) |  |
| Missing | 0 (0.0) | 0 (0.0) |  |
| **Cancer grade** |  |  |  |
| Benign | 5,098 (72.46) | 736 (63.66) | 195 (37.79) |
| ISUP 1 (3+3) | 958 (13.62) | 153 (13.23) | 87 (16.86) |
| ISUP 2 (3+4) | 380 (5.40) | 76 (6.55) | 45 (8.72) |
| ISUP 3 (4+3) | 240 (3.41) | 74 (6.37) | 77 (14.92) |
| ISUP 4 (4+4, 3+5, 5+3) | 203 (2.89) | 53 (4.56) | 54 (10.46) |
| ISUP 5 (4+5, 5+4, 5+5) | 157 (2.23) | 64 (5.51) | 58 (11.24) |
| Missing | 0 (0.0) | 0 (0.0) | 0 (0.0) |

| **Tuning** | **STHLM3** | **KUH-1** | **RUMC** |
|---|---|---|---|
| **No. participants (%)** | **n=24** | **n=73** | **n=72** |
| **Age, years** |  |  |  |
| <=49 yrs | 0 (0.0) | 2 (2.73) |  |
| 50 - 54 yrs | 1 (4.16) | 5 (6.84) |  |
| 55 - 59 yrs | 2 (8.33) | 10 (13.69) |  |
| 60 - 64 yrs | 8 (33.33) | 12 (16.43) | / |
| 65 - 69 yrs | 13 (54.16) | 15 (20.54) |  |
| >= 70 yrs | 0 (0.0) | 29 (39.72) |  |
| Missing | 0 (0.0) | 0 (0.0) |  |
| **Prostate-specific antigen** |  |  |  |
| <3 ng/mL | 3 (12.50) | / | / |
| 3 - <5 ng/mL | 12 (50.00) | | |
| 5 - <10 ng/mL | 5 (20.83) | | |
| >= 10 ng/mL | 4 (16.66) | | |
| Missing | 0 (0.0) | | |
| **No. slides (%)** | **n=276** | **n=330** | **n=195** |
| **Cancer length** | | | |
| No cancer | 192 (69.57) | 108 (32.72) | / |
| >0 - 1 mm | 32 (11.59) | 33 (10.00) | |
| >1 - 5 mm | 27 (9.67) | 77 (23.33) | |
| >5 - 10 mm | 16 (5.73) | 75 (22.72) | |
| >10 mm | 9 (3.22) | 37 (11.21) | |
| Missing | 0 (0.0) | 0 (0.0) | |
| **Cancer grade** | | | |
| Benign | 192 (69.57) | 108 (32.72) | 95 (48.72) |
| ISUP 1 (3+3) | 28 (10.14) | 65 (19.70) | 24 (12.31) |
| ISUP 2 (3+4) | 18 (6.52) | 63 (19.09) | 15 (7.69) |
| ISUP 3 (4+3) | 13 (4.71) | 49 (14.85) | 15 (7.69) |
| ISUP 4 (4+4, 3+5, 5+3) | 13 (4.71) | 19 (5.76) | 19 (9.74) |
| ISUP 5 (4+5, 5+4, 5+5) | 12 (4.35) | 26 (7.88) | 27 (13.85) |
| Missing | 0 (0.0) | 0 (0.0) | 0 (0.0) |

| **External validation** | **AMU** | **AQ** | **AUH** |
|---|---|---|---|
| **No. participants (%)** | **n=43** | **n=135** | **n=42** |
| **Age, years** | | | |
| <= 49 yrs | / | / | / |
| 50 - 54 yrs | | | |
| 55 - 59 yrs | | | |
| 60 - 64 yrs | | | |
| 65 - 69 yrs | | | |
| >= 70 yrs | | | |
| Missing |  |  |  |
| **Prostate-specific antigen** |  |  |  |
| <3 ng/mL | 1 (2.32) | / | / |
| 3 - <5 ng/mL | 1 (2.32) |  |  |
| 5 - <10 ng/mL | 11 (25.58) |  |  |
| >= 10 ng/mL | 30 (69.76) |  |  |
| Missing | 0 (0.0) |  |  |
| **No. slides (%)** | **n=73** | **n=136** | **n=102** |
| **Cancer length** |  |  |  |
| No cancer |  |  | 43 (42.15) |
| >0 - 1 mm |  |  | 5 (4.90) |
| >1 - 5 mm | / | / | 18 (17.64) |
| >5 - 10 mm |  |  | 24 (23.52) |
| >10 mm |  |  | 12 (11.76) |
| Missing |  |  | 0 (0.0) |
| **Cancer grade** |  |  |  |
| Benign | 0 (0.0) | 122 (89.70) | 43 (42.15) |
| ISUP 1 (3+3) | 0 (0.0) | 1 (0.73) | 26 (25.49) |
| ISUP 2 (3+4) | 0 (0.0) | 1 (0.73) | 25 (24.50) |
| ISUP 3 (4+3) | 6 (8.21) | 0 (0.00) | 1 (0.98) |
| ISUP 4 (4+4, 3+5, 5+3) | 22 (28.76) | 0 (0.00) | 7 (6.86) |
| ISUP 5 (4+5, 5+4, 5+5) | 45 (60.27) | 1 (0.73) | 0 (0.0) |
| Missing | 0 (0.0) | 11 (8.08) | 0 (0.0) |

| **External validation** | **KUH-2** | **MLP** | **MUL** |
|---|---|---|---|
| **No. participants (%)** | **n=89** | **n=199** | **n=207** |
| **Age, years** |  |  |  |
| <=49 yrs |  |  | 2 (0.96) |
| 50 - 54 yrs | / | / | 4 (1.93) |
| 55 - 59 yrs |  |  | 10 (4.83) |
| 60 - 64 yrs |  |  | 29 (14.00) |

| 65 - 69 yrs | | | 50 (24.15) |
|---|---|---|---|
| >= 70 yrs | | | 108 (52.17) |
| Missing | | | 4 (1.96) |
| **Prostate-specific antigen** | | | |
| <3 ng/mL | | 19 (9.54) | |
| 3 - <5 ng/mL | | 26 (13.06) | |
| 5 - <10 ng/mL | / | 65 (32.66) | / |
| >= 10 ng/mL | | 85 (42.71) | |
| Missing | | 4 (2.03) | |

| **No. slides (%)** | **n=146** | **n=1,964** | **n=1,959** |
|---|---|---|---|
| **Cancer length** | | | |
| No cancer | | 302 (15.37) | |
| >0 - 1 mm | | 24 (1.22) | |
| >1 - 5 mm | / | 207 (10.53) | / |
| >5 - 10 mm | | 191 (9.72) | |
| >10 mm | | 1,189 (60.53) | |
| Missing | | 54 (2.63) | |
| **Cancer grade** | | | |
| Benign | 103 (70.54) | 323 (16.44) | 1,483 (75.70) |
| ISUP 1 (3+3) | 34 (23.28) | 433 (22.04) | 161 (8.21) |
| ISUP 2 (3+4) | 5 (3.42) | 506 (25.76) | 58 (2.96) |
| ISUP 3 (4+3) | 0 (0.0) | 216 (10.99) | 74 (3.77) |
| ISUP 4 (4+4, 3+5, 5+3) | 0 (0.0) | 133 (6.77) | 65 (3.31) |
| ISUP 5 (4+5, 5+4, 5+5) | 0 (0.0) | 353 (17.97) | 118 (6.02) |
| Missing | 4 (2.73) | 0 (0.0) | 0 (0.0) |

| **External validation** | **SCH** | **SFI** | **SFR** |
|---|---|---|---|
| **No. participants (%)** | **n=199** | **n=99** | **n=84** |
| **Age, years** | | | |
| <=49 yrs | 3 (1.50) | / | 1 (1.19) |
| 50 - 54 yrs | 3 (1.50) | | 5 (5.95) |
| 55 - 59 yrs | 22 (11.05) | | 11 (13.09) |
| 60 - 64 yrs | 27 (13.56) | | 11 (13.09) |
| 65 - 69 yrs | 46 (23.11) | 2 (2.0) | 21 (25.00) |
| >= 70 yrs | 98 (49.24) | 3 (3.03) | 35 (41.66) |
| Missing | 0 (0.0) | 94 (94.97) | 0 (0.0) |
| **Prostate-specific antigen** | | | |
| Low | 0 (0.0) | 2 (2.02) | 0 (0.0) |
| Normal | 0 (0.0) | 2 (2.02) | 0 (0.0) |
| Elevated | 19 (9.54) | 8 (8.08) | 0 (0.0) |
| <3 ng/mL | 3 (1.50) | 2 (2.02) | 1 (1.35) |
| 3 - <5 ng/mL | 21 (10.55) | 8 (8.08) | 6 (7.14) |
| 5 - <10 ng/mL | 45 (22.61) | 39 (39.39) | 51 (60.71) |
| >= 10 ng/mL | 39 (19.59) | 32 (32.32) | 16 (19.04) |
| Missing | 72 (36.18) | 6 (6.06) | 10 (11.75) |
| **No. slides (%)** | **n=2,434** | **n=537** | **n=515** |
| **Cancer length** | | | |
| No cancer | 1,580 (64.91) | 311 (57.91) | 373 (72.42) |
| >0 - 1 mm | 22 (0.90) | 16 (2.97) | 1 (0.19) |
| >1 - 5 mm | 156 (6.39) | 39 (7.26) | 34 (6.60) |
| >5 - 10 mm | 88 (3.60) | 30 (5.58) | 32 (6.21) |
| >10 mm | 565 (23.27) | 54 (10.05) | 69 (13.39) |
| Missing | 23 (0.94) | 87 (16.42) | 6 (0.97) |
| **Cancer grade** | | | |
| Benign | 1,580 (64.91) | 311 (57.91) | 373 (72.42) |
| ISUP 1 (3+3) | 325 (13.31) | 61 (11.35) | 87 (16.89) |
| ISUP 2 (3+4) | 201 (8.25) | 51 (9.49) | 28 (5.43) |
| ISUP 3 (4+3) | 183 (7.51) | 50 (9.31) | 3 (0.58) |
| ISUP 4 (4+4, 3+5, 5+3) | 94 (3.86) | 16 (2.97) | 10 (1.94) |
| ISUP 5 (4+5, 5+4, 5+5) | 47 (1.93) | 30 (5.58) | 6 (1.16) |
| Missing | 4 (0.16) | 18 (3.39) | 8 (1.55) |

| External validation | SPROB20 | UKK | WNS |
|---|---|---|---|
| **No. participants (%)** | **n=452** | **n=50** | **n=50** |
| **Age, years** | | | |
| <=49 yrs | / | / | / |
| 50 - 54 yrs | | | |
| 55 - 59 yrs | | | |
| 60 - 64 yrs | | | |
| 65 - 69 yrs | | | |
| >= 70 yrs | | | |
| Missing | | | |
| **Prostate-specific antigen** | | | |
| <3 ng/mL | 13 (2.87) | / | / |
| 3 - <5 ng/mL | 14 (3.09) | | |
| 5 - <10 ng/mL | 30 (6.63) | | |
| >= 10 ng/mL | 190 (42.03) | | |
| Missing | 205 (45.35) | | |
| **No. slides (%)** | **n=2,570** | **n=50** | **n=50** |
| No cancer | / | / | / |
| >0 - 1 mm | | | |
| >1 - 5 mm | | | |
| >5 - 10 mm | | | |
| >10 mm | | | |
| Missing | | | |
| **Cancer grade** | | | |
| Benign | 950 (36.96) | 0 (0.0) | 0 (0.0) |
| ISUP 1 (3+3) | 543 (21.12) | 12 (24.0) | 10 (20.0) |
| ISUP 2 (3+4) | 700 (27.23) | 8 (16.0) | 10 (20.0) |
| ISUP 3 (4+3) | 186 (7.23) | 12 (24.0) | 12 (24.0) |
| ISUP 4 (4+4, 3+5, 5+3) | 103 (4.00) | 8 (16.0) | 8 (16.0) |
| ISUP 5 (4+5, 5+4, 5+5) | 88 (3.42) | 10 (20.0) | 10 (20.0) |
| Missing | 0 (0.0) | 0 (0.0) | 0 (0.0) |

**Table 2. Overview of image acquisition attributes and WSIs.** Cohorts marked with (*) (i.e. STHLM3, STG, and MUL) contain overlapping subsets of slides digitised with different scanners. Other cohorts were either digitised with a single scanner or contain non-overlapping subsets of slides digitised with different scanners. WSI=whole slide image, STHLM3=Stockholm3, SUH=Stavanger University Hospital, RUMC=Radboud University Medical Center, STG=Capio S:t Göran Hospital, KUH-1=Karolinska University Hospital, AMU=Aichi Medical University, AQ=Aquesta Uropathology, AUH=Aarhus University Hospital, KUH-2=Karolinska University Hospital morphological subtypes, MLP=Mehiläinen Länsi-Pohja, MUL=Medical University of Lodz, SCH=Synlab Switzerland, SFI=Synlab Finland, SFR=Synlab France, SPROB20=Spear Prostate Biopsy 2020, UKK=University Hospital Cologne, WNS=Hospital Wiener Neustadt.

| Split | Cohort | Scanning location | Scanning period | Scanner Vendor | Scanner Model | Scanner Serial no. | Magnification (Pixel size) | WSI format | WSI number |
|---|---|---|---|---|---|---|---|---|---|
| Development, tuning and internal validation cohorts | STHLM3* | Department of Medical Epidemiology and Biostatistics, Karolinska Institutet, Solna, Sweden | 07/2014 - 11/2014 | Hamamatsu | NanoZoomer 2.0-HT C9600-12 | 760347 | 20x (0.4520 μm) | .ndpi | 5,726 |
| | | | | | | | | .tiff | 3,417 |
| | | SciLifeLab, Uppsala, Sweden | 09/2017 - 06/2019 | Aperio | AT2 DX | RUD-D10971 | 20x (0.5032 μm) | .svs | 3,667 |
| | | | | | | | | .tiff | 2,445 |
| | | Department of Medical Epidemiology and Biostatistics, Karolinska Institutet, Solna, Sweden | 03/2018 - 06/2019 | Hamamatsu | NanoZoomer XR C12000-02 | 870003 | 20x (0.4536 μm) | .ndpi | 17,973 |
| | | Department of Medical Epidemiology and Biostatistics, Karolinska Institutet, Solna, Sweden | 10/2019 - 06/2020 | Philips | IntelliSite UFS | FMT0047 | 40x (0.2500 μm) | .isyntax | 32,078 |
| | | Department of Medical Epidemiology and Biostatistics, Karolinska Institutet, Solna, Sweden | 02/2023 - 03/2023 | Grundium | Ocus40 | MGU-00003-000184 | 40x (0.2505 μm) | .svs | 2,289 |
| | SUH | Department of Pathology, Stavanger University Hospital, Stavanger, Norway | 02/2022 - 03/2023 | Hamamatsu | NanoZoomer S60 C13210-01 | 000266 | 40x (0.2199 μm) | .ndpi | 5,762 |
| | RUMC | Radboud University Medical Center, Nijmegen, The Netherlands | 01/2019 - 12/2019 | 3DHISTECH | Pannoramic Scan ll | N/A | 20x (0.4861 μm) | .tiff | 5,275 |
| | STG* | Department of Immunology, Genetics, and Pathology, Uppsala University, Uppsala, Sweden | 09/2018 - 10/2018 | Hamamatsu | C13210 | 000058 | 20x (0.4405 μm) | .ndpi | 74 |
| | | Department of Immunology, Genetics, and Pathology, Uppsala University, Uppsala, Sweden | 10/2018 | Hamamatsu | C13210 | 000044 | 20x (0.4409 μm) | .ndpi | 67 |
|  |  | SciLifeLab, Uppsala, Sweden | 12/2018 | **Aperio** | **AT2 DX** | **RUD-D10971** | 20x (0.5032 μm) | **.svs** | 247 |
|  | **KUH-1** | Department of Pathology, Karolinska University Hospital, Solna, Sweden | 07/2019 - 08/2019 | **Hamamatsu** | **NanoZoomer S360 C13220-01** | **000077** | 20x (0.4604 μm) | **.ndpi** | 330 |
| **External and partly external validation cohorts** | **AMU** | Aichi Medical University, Nagakute, Japan | 01/2023 - 12/2023 | **Hamamatsu** | **C13210** | **000218** | 40x (0.2211 μm) | **.ndpi** | 73 |
|  | **AQ** | Department of Medical Epidemiology and Biostatistics, Karolinska Institutet, Solna, Sweden | 10/2019 - 06/2020 | **Philips** | **IntelliSite UFS** | **FMT0047** | 40x (0.2500 μm) | **.isyntax** | 58 |
|  |  | Department of Medical Epidemiology and Biostatistics, Karolinska Institutet, Solna, Sweden | 01/2024 - 02/2024 | **Grundium** | **Ocus40** | **MGU-00003-000184** | 40x (0.2505 μm) | **.svs** | 78 |
|  | **AUH** | Department of Pathology, Aarhus University Hospital, Aarhus, Denmark | 11/2019 - 06/2020 | **Hamamatsu** | **NanoZoomer 2.0-HT C9600-12** | **1Z0209** | 20x (0.4545 μm) | **.ndpi** | 102 |
|  | **KUH-2** | Department of Pathology, Karolinska University Hospital, Solna, Sweden | 07/2022 | **Aperio** | **AT2 DX** | **SS7033** | 20x (0.5032 μm) | **.svs** | 146 |
|  | **MLP** | Finnish Institute of Molecular Medicine, Helsinki, Finland | 10/2019 - 03/2020 | **3DHISTECH** | **Pannoramic 250 Flash III** | **01702** | 40x (0.2427 μm) | **.mrxs** | 1,964 |
|  | **MUL*** | Department of Medical Epidemiology and Biostatistics, Karolinska Institutet, Solna, Sweden | 12/2019 - 01/2020 | **Philips** | **IntelliSite UFS** | **FMT0047** | 40x (0.2500 μm) | **.isyntax** | 503 |
|  |  | Department of Medical Epidemiology and Biostatistics, Karolinska Institutet, Solna, Sweden | 01/2023 - 03/2023 | **Grundium** | **Ocus40** | **MGU-00003-000184** | 40x (0.2505 μm) | **.svs** | 1,945 |
|  | **SCH & SFI & SFR** | Synlab italia srl, Monza, Italy | 06/2022 - 02/2023 | **Philips** | **IntelliSite UFS** | N/A | 40x (0.2500 μm) | **.isyntax** | 3,486 |
|  | **SPROB20** | Uppsala University Hospital, Uppsala, Sweden | 2020 | **Hamamatsu** | **NanoZoomer S360 C13210** | N/A | 40x (0.2204 μm) | **.tif** | 2,570 |
|  | **UKK** | Institute of Pathology, University Hospital Cologne, Cologne, Germany | N/A | **Hamamatsu** | **NanoZoomer S360** | N/A | 40x (0.2305 μm) | **.ome.tiff** | 50 |
|  | **WNS** | Hospital Wiener Neustadt, Wiener Neustadt, Austria | N/A | **Hamamatsu** | **NanoZoomer S360** | N/A | 40x (0.2305 μm) | **.ome.tiff** | 50 |

**Table 3. Reference standard protocols with respect to grading.** Reference standard protocols are divided into three categories: single reader, consensus and panel. In the single reader category, a sole reader assessed each slide. In the consensus category, assessments from multiple readers were combined based on site-specific criteria for consensus. In the panel category, readers provided independent assessments in a blinded manner. STHLM3=Stockholm3, SUH=Stavanger University Hospital, RUMC=Radboud University Medical Center, STG=Capio S:t Göran Hospital, KUH-1=Karolinska University Hospital, AMU=Aichi Medical University, AQ=Aquesta Uropathology, AUH=Aarhus University Hospital, KUH-2=Karolinska University Hospital morphological subtypes, MLP=Mehiläinen Länsi-Pohja, MUL=Medical University of Lodz, SCH=Synlab Switzerland, SFI=Synlab Finland, SFR=Synlab France, SPROB20=Spear Prostate Biopsy 2020, UKK=University Hospital Cologne, WNS=Hospital Wiener Neustadt.

| | Cohorts | | | Reference standard protocol | | | |
|---|---|---|---|---|---|---|---|
| Split | Cohort | Cohort subset | Slide number | Type | Total number of readers | Level | |
| Development, tuning and internal validation cohorts | STHLM3 | STHLM3 full cohort | 36,848 | Single reader (L.E.) | 1 | Slide | Patient |
| | | ImageBase | 90 | Panel | 23 | Slide | |
| | | PANDA Swedish private validation set | 212 | Consensus | 3 | | |
| | | STHLM3 morphological subtypes | 24 | Single reader (L.E.) | 1 | | |
| | SUH | SUH full cohort | 5,762 | Single reader | 14 | Slide | |
| | | Re-graded | 66 | Single reader (L.E.) | 1 | | |
| | RUMC | RUMC full cohort | 5,275 | Single reader | multiple | Slide | |
| | | PANDA RUMC tuning set | 195 | Panel | 3 | | |
| | | PANDA RUMC private validation set | 333 | Panel | 3 | | |
| | | Re-graded | 66 | Single reader (L.E.) | 1 | | |
| | STG | STG full cohort | 247 | Single reader (L.E.) | 1 | Slide | |
| | KUH-1 | KUH-1 full cohort | 330 | Single reader (L.E.) | 1 | Slide | Patient |
| External and partly external validation cohorts | AMU | AMU full cohort | 73 | Single reader | 1 | Patient | |
| | AQ | AQ full cohort | 136 | Single reader | 1 | Slide | |
| | AUH | AUH full cohort | 102 | Single reader | 1 | Slide | |
| | | Re-graded | 41 | Single reader (L.E.) | 1 | | |
| | KUH-2 | KUH-2 full cohort | 146 | Single reader (L.E.) | 1 | Slide | |
| | MLP | MLP full cohort | 1,964 | Single reader | multiple | Location | |
| | | Re-graded | 66 | Single reader (L.E.) | 1 | Slide | |

| | | | | | | |
|---|---|---|---|---|---|---|
| | **MUL** | MUL full cohort | | 1,959 | Consensus | 2 | Slide |
| | | Re-graded | 66 | Single reader (L.E.) | 1 | |
| | **SCH** | SCH full cohort | | 2,434 | Single reader | multiple | Location |
| | | Re-graded | 72 | Single reader (L.E.) | 1 | Slide |
| | **SFI** | SFI full cohort | | 537 | Single reader | multiple | Location |
| | | Re-graded | 67 | Single reader (LE) | 1 | Slide |
| | **SFR** | SFR full cohort | | 515 | Single reader | multiple | Location |
| | | Re-graded | 49 | Single reader (LE) | 1 | Slide |
| | **SPROB20** | SPROB20 full cohort | | 2,570 | Single reader | multiple | Patient |
| | | Re-graded | 50 | Single reader (LE) | 1 | Slide |
| | **UKK** | UKK full cohort | | 50 | Panel | 11 | Slide |
| | **WNS** | WNS full cohort | | 50 | Panel | 10 | Slide |

**Table 4. Reference standard protocols with respect to PNI and cribriform cancer.** Reference standard protocols are divided into three categories: single reader, consensus and panel. In the single reader category, a sole reader assessed each slide. In the consensus category, assessments from multiple readers were combined based on site-specific criteria for consensus. In the panel category, readers provided independent assessments in a blinded manner. The SUH, MUL and SCH cohorts do not have consistent original reporting on cribriform cancer and PNI. PNI=perineural invasion, WSI=whole slide image, STHLM3=Stockholm3, SUH=Stavanger University Hospital, AMU=Aichi Medical University, MUL=Medical University of Lodz, SCH=Synlab Switzerland.

| | Cohorts | | | Reference standard protocol | | | |
|---|---|---|---|---|---|---|---|
| **Split** | **Cohort** | **Cohort subset** | **Slide number** | **Type** | **Total number of readers** | **Level** | |
| Development, tuning and internal validation cohorts | STHLM3 | STHLM3 full cohort | 36,848 | Single reader (L.E.) | 1 | Slide | Patient |
| | | Re-assessed cribriform cancer (round 1) | 702 | Single reader (L.E.) | 1 | Slide | Pixel |
| | | Re-assessed cribriform cancer (round 2) | 304 | Panel | 9 | Slide | |
| | | Re-assessed PNI (round 1) | 485 | Single reader (L.E.) | 1 | Slide | Pixel |
| | | Re-assessed PNI (round 2) | 212 | Panel | 4 | Slide | |
| | SUH | SUH full cohort | N/A | N/A | N/A | N/A | |
| | | Re-assessed cribriform cancer (round 1) | 332 | Single reader (A.B.) | 1 | Slide | |
| | | Re-assessed cribriform cancer (round 2) | 200 | Single reader (L.E.) | 1 | Slide | |
| | | Re-assessed PNI (round 1) | 509 | Single reader (A.B.) | 1 | Slide | |
| | | Re-assessed PNI (round 2) | 185 | Single reader (L.E.) | 1 | Slide | |
| External validation cohorts | AMU | AMU full cohort | 73 | Single reader | 1 | Slide | |
| | MUL | MUL full cohort | N/A | N/A | N/A | N/A | |
| | | Re-assessed cribriform cancer | 276 | Consensus | 2 | Slide | |
| | | Re-assessed PNI | 276 | Consensus | 2 | Slide | |
| | SCH | SCH full cohort | N/A | N/A | N/A | N/A | |
| | | Re-assessed cribriform cancer | 56 | Single reader (H.S.) | 1 | Slide | |
| | | Re-assessed PNI | 94 | Single reader (B.D.) | 1 | Slide | |

**Table 5. Summary of slides representing various morphological subtypes.** The AQ cohort contains partly external validation data (scanner was used in the development) and fully external validation data (scanner was not used in the development). A single slide can be associated with multiple subtypes. Instead of morphological subtypes, the samples denoted with (*) represent other types of specimens than core needle biopsies. Besides assessing performance on unusual and potentially challenging morphologies, we will assess how the AI system intended for needle biopsies will respond to other specimen types and evaluate frameworks for automatically flagging outlier cases (Olsson *et al.*, 2022). STHLM3=Stockholm3, AQ=Aquesta Uropathology, KUH-2=Karolinska University Hospital morphological subtypes, PIN=prostatic intraepithelial neoplasia, TUR-P=transurethral resection of the prostate.

| Morphological subtype | Internal validation cohort | Partly external validation cohort | External validation cohort | |
|---|---|---|---|---|
| | STHLM3 (n=24) | AQ (n=58) | AQ (n=78) | KUH-2 (n=146) |
| Adenosis | 4 | 18 | 7 | 34 |
| Atrophy | 0 | 0 | 0 | 38 |
| Partial atrophy | 0 | 7 | 9 | 0 |
| Simple atrophy | 0 | 1 | 15 | 0 |
| Basal cell hyperplasia | 0 | 10 | 7 | 20 |
| Cancer of atrophic type | 7 | 1 | 3 | 2 |
| Clear cell cribriform hyperplasia | 0 | 0 | 5 | 3 |
| Cowper's glands | 0 | 2 | 14 | 6 |
| Foamy gland cancer | 0 | 4 | 2 | 13 |
| Increased number of glands | 0 | 0 | 5 | 0 |
| Postatrophic hyperplasia | 0 | 2 | 2 | 4 |
| Prostatectomy* | 0 | 1 | 4 | 0 |
| PIN-like cancer | 3 | 0 | 0 | 0 |
| Pseudohyperplastic cancer | 9 | 4 | 0 | 24 |
| Sclerosing adenosis | 0 | 4 | 2 | 0 |
| Seminal vesicle | 0 | 5 | 7 | 0 |
| Small cell cancer | 0 | 0 | 0 | 4 |
| TUR-P* | 0 | 13 | 4 | 0 |

**Table 6. Summary of slides with IHC staining confirming the diagnosis.** Number of slides stratified by ISUP grade with/without IHC staining performed for confirming the diagnosis. For the SCH and SFR cohorts, where pathology reporting was performed on anatomical location or patient level, the total summed numbers of slides associated with an IHC-supported diagnosis are shown. IHC=immunohistochemistry, ISUP=International Society of Urological Pathology, SUH=Stavanger University Hospital, SCH=Synlab Switzerland, SFR=Synlab France.

| Split | Cohort | IHC performed | Number of slides | | | | | | |
|---|---|---|---|---|---|---|---|---|---|
| | | | All | Benign | ISUP 1 (3+3) | ISUP 2 (3+4) | ISUP 3 (4+3) | ISUP 4 (4+4, 3+5, 5+3) | ISUP 5 (4+5, 5+4, 5+5) |
| **Internal validation** | SUH | Yes | 247 | 132 | 60 | 16 | 10 | 9 | 20 |
| | | No | 909 | 604 | 93 | 60 | 64 | 44 | 44 |
| **External validation** | SCH | Yes | 365 | 120 | 131 | 47 | 46 | 9 | 12 |
| | | No | 2,064 | 1,455 | 194 | 154 | 137 | 85 | 35 |
| | | Missing | 5 | 5 | 0 | 0 | 0 | 0 | 0 |
| | SFR | Yes | 116 | 66 | 41 | 4 | 1 | 1 | 0 |
| | | No | 398 | 306 | 46 | 24 | 2 | 9 | 6 |
| | | Missing | 1 | 1 | 0 | 0 | 0 | 0 | 0 |